\documentclass[nonblindrev]{informs3_noinforms}
\usepackage{pdfsync}
\usepackage{enumerate}
\usepackage{dsfont}
\usepackage{enumitem}
\usepackage[dvipsnames]{xcolor}
\usepackage{amsmath}
\usepackage[normalem]{ulem}

\usepackage{comment}
\usepackage{multirow}
\usepackage{subfigure}
\usepackage[hypertexnames=false]{hyperref}
\usepackage{cleveref}
\usepackage{float}
\hypersetup{hidelinks}
\usepackage[
  margin=1in,
  marginparwidth=2.25cm,
  marginparsep=0.1cm
]{geometry}

\usepackage[ruled,linesnumbered]{algorithm2e}

\OneAndAHalfSpacedXI %

\usepackage{endnotes}
\let\footnote=\endnote

\newcommand{\cL}{{\cal L}}
\newcommand{\bd}{\mathbf{d}}
\newcommand{\bii}{\mathbf{i}}
\newcommand{\bj}{\mathbf{j}}
\newcommand{\bbe}{\mathbb{E}}
\newcommand{\bE}{\mathbb{E}}

\def\bbR{{\Bbb{R}}} %
\def\bbP{{\Bbb{P}}}
\def\bbI{{\Bbb{I}}}

\newcommand{\GE}{\mathrm{GE}}
\newcommand{\EE}{\mathrm{EE}}
\newcommand{\ApproxE}{\mathrm{AE}}
\newcommand{\Pdim}{\mathrm{P\text{-}dim}}
\newcommand{\Pgamma}[1]{\mathrm{P}_{#1}\text{-}\mathrm{dim}}
\newcommand{\Rade}{\mathfrak{R}}
\newcommand{\Hlower}{\underline{H}}
\newcommand{\opt}{\mathrm{opt}}

\numberwithin{subcase}{case}

\numberwithin{subsubcase}{subcase}

\renewcommand{\theequation}{\thesection.\arabic{equation}}

\usepackage{color}              %
\usepackage{color-edits}

\usepackage[english]{babel}
\usepackage[autostyle, english = american]{csquotes}
\MakeOuterQuote{"}

\usepackage{natbib}
 \bibpunct[, ]{(}{)}{,}{a}{}{,}%
 \def\bibfont{\small}%

\TheoremsNumberedThrough     %
\ECRepeatTheorems

\EquationsNumberedThrough    %

\begin{document}

\RUNTITLE{VC Theory for Inventory Policies}
\TITLE{VC Theory for Inventory Policies}

\ARTICLEAUTHORS{
\AUTHOR{\bf Yaqi Xie}
\AFF{Booth School of Business, University of Chicago, Chicago, IL 60637}
\AFF{yaqi.xie@chicagobooth.edu}
\AUTHOR{\bf Will Ma}
\AFF{Graduate School of Business and Data Science Institute, Columbia University, New York 10027}
\AFF{wm2428@gsb.columbia.edu}
\AUTHOR{\bf Linwei Xin}
\AFF{School of Operations Research and Information Engineering, Cornell University, Ithaca, NY, 14853}
\AFF{lx267@cornell.edu}
\RUNAUTHOR{Xie, Ma and Xin}
}

\ABSTRACT{%
There has been growing interest in applying reinforcement learning (RL) to inventory management, either by optimizing over temporal transitions or by learning directly from full historical demand trajectories. This contrasts sharply with classical data-driven approaches, which first estimate demand distributions from past data and then compute well-structured optimal policies via dynamic programming. This paper considers a hybrid approach that combines trajectory-based RL with policy regularization imposing base-stock and $(s,S) $ structures. We provide generalization guarantees for this combined approach for several well-known classes in a $T$-period dynamic inventory model,
using tools from the celebrated Vapnik-Chervonenkis (VC) theory, such as the Pseudo-dimension and Fat-shattering dimension.
Our results have implications for regret against the best-in-class policies, and allow for an arbitrary distribution over demand sequences, which makes no assumptions such as independence across time.

Surprisingly, we prove that the class of policies defined by $T$ non-stationary base-stock levels exhibits a generalization error that does not grow with $T$, whereas the two-parameter $(s, S)$ policy class has a generalization error growing logarithmically with $T$. Overall, our analysis leverages specific inventory structures within the learning theory framework, and improves %
sample complexity guarantees even compared to existing results assuming independent demands.
}


\HISTORY{January 31, 2026}

\maketitle

\section{Introduction}
Inventory management plays a crucial role in businesses, ranging from manufacturing to distribution to retailing, and is a foundational research topic for operations and supply chain management.
Algorithmic, data-driven approaches to inventory decisions have long been central in the literature, reflecting the gradual shift in inventory management from human judgment to automated decision rules (see the survey \citealt{gijsbrechts2025ai}).
A typical approach to data-driven
inventory algorithms involves using past data to estimate demand distributions, and then optimizing over a time horizon using dynamic programming (DP) (e.g., \citealt{levi2007provably}, \citealt{cheung2019sampling}).
In addition, policies derived often have simple, interpretable structures (e.g., defined by base-stock levels, reorder points) that are specific to inventory, as established %
in decades of inventory theory.

Another approach is to "simulate" on available trajectories and find a policy that performs well,
an idea whose application to inventory problems dates back to \citet{glasserman1995sensitivity}.
More recent developments focus on learning a policy end-to-end via reinforcement learning (RL) without explicitly estimating a demand model.
These off-the-shelf methods can moreover incorporate high-dimensional contextual information through deep neural networks.
Indeed, by treating the inventory problem as a generic Markov Decision Process (MDP), off-the-shelf RL algorithms can be implemented (e.g., \citealt{gijsbrechts2022can}, \citealt{liu2023ai}), with a differentiable simulator approach that directly optimizes trajectory-level loss functions also recently being proposed (see \citealt{madeka2022deep}, \citealt{alvo2025deepreinforcement}).
Regardless, these generic approaches are typically unstructured and difficult to interpret, and could exhibit unstable performance especially in small-data regimes.

In this paper, we consider a hybrid approach that combines RL with regularization, where the learned policies are constrained to satisfy classical structures from inventory theory.
We show that this combined approach is particularly, surprisingly powerful for inventory management: it achieves better guarantees than the model-based approach even when the model is well-specified.
Specifically, we consider Empirical Risk Minimization (ERM), %
which bears resemblance to the differentiable simulator approach, but crucially, we restrict the policy class to the well-established base-stock and $(s, S)$ policies (see definitions in \Cref{sec-approximation}).
We study its learning guarantees in dynamic settings where inventory is durable and replenishment decisions span multiple time periods.

To elaborate, we formulate the problem as follows:
\begin{itemize}
\item $\bd\triangleq(d^t)_{t=1}^T$ denotes the demand sequence for a single item over $T$ periods; %
\item $\pi$ is a (deterministic) policy specifying how much inventory to order based on the current inventory state; %
\item $\ell(\pi,\bd)$ represents the loss, measuring the costs incurred over the $T$ periods from executing policy $\pi$ under demand sequence $\bd$, accounting for inventory holding costs, shortage costs, and fixed ordering costs (details in \Cref{sec-model});
\item we are given $N$ IID samples $\bd_1,\ldots,\bd_N$ drawn from an unknown distribution $\mathcal{D}$ over demand sequences of length $T$;
\item the goal is to learn a policy $\hat{\pi}$ with low out-of-sample expected loss (or ``Risk'')
when evaluated on the same distribution $\mathcal{D}$, defined as $R(\hat{\pi})$, where $R(\pi) \triangleq \mathbb{E}_{\bd\sim \mathcal{D}} [\ell( \pi ,\bd) ]$.
\end{itemize}
This is a standard formulation from the perspective of learning theory, but we do not study classification or regression problems as in supervised learning. Instead, we restrict attention to policy classes $\Pi$ defined by base-stock levels and reorder points that are provably optimal in certain inventory regimes (see \Cref{sec-approximation}).
In particular, our approach proposes the $\Pi$-constrained ERM solution $\hat{\pi}\in \arg \inf_{\pi\in \Pi}\hat{R}(\pi)$, where $\hat{R}(\pi) \triangleq \frac{1}{N} \sum_{i=1}^N \ell(\pi, \bd_i)$, and we evaluate its Estimation Error, defined as $\EE \triangleq R(\hat{\pi}) - R(\pi^*)$.
Here, $\pi^*\in \arg\inf_{\pi\in\Pi}R(\pi)$ denotes the best inventory policy in the same class $\Pi$ that a clairvoyant with full knowledge of the underlying distribution $\mathcal{D}$ could have used.

In contrast to the aforementioned theoretical work on data-driven inventory, we make no assumptions on the sequence distribution $\mathcal{D}$, such as random demands $d^t$ being independent across time $t$. Instead, we restrict $\pi^*$ to lie within the same policy class $\Pi$ as the learned policy $\hat{\pi}$.
Our guarantees on EE hold for arbitrary correlated demand sequences, and the comparison policy $\pi^*$ is optimal when demands are indeed independent across time. This implies that under the independence assumption, we are indeed comparing against the true optimal policy, and our measure of EE coincides with the standard notion of "regret" in the data-driven inventory literature.
Surprisingly, our results generally \textit{improve regret guarantees} for data-driven inventory, despite the fact that we are bounding the EE under arbitrarily correlated distributions, which is a \textit{strictly harder problem} than bounding the regret under independent distributions.
This demonstrates the power of our $\Pi$-constrained ERM approach, %
which avoids the need to analyze how DP errors propagate under independence assumptions.

All in all, our $\Pi$-constrained ERM approach exhibits both striking theoretical guarantees and excellent simulation performance, as discussed in the following subsections.
Moreover, ideas based on this $\Pi$-constrained ERM approach have been deployed in modern high-dimensional settings and demonstrated to outperform unconstrained algorithms (\citealp{xie2025deepstock}).

\subsection{Theoretical Results---(Nearly) Horizon-free Estimation Error}\label{sec:outlineResults}

It can be shown that the EE of the policy $\hat{\pi}$ is upper-bounded by the (uniform) Generalization Error $\GE \triangleq \sup_{\pi \in \Pi}\left\{ R( {\pi}) - \hat{R}( {\pi}) \right\}$, which measures the difference between in-sample and out-of-sample risk for any policy $\pi$; see \eqref{ineq-ee-ge}. Our analysis focuses on the GE and achieves the following.
\begin{itemize}
\item It provides an upper bound on the GE for any distribution $\mathcal{D}$. As a result, it provides an upper bound on the true risk $R(\pi)$ of the learned policy $\pi$, of the form $R(\pi)\le\hat{R}(\pi)+\GE$ (where $\hat{R}(\pi)$ can be evaluated in-sample), even if $\pi$ differs from $\hat{\pi}$.

\item When demands are actually independent over time (as assumed by the classical approach), our analysis also provides a new theoretical framework for analyzing methods that first estimate per-period demand distributions and then optimize via DP (\citealt{levi2007provably}, \citealt{cheung2019sampling}). We refer to this resulting policy as Product ERM (PERM); see \Cref{sec-PERM} for further details.
\end{itemize}

We upper-bound the GE using the celebrated Vapnik-Chervonenkis (VC) theory. Specifically, the GE is upper-bounded by the Rademacher complexity, which is further upper-bounded using variants of the VC-dimension, such as the Pseudo-dimension and the Fat-shattering dimension (see definitions in \Cref{sec-VC}).
These complexity measures are combinatorial in nature, distribution-free, and quantify the capability of a policy class to fit complex patterns.
Particularly, we leverage the specific structures of inventory problems and policies to derive both upper and lower bounds on these VC-dimension variants.

In our theoretical results,
demands, cost parameters, and policy parameters are assumed to be bounded, and costs are normalized by the horizon length $T$, so that the loss of any policy under any demand realization remains bounded by a constant independent of $T$.
Under this normalized loss function $\ell(\pi,\bd)$, we investigate how the uniform GE of a policy class $\Pi$ scales with the sample size $N$ and the horizon length $T$.
We consider three policy classes for $\Pi$: stationary base-stock $S$ policies, stationary $(s,S)$ policies, and non-stationary base-stock policies $(S^t)$, in a backlogged demand setting with a deterministic lead time. (See \Cref{sec-model} for the inventory dynamics of these policies and the associated costs.)

The GE is a random variable with respect to (w.r.t.)\ the realization of the $N$ samples, and
we bound its expectation.
We present the following nearly-tight characterizations:
\begin{enumerate}
\item \textbf{$S$ policies}: Expected GE is $\Theta(\sqrt{1/N})$;
\item \textbf{$(s,S)$ policies}: Expected GE is $O(\sqrt{\log T/N})$ and $\Omega(\sqrt{(\log T/\log\log T)/N})$;
\item \textbf{$(S^t)$ policies}: Expected GE is $\Theta(\sqrt{1/N})$.
\end{enumerate}
Interestingly, $(S^t)$ policies have a GE independent of $T$, despite involving $T$ parameters.
In stark contrast, $(s,S)$ policies exhibit a GE that grows with $T$, despite having only two parameters.
We note there is a simple lower bound of $\Omega(\sqrt{1/N})$ due to statistical noise \citep[see e.g.,][Prop.~1]{zhang2020closing}, showing that results 1.\ and 3.\ are tight.

Our results can also be stated in equivalent forms, such as sample complexity. In this form, they imply that achieving an expected GE of at most $\epsilon$ requires a number of samples that grows with the horizon length $T$ for $(s,S)$ policies, but remains independent of $T$ for $S$ and $(S^t)$ policies. Specifically, by setting our upper bounds on the expected GE to be $\epsilon$ and solving for $N$, our results imply that the sample complexities for learning base-stock, $(s, S)$, and $(S^t)$ policies are $O(1/\epsilon^2)$, $O(\log T/\epsilon^2)$, and $O(1/\epsilon^2)$, respectively.
All of these expectation bounds can also be converted into high-probability bounds using standard concentration inequalities (see \Cref{sec-estimation}).

\subsection{Practical Insights and Further Evidence from Simulations}\label{sec:insights}
First, our results show that under the $\Pi$-constrained ERM approach, the (normalized) regret in inventory problems grows at most logarithmically with the time horizon $T$, indicating minimal dependence on $T$. That is, apart from the trivial accumulation of errors over $T$ periods (which is offset by normalization), data-driven inventory problems do not become significantly harder as the horizon length increases.
This finding is further supported by numerical simulations in \Cref{subsec-num-T}.
Collectively, our results challenge the conventional wisdom in the literature that learning near-optimal policies requires significantly more samples as the horizon length increases.

Comparing these three policy classes, we can further conclude that the number of parameters is not always an accurate measure of policy complexity. As we showed, the EE and GE for both $S$ policies and $(S^t)$ policies are of the same $T$-free order, $\Theta(\sqrt{1/N})$, whereas those for $(s,S)$ policies grow mildly with $T$, on the order of $\Theta(\sqrt{\log T/N})$ (up to a $\log\log T$ factor).
These tight bounds on EE provide insight into choosing a policy class based on the trade-off between EE and approximation error (AE; see \Cref{sec-approximation}). Particularly, restricting the policy class can reduce EE at the cost of introducing AE due to its reduced expressiveness. This trade-off may be worthwhile under limited data, as the EE term tends to dominate when $N$ is small, as shown empirically in \Cref{subsec-num-N}.
This highlights an important managerial implication of our findings: the principle of ``learning less is more'' in inventory management, where depending on the amount of available data, one may want to learn within a ``simple'' policy class with lower EE.

Finally, the comparison between the $\Pi$-constrained ERM and PERM approaches highlights a fundamental trade-off between sample efficiency and the preservation of temporal dependence in demand trajectories. While PERM is generally more sample efficient, it loses a significant amount of temporal information when the demand process deviates from independent. In contrast, $\Pi$-constrained ERM can capture demand correlations over time while still learning well-structured policies, but may suffer from data scarcity when the sample size is extremely small.
Fortunately, for the inventor context, our results show that $\Pi$-constrained ERM also exhibits surprisingly strong sample efficiency. Indeed, with a modest sample size, ERM achieves performance comparable to PERM when the demand process is close to independent, and outperforms PERM when the demand process exhibits strong temporal dependence, as demonstrated numerically in \Cref{subsec-num-perm}.

\section{Literature Review}
In this section, we review related work on data-driven inventory control, generic RL theory, stochastic optimization, and data-driven algorithm design.  We place particular emphasis on comparing our theoretical bounds with existing results in the literature.

\subsection{Comparison with Existing Theoretical Results}\label{sec-comparison}

Our paper falls under the domain of offline learning with finite-sample guarantees. Below, we compare our results with existing literature in data-driven inventory control, RL theory, and stochastic optimization, highlighting the surprising nature of our EE bounds: constant in the horizon length $T$ for $(S^t)$ policies and logarithmic in $T$ for $(s,S)$ policies.

We first highlight that our $O(\sqrt{1/N})$ result for the EE of $S$ policies provides a novel way to establish $O(\sqrt{1/N})$ regret even in the "Newsvendor" special case (where $T=1$) by using the Pseudo-dimension concept from learning theory, significantly differing from the existing Newsvendor literature (see \citealt{chen2024survey}).

For $(s, S)$ policies, the closest related work is \cite{fan2024don}, who study an infinite-horizon discounted setting with IID demands over time.  Ignoring polynomial terms in the state and action space cardinalities and making a best-effort conversion from sample complexity for IID demands, their result would imply a sample complexity of $O(T /\epsilon^2)$ in our setting by plugging in $1-\gamma = 1/T$
(where $\gamma$ denotes the discount factor), which is still worse than our sample complexity guarantee of $O(\log T/\epsilon^2)$.

For $(S^t)$ policies, classical data-driven inventory control papers typically assume independent demands, under which the optimal policy is known to lie within the $(S^t)$ policy class. In particular, \cite{levi2007provably} and \cite{cheung2019sampling} establish sample complexity bounds of $O(T^5 \log T/\epsilon^2)$ and $O(T^6 \log T/\epsilon^2)$, respectively, in a slightly different setting with multiplicative regret.
These bounds imply the same sample complexity guarantees when converted to our setting, much worse than our bound of $O(1/\epsilon^2)$,
highlighting the surprising nature of our $T$-free sample complexity (that applies to EE without the independence assumption).
Assuming independence, \citet{qin2023sailing} recently establish a sample complexity of $O(T^3/\epsilon^2)$ for additive regret under unnormalized loss, which when converted to our setting implies a sample complexity of $O(T/\epsilon^2)$, still growing with $T$.

Meanwhile, general RL theory papers study the sample complexity of finding a near-optimal policy for Markov decision processes, typically in the \textit{tabular} setting (i.e., finite state and action spaces), which does not exactly capture our inventory setting.
Nonetheless, closely related works are \citet{yin2021near} and \cite{li2024breaking}, who consider an episodic (i.e., finite-horizon) time-inhomogeneous (i.e., corresponding to our setting with independent, non-identical demands) RL problem and establish a sample complexity that translates to $O(T \log T / \epsilon^2)$ in our setting with a normalized loss function.
They consider offline RL, assuming the data sufficiently covers transitions under relevant state-action pairs, a condition that is indeed satisfied in our inventory problem with perfect hindsight evaluation.
Meanwhile, \cite{zhang2022horizon} study online episodic RL, which can also apply to our inventory setting if each trajectory $i=1,\ldots,N$ is used to simulate one episode.
They establish a regret bound of $O(\sqrt{\log N/N})$ that does not depend on the horizon length; however, this result holds only under time-homogeneous transitions. Their result translates to $O(\sqrt{T\log N/N})$ under time-inhomogeneous transitions.
We note that the horizon-independent guarantee relies on a normalization that bounds the total reward (or loss) by a constant, thereby eliminating reward-scaling effects, as discussed in \citet{jiang2018open}. This normalization is standard in the RL literature (e.g., \citealt{zhang2022horizon}) and is adopted throughout our paper.

A final stream of related literature is stochastic optimization, whose papers typically treat the sample sequence as a high-dimensional vector, ignoring correlations between entries, which corresponds to arbitrary correlated demands in our model. These papers analyze the sample complexity of Sample Average Approximation (SAA), which involves minimizing the in-sample risk using $\hat{\pi}$,
w.r.t.\ the dimension of the decision space (i.e., $T$ for $(S^t)$ policies).
Our problem with $(S^t)$ policies can be framed as a stochastic optimization problem where the loss function is Lipschitz continuous in the decision variables.
The sample complexity for such problems is $O(T/\epsilon^2)$, as shown in, e.g., Theorem 5 of \cite{shalev2009stochastic} and Theorem 5.18 of \cite{shapiro2021lectures}, and it likewise grows with $T$. These results apply to the SAA (in-sample optimal) solution, which may be intractable to compute exactly.

We summarize results relating to our $O(\sqrt{1/N})$ bound for $(S^t)$ policies in \Cref{table-compare}.  As shown, all of these existing results have a dependence on the horizon length $T$.  This is necessarily the case for the RL and SAA results, which consider more general Markov decision processes or stochastic optimization problems.  Meanwhile, the inventory results accrue a dependence on $T$ through the PERM approach, instead of using our $\Pi$-constrained ERM approach.

\begin{table}[h]
\renewcommand{\arraystretch}{1.6}
\centering
\caption{Finite-Sample Estimation Error Bounds for $(S^t)$ Policies}\label{table-compare}
\begin{tabular}{|l|l|}
\hline
Independent Demands & Arbitrary Demands \\ \hline
\begin{tabular}[c]{@{}l@{}}
\cite{levi2007provably}: $O(\sqrt{T^5 \log T /N})$ \\
\cite{cheung2019sampling}: $O(\sqrt{T^6 \log T /N})$ \\
\cite{qin2023sailing}: $O(\sqrt{T/N})$\\
\cite{yin2021near}, \cite{li2024breaking}: $O(\sqrt{T\log T/N})$ \\
\cite{zhang2022horizon}: $O(\sqrt{T\log N/N})$
\end{tabular}
&
\begin{tabular}[c]{@{}l@{}}
\cite{shalev2009stochastic}: $O(\sqrt{T/N})$ \\
\cite{shapiro2021lectures}: $O(\sqrt{T/N})$
\end{tabular} \\ \hline
\multicolumn{2}{|c|}{\textbf{Our Paper: $O\left(\sqrt{1/N} \right)$}} \\ \hline
\end{tabular}
\end{table}

\subsection{Further Related Work}\label{sec-literature}

\noindent\textbf{Data-driven inventory management:}
Besides the papers on multi-period dynamic inventory models discussed in \Cref{sec-comparison}, most data-driven inventory management studies focus on the Newsvendor model, including \cite{levi2015data, lin2022data, besbes2023big, besbes2023quality}.
In contrast to the uncensored demand setting assumed in our paper, another stream of the literature considers settings in which offline demand data are censored due to lost sales.
To the best of our knowledge,  existing theoretical guarantees under censored demand primarily focus on base-stock or $(s,S)$ policy classes. For example, \citet{huh2009nonparametric}, \citet{shi2016nonparametric}, and \citet{agrawal2022learning} study base-stock policies, while \citet{ban2020confidence}, and \cite{yuan2021marrying} analyze $(s,S)$ policies.
Given the well-documented difficulty of censored-demand settings, we view our focus on uncensored demand as a natural and tractable starting point.
Moreover, our paper focuses on nonparametric approaches, as well as a model-based PERM approach used for comparison. These stand in contrast to parametric data-driven methods that estimate demand distributions within a prescribed parametric family.

It is quite surprising that the study of VC-type  complexity, even for the simplest classes such as base-stock and ($s, S$) policies, is missing; this is the gap we aim to fill with this paper. It is worth noting that, recently, \cite{chen2022using} and \cite{han2023deep} apply existing bounds on the generalization error of neural networks directly to a Newsvendor problem. Their model is based on the contextual Newsvendor model that is pioneered by \citet{ban2019big}. By contrast, our analysis focuses on the dynamics of inventory state transitions over time and how this relates to learning-theoretic dimensions. We note that recently, \citet{guan2022randomized} ask similar questions in the context of optimal stopping, which also involves dynamic decision-making over time but is otherwise unrelated to inventory.

\noindent\textbf{Data-driven algorithm design:}
We are tuning the parameters of policy classes based on historical sequences (instances), a process known as \emph{data-driven algorithm design} \citep{balcan2020data,gupta2020data}.
However, we derive problem-specific results by analyzing the sensitivity of inventory state transitions to policy parameters, which appear difficult to deduce from general theory (e.g., \citealp{balcan2021much}).
We also believe that inventory problems, which naturally combine dynamic decision-making, state transitions, and distributions over arbitrary demand sequences, are rather underexplored in this area and may inspire future data-driven algorithm design for other online algorithms problems.
We note that online algorithms can be tuned based on past arrival sequences by using the so-called "Algorithms with Predictions" framework \citep{lykouris2021competitive}; however, the theoretical results therein are of a completely different nature.

\section{Model and Preliminaries}\label{sec-model}

In this paper, we consider a discrete-time backlogged inventory model with lead time over a finite-horizon. There is a lead time $L\in \mathbb{Z}_{\geq 0}$ between when an order is placed and when it is delivered. There are  $T+L$ periods in the entire horizon. The demand process follows a stochastic process with distribution $\mathcal{D}$, which can be non-stationary or correlated across time. A realized demand sequence $\bd\triangleq (d^1, \ldots, d^{T+L})$ consists of demand $d^t $ for period $ t= 1, \ldots, T+L$.
Let $h$ and $b$ be the per-unit holding and backlogging costs, respectively, and let $K$ be the fixed ordering cost.
Without loss of generality, we normalize $h, b\in [0,1]$, %
assume $K\geq 0$, and assume that $\mathcal{D}$ is supported on $[0,U]^{T+L}$ for a given constant $U>0$.
The decision-maker must specify an inventory control policy $\pi$ that determines an order quantity at each time point with the objective of minimizing the expected average loss over the entire horizon.

Let $x^t$ be the inventory level at the beginning of period $t$. Let $ (q^{t-L}, \ldots, q^{t-1})$ be the pipeline inventory at the beginning of period $t$, where $q^t$ denotes the inventory ordered in period $t$ that is to arrive in period $t+L$. We assume empty initial pipeline: $q^t = 0$ for $t\leq 0$. Let $I^t$ be the inventory position at the beginning of period $t$ before ordering, namely, $I^t = x^t + \sum_{t'=t-L}^{t-1} q^{t'}$. Let
$y^t$ be the inventory level after replenishment but before seeing demand in period $t$, namely,  $y^t  = x^t + q^{t-L}$. Demands are backlogged and the inventory level in period $t+1$ is updated according to $ x^{t+1} = y^t - d^t$. The sequence of events in period $t$ is summarized as follows:
\begin{enumerate}
    \item The decision-maker observes the inventory level $x^t $ and pipeline vector $\left(q^{t-L}, \ldots, q^{t-1}\right)$;
    \item A new order $q^t$ is placed based on policy $\pi$;
    \item The inventory level is updated according to $y^t =  x^t + q^{t-L}$;
    \item Demand $d^t$ is realized and  the loss $\ell^t(\pi, \bd)$ is incurred.
\end{enumerate}
In particular, define $c(x) \triangleq  h[x]^+ +  b[-x]^+$, where $[\cdot]^+\triangleq\max\{\cdot,0\}$, and the loss  $\ell^t(\pi, \bd) \triangleq c(y^t - d^t)+ K \cdot \bbI\{q^{t-L} > 0\}$.
Here, $\ell^t$ includes the holding and backlogging costs in period $t$, as well as the fixed ordering cost associated with the order placed in period $t-L$ that is arriving during period $t$. Define $\ell(\pi, \bd) \triangleq \frac{1}{T} \sum_{t=L+1}^{T+L} \ell^t(\pi, \bd)$ as the losses from period $L+1$ to period $T+L$ of a policy $\pi$ (averaged over time) on a given demand sequence $\bd$. Here, we ignore the losses associated with periods 1 to $L$ because our first action $q^1$ only impacts losses starting from period $L+1$.

For the data-driven problem, the training dataset contains $N$ demand samples $\bd_1, \ldots, \bd_N$ drawn IID from  $\mathcal{D}$, and each of the samples $\bd_i \triangleq (d_i^t)_{t=1}^{T+L}$ consists of demand realizations from period $1$ to  period $T+L$. Note that this assumption is also made in \cite{ban2020confidence} and \citet[\S7]{madeka2022deep}. Here, the $N$ samples can be interpreted as the demands of $N$ similar products over $T+L$ periods, or as $N$ records of one product's demands over $T+L$ consecutive periods (e.g., each sample $i=1,\ldots,N$ represents one year and consists of $T+L$ periods).
The samples can also correspond to the life cycles of the $N$ past generations of a product (e.g.,\ a magazine or video game), %
whose sales are correlated within cycle (e.g.,\ one generation was popular and had high demands throughout its life cycle) but each cycle can be viewed as an IID realization.

\subsection{Estimation Error}\label{sec-estimation}
Recall that $R(\pi) = \mathbb{E}_{\bd\sim \mathcal{D}} [\ell( \pi ,\bd) ]$ and $\hat{R}(\pi) = \frac{1}{N} \sum_{i=1}^N \ell(\pi, \bd_i) $ denote the true risk and the empirical risk of the policy $\pi$, respectively, and $\pi^* \in \arg \inf_{\pi \in \Pi} R (\pi)$ denotes the best policy within the policy class $\Pi$. Define $\cL(\Pi)  \triangleq \{\ell ( \pi ,\cdot): \pi \in \Pi \}$ as the set of loss functions induced by the policy class $\Pi$, where each function corresponds to a policy $\pi \in \Pi$.

Given the dataset $(\bd_i)_{i=1}^N$, the decision-maker uses a procedure to specify a policy $\pi \in \Pi$. We focus on the $\Pi$-constrained Empirical Risk Minimization (ERM) approach,
which produces an output  $\hat{\pi} \in \arg \inf_{\pi\in \Pi} \hat{R}(\pi)$ minimizing in-sample risk.
The question of interest is to bound the expected risk gap between the policy $\hat{\pi}$ produced by ERM and the best one $\pi^*$ within the policy class $\Pi$. This gap is known as the Estimation Error (EE) of ERM w.r.t.\ the function class  $\cL(\Pi)$ in the literature (e.g., Section 5.2 of \citealt{shalev2014understanding}), denoted by
$\EE(\cL(\Pi)) \triangleq R(\hat{\pi}) - R(\pi^*).$
The expected EE w.r.t.\ the $N$ samples $(\bd_i)_{i=1}^N $  can be upper-bounded as follows:
\begin{align}\label{ineq-ee-ge}
    \mathbb{E}_{ \bd_1, \ldots, \bd_N  \sim \mathcal{D} }[\EE(\cL(\Pi))]& =  \mathbb{E}_{ \bd_1, \ldots, \bd_N  \sim \mathcal{D} }\left[R(\hat{\pi}) - \hat{R}(\hat{\pi}) + \hat{R}(\hat{\pi}) - \hat{R}(\pi^*) +  \hat{R}(\pi^*) - R(\pi^*) \right] \nonumber\\
    &\leq  \mathbb{E}_{ \bd_1, \ldots, \bd_N  \sim \mathcal{D} }\left[R(\hat{\pi}) - \hat{R}(\hat{\pi}) \right] \nonumber\\
    &\leq \mathbb{E}_{ \bd_1, \ldots, \bd_N  \sim \mathcal{D} } \left[\sup_{\pi \in \Pi}\left\{ R( {\pi}) - \hat{R}( {\pi}) \right\}\right],
\end{align}
where the  first inequality comes from the optimality of $\hat{\pi}$ and the fact that $\mathbb{E}\left[\hat{R}(\pi^*) - R(\pi^*) \right]=0$.
The supremum term in \eqref{ineq-ee-ge} is referred to as the (uniform) Generalization Error (GE) or the representativeness of the policy class $\Pi$ in the literature, which describes the worst-case scenario of overfitting within the policy class.
Formally, the GE of the policy class $\Pi$ w.r.t.\ the function class  $\cL(\Pi)$ is defined as $\GE(\cL(\Pi)) \triangleq \sup_{\pi \in \Pi}\left\{ R( {\pi}) - \hat{R}( {\pi}) \right\}.$
It quantifies the generalization ability of a policy class $\Pi$, namely, the worst-case gap between in-sample and out-of-sample risks of a policy within $\Pi$.

In this paper, we focus on analyzing the expected GE given samples $(\bd_i)_{i=1}^N $, with the objective of quantifying the learning complexity intrinsic to three different policy classes by leveraging the VC-dimension variants, such as the Pseudo-dimension and the Pseudo$_\gamma$-dimension (see \Cref{sec-VC}). More specifically, we aim to upper-bound the expected GE w.r.t.\ both the horizon length $T$ and the sample size $N$, which would then lead to an upper bound on the expected EE. Intuitively, with a larger horizon length $T$, the inventory dynamics simulated by the $N$ samples become more unstable and could lead to greater overfitting. Surprisingly, we show that the expected GE in inventory problems exhibit nearly $T$-free dependence.

We remark that the literature often focuses on high-probability bounds for the EE instead of expectation bounds, in which case the definition of GE can be $\sup_{\pi \in \Pi} \left\{ \left|R( {\pi}) - \hat{R}( {\pi})\right| \right\}$.
Regardless, our expectation upper bounds imply high-probability upper bounds after applying standard tools such as McDiarmid's inequality (e.g., Theorem 26.5 of \citealt{shalev2014understanding}), and our analysis based on $\sup_{\pi \in \Pi} \left\{R( {\pi}) - \hat{R}( {\pi})  \right\}$ implies upper bounds on $\sup_{\pi \in \Pi} \left\{ \left|R( {\pi}) - \hat{R}( {\pi})\right| \right\}$ after scaling by a constant factor.

\subsection{Approximation Error}\label{sec-approximation}
The Approximation Error (AE) of a policy class $\Pi$ is defined as the true risk gap between the best policy within the class, $\pi^* \in \arg\inf_{\pi\in \Pi} R(\pi)$, and the true optimal policy, $\pi^{\opt} \in \arg\inf_{\pi} R(\pi)$. More formally, $\ApproxE(\cL(\Pi)) \triangleq R(\pi^*) - R(\pi^{\opt})$.
It depends only on the richness of the policy class $\Pi$ and is independent of the sample size $N$.

The true risk gap between the ERM policy $\hat{\pi}$ and the true optimal policy $\pi^{\opt}$ is equal to the sum of $\EE(\cL(\Pi))$ and $\ApproxE(\cL(\Pi))$.
In general, there is a trade-off between EE and AE: a richer policy class results in a lower AE but might lead to overfitting and a larger EE. The problem of designing policy classes that balance this trade-off has been explored in the learning theory literature. One common approach involves using prior or domain knowledge about the specific problem to impose structural restrictions on the policy class $\Pi$. Moreover, because AE does not depend on the sample size $N$, it motivates analyzing EE w.r.t.\ $N$ for specific policy classes.

We focus on three policy classes that have been extensively studied in the inventory literature and are widely used in practice:
\begin{itemize}
\item $\Pi_S \triangleq \{S: S\in [0,H]\}$: the class of all base-stock policies with a stationary base-stock level $S$. These policies order to bring the inventory position up to $S$ whenever the inventory position falls below $S$. See \Cref{sec-base-stock} for a formal specification.
\item $\Pi_{(s, S)} \triangleq \{(s, S): \Hlower\leq s\leq S \leq H, S\geq 0\}$: the class of all $(s, S)$ policies. These policies order to bring the inventory position up to $S$ whenever the inventory position is below the reorder point $s$. See \Cref{sec-SS-ub} for a formal specification.
\item $\Pi_{(S^t)} \triangleq \{(S^t)_{t=1}^{T+L}: (S^t)_{t=1}^{T+L}\in [0, H]^{T+L}\}$: the class of all base-stock policies with non-stationary levels $(S^t)_{t=1}^{T+L}$. These policies order to bring the inventory position up to $S^t$ when the inventory position falls below $S^t$ in period $t$. See \Cref{sec-ST} for a formal specification.
\end{itemize}
Here, $H$ and~$\Hlower$ are fixed constants that depend on $h$, $b$, and $U$; see the corresponding sections for their explicit definitions. Note that we assume that the parameter domains are continuous, allowing $s,S,S^t$ to assume fractional values.
To ensure that a replenishment is always ordered in period $t=1$, we assume that $x^1$ is a non-positive fixed constant satisfying $x^1 \leq \Hlower$ in the $(s,S)$ setting, and that $x^1\le 0$ also in the $S$ and $(S^t)$ settings.

From classical inventory theory (e.g., \citealt{simchi2013logic,snyder2019fundamentals}), it is well established that these policies can achieve optimality (i.e., zero AE) in backlogged inventory models under certain conditions:
\begin{itemize}
    \item In settings where the fixed cost $K=0$ and demands are IID across time, an optimal base-stock policy exists with a stationary base-stock level for each period, falling within the class $\Pi_S$ with $H = (L+1)U$.
    \item In settings where $K>0$ and demands are IID across time, an $(s, S)$ policy with both stationary order-up-to level $S$ and reorder point $s$ is asymptotically optimal (as $T\to \infty$).
    \item In settings where the fixed cost $K=0$ and demands are independently but not necessarily identically distributed across time, an optimal $(S^t)$ policy exists. This policy features a time-dependent base-stock level for period $t$, which falls within the class $\Pi_{(S^t)}$ with $H = (L+1)U$.
\end{itemize}

Our results for base-stock policies and $(s,S)$ policies make no assumptions about the cost structure or the stationarity/independence of demand over time. In contrast, our result for $(S^t)$ policies does require the absence of fixed costs, i.e.,~$K=0$. Regardless, because we never assume independence, we are always upper-bounding the EE in settings where the AE is non-zero.

\subsection{Rademacher Complexity, VC-Dimension, Pseudo-Dimension, and Pseudo$_\gamma$-Dimension}\label{sec-VC}
In this section, we review several well-known results from learning theory by using the notions of Rademacher complexity, VC-dimension, Pseudo-dimension, and Pseudo$_\gamma$-dimension.
Throughout the paper, we consider general function classes $\mathcal{F}(\Pi) = \left\{f(\pi, \cdot): \pi\in \Pi\right\}$, where for each policy $\pi$, the induced function $f(\pi,\cdot): [0,U]^{T+L}\to \bbR$ may represent either a loss function or an inventory level.
Let $[m]\triangleq \{1,\ldots,m\}$.
We first define a fundamental concept known as Rademacher complexity and present a classical inequality from the literature to bound the expected GE (e.g., Lemma 26.2 of \citealt{shalev2014understanding}).
\begin{definition}[Rademacher complexity]
    The empirical Rademacher complexity of a function class $\mathcal{F}(\Pi) = \{f(\pi, \cdot): \pi\in \Pi\}$ w.r.t.\ a dataset $(\bd_i)_{i=1}^N\in ([0, U]^{T+L})^N$ is defined as
    \[
   \Rade \left(\mathcal{F}(\Pi) \circ (\bd_i)_{i=1}^N \right)\triangleq\ \mathbb{E}_{\sigma_1,\ldots, \sigma_N}\left[\sup_{\pi \in \Pi} \frac{1}{N} \sum_{i=1}^N \sigma_i \cdot f(\pi, \bd_i)   \right],
    \]
    where $\sigma_1,\ldots, \sigma_N$ are random variables IID drawn from the Rademacher distribution, i.e., $\bbP(\sigma_i = +1) = \bbP(\sigma_i = -1) = 1/2$.
\end{definition}
\begin{proposition}\label{prop-rademacher}
For any distribution $\mathcal{D}\in \Delta([0, U]^{T+L})$, sample size $N\in\mathbb{Z}_{>0}$, and  function class $\mathcal{F}(\Pi) = \left\{f(\pi, \cdot): \pi\in \Pi\right\}$, the following inequality holds:
$$
\mathbb{E}_{ \bd_1, \ldots, \bd_N  \sim \mathcal{D} }\left[ \GE (\mathcal{F}(\Pi)) \right]    \leq   2  \mathbb{E}_{ \bd_1, \ldots, \bd_N  \sim \mathcal{D} } \left[  \Rade\left(\mathcal{F}(\Pi) \circ (\bd_i)_{i=1}^N \right) \right].
$$
\end{proposition}

The empirical Rademacher complexity measures the richness and complexity of the function class $\mathcal{F}(\Pi)$ given the dataset $(\bd_i)_{i=1}^N$. However, computing the empirical Rademacher complexity is NP-hard (see, e.g., the second paragraph on page 29 of \citealt{mohri2018foundations}). To address this challenge, combinatorial concepts such as the VC-dimension (\citealt{vapnik1971uniform}) and the Pseudo-dimension (\citealt{Pollard84}) have been introduced.
Compared to Rademacher complexity, VC-dimension and Pseudo-dimension are often relatively easier to bound or estimate, providing a simpler alternative for bounding the expected GE.

\begin{definition}[Pseudo-dimension] \label{def:pseudo}
    A dataset  $(\bd_i)_{i=1}^m\in \left([0, U]^{T+L}\right)^m$ is shattered by a function class $\mathcal{F}(\Pi) = \{f(\pi, \cdot): \pi\in \Pi\}$ with witnesses $\tau_1,\ldots, \tau_m$ if
    there exist $\tau_1,\ldots, \tau_m \in \bbR$ such that for any $A\subseteq [m]$, there exists $\pi \in \Pi$  such that
    $
       f(\pi, \bd_i) > \tau_i \ \forall i\in A,
    $ and  $
       f(\pi, \bd_i) \leq \tau_i  \ \forall i\notin A
    $.
    The Pseudo-dimension of $\mathcal{F}(\Pi)$ denoted by $\Pdim(\mathcal{F}(\Pi))$ is the maximum size of a dataset that can be shattered by $\mathcal{F}(\Pi)$.
\end{definition}

In \Cref{def:pseudo}, if $f(\pi,\cdot)=\ell(\pi,\cdot)$ is the loss function, then $A$ can be interpreted as the subset of samples $i$ on which $\pi$ performs "poorly," i.e., it incurs a loss exceeding the threshold $\tau_i$.
Note that an equivalent definition of shattering is that there exist thresholds $\tau_1,\ldots, \tau_m \in \bbR$ such that
\[
   \left| \{\left(\bbI\{f (\pi, \bd_1) > \tau_1\}, \ldots,  \bbI\{f (\pi, \bd_m) > \tau_m\} \right): \pi \in \Pi\} \right| = 2^m,
\]
where $|\cdot|$ denotes the cardinality of a set.
Here, shattering refers to the ability to realize every possible combination of binary labels on a set of points via the indicator $\bbI\{f (\pi, \bd_i) > \tau_i\}$. The Pseudo-dimension quantifies the richness of a function class in terms of its ability to fit all possible datasets, that is, a higher Pseudo-dimension indicates a more complex class, which may increase the risk of overfitting.
Moreover, the Pseudo-dimension generalizes the concept of VC-dimension, which is designed only for binary classification.
Specifically, when the function class is binary-valued, we can trivially set $\tau_i=0$ for all $i\in [m]$, reducing the Pseudo-dimension to the classical definition of VC-dimension.
We refer interested readers to Section \ref{sec-background} in the appendix, which provides some examples to illustrate the calculation of VC-dimension and Pseudo-dimension.

We next present a classical result in learning theory that uses Pseudo-dimension to derive a distribution-free bound on the Rademacher complexity (e.g., Section 5.2 of \citealt{bousquet2003introduction}).

\begin{proposition}\label{prop-P-dim}
    For any distribution $\mathcal{D}\in\Delta([0,U]^{T+L})$, sample size $N\in\mathbb{Z}_{>0}$, and function class $\mathcal{F}(\Pi)=\{f(\pi,\cdot):\pi\in\Pi\}$ where $f(\pi,\bd)\in[0,B]$ for all $\pi\in\Pi$ and $\bd\in[0,U]^{T+L}$, there exists an absolute constant $C_0>0$ such that
    $$
    \mathbb{E}_{ \bd_1, \ldots, \bd_N  \sim \mathcal{D} }\left[ \Rade\left(\mathcal{F}(\Pi) \circ (\bd_i)_{i=1}^N \right) \right]  \leq  C_0  B\sqrt{\frac{\Pdim(\mathcal{F}(\Pi))}{N}}.
    $$
\end{proposition}

Combining \Cref{prop-rademacher,prop-P-dim}, an upper bound on the Pseudo-dimension implies an upper bound on the expected GE. %
The concept of shattering can be extended to $\gamma$-shattering, leading to the notion of the Pseudo$_\gamma$-dimension (\citealt{kearns1994efficient}), also known in the literature as the Fat-shattering dimension. This measure, which is sensitive to the parameter $\gamma\geq 0$, provides tighter bounds
on the expected GE. In our analysis of the policy class $\Pi_{(S^t)}$, we will show that the Pseudo$_\gamma$-dimension enables us to achieve a substantially tighter upper bound than what is achievable through the more traditional Pseudo-dimension approach.
It is worth highlighting that analyzing the Pseudo$_\gamma$-dimension is rare in the literature, because the more standard Pseudo-dimension is typically sufficient for bounding the expected GE (e.g., \citealt{morgenstern2015pseudo,balcan2021much,balcan2023generalization}).

\begin{definition}[Pseudo$_\gamma$-dimension]
    Given $\gamma\geq 0$,
    a dataset  $(\bd_i)_{i=1}^m\in \left([0, U]^{T+L}\right)^m$ is $\gamma$-shattered by $\mathcal{F}(\Pi) = \{f(\pi, \cdot): \pi\in \Pi\}$ with witnesses $\tau_1,\ldots, \tau_m$ if
    there exist  $\tau_1,\ldots, \tau_m \in \bbR$ such that for any $  A\subseteq [m]$, there exists $\pi \in \Pi$  such that
    $
       f(\pi, \bd_i) > \tau_i + \gamma\ \forall i\in A,
    $ and  $
      f(\pi, \bd_i) \leq \tau_i - \gamma\ \forall i\notin A
    $.
    The Pseudo$_{\gamma}$-dimension of $\mathcal{F}(\Pi) $ denoted by $\Pgamma{\gamma}(\mathcal{F}(\Pi))$ is the maximum size of a dataset that can be $\gamma$-shattered by $\mathcal{F}(\Pi)$.
\end{definition}

The Pseudo$_\gamma$-dimension quantifies not only whether a function class $\mathcal{F}(\Pi)$ can separate samples, but also whether it can do so with a margin of at least $\gamma$.
When $\gamma = 0$, the Pseudo$_\gamma$-dimension reduces to the Pseudo-dimension. We now present a useful distribution-free bound on the Rademacher complexity based on the Pseudo$_\gamma$-dimension (e.g., Theorems 10 and 17 of \citealt{mendelson2003few}). Note that in our setting, the integral in their bound can be evaluated between 0 and 1, because the covering number, under the $L_2$ norm over $\mathcal{F}(\Pi)$ that is bounded within $[-1, 1]$ and with range being no less than $1$, is equal to 1.
\begin{proposition}\label{prop-gamma-dim}
    For any distribution $\mathcal{D}\in\Delta([0,U]^{T+L})$, sample size $N\in\mathbb{Z}_{>0}$, and function class $\mathcal{F}(\Pi)=\{f(\pi,\cdot):\pi\in\Pi\}$ where $f(\pi,\bd)\in[-1, 1]$ for all $\pi\in\Pi$ and $\bd\in[0,U]^{T+L}$, there exist absolute constants $C_1, C_2>0$ such that
    $$
   \mathbb{E}_{ \bd_1, \ldots, \bd_N  \sim \mathcal{D} } \left[  \Rade\left(\mathcal{F}(\Pi) \circ (\bd_i)_{i=1}^N \right) \right] \leq \   \frac{C_1}{\sqrt{N}} \int_0^1 \sqrt{ \log\left(\frac{2}{\gamma} \right)\cdot \Pgamma{C_2\gamma}(\mathcal{F}(\Pi)) }\ d\gamma.
    $$
\end{proposition}

\section{Theoretical Results}\label{sec-analysis}
In this section, we leverage the specific structures of the loss functions induced by the policy classes $\Pi_S$, $\Pi_{(s, S)}$, and $\Pi_{(S^t)}$ to analyze their Pseudo-dimension/Pseudo$_{\gamma}$-dimension. By combining with \Cref{prop-rademacher,prop-P-dim,prop-gamma-dim}, this leads to distribution-free bounds on the expected GE of these policy classes w.r.t.\ the horizon length $T$ and sample size $N$.

In \Cref{sec-base-stock}, we analyze the class of stationary base-stock policies $\Pi_S$, and demonstrate an upper bound of $O(\sqrt{1/N})$ for the expected GE of $\cL(\Pi_S)$ in \Cref{col-basestock}. We then analyze the class of $(s, S)$ policies $\Pi_{(s, S)}$, presenting an upper bound of $O(\sqrt{\log T/N})$ and a lower bound of $\Omega(\sqrt{\log T/\left(N\log \log T\right)})$ for the expected GE of $\cL(\Pi_{(s, S)})$ in \Cref{col-ss} of \Cref{sec-SS-ub} and \Cref{col-ss-lb-2} of \Cref{sec-SS-lb}, respectively. In \Cref{sec-ST}, we analyze the class of non-stationary base-stock policies $\Pi_{(S^t)}$, and show an upper bound of $O(\sqrt{1/N})$ for the expected GE of $\cL(\Pi_{(S^t)})$ in \Cref{col-st}. Finally, in \Cref{sec-PERM}, we extend these results from solving ERM with arbitrary demand sequences to solving DP with independent demand sequences.

\subsection{The Class of Stationary Base-Stock Policies}\label{sec-base-stock}

In this section, we analyze the class of stationary base-stock policies $\Pi_{S} = \{S: S\in [0,H] \}$, where $H$ is set to be $(L+1)U$.
We note that $H$ is the maximum desirable base-stock level needed to cover the demand over $L+1$ periods.
Under such a policy, if the inventory position $I^t$ falls below $S$ in period $t$, then an order is placed to bring the inventory position up to $S$.
Given $x^1\leq 0$, we have $q^1 = S-I^1=S-x^1$, $y^t=x^1-\sum_{t'=1}^{t-1} d^{t'}$ for all $1\leq t\leq L$, and
    \begin{align*}
       I^t = x^t+\sum_{t'=t-L}^{t-1} q^{t'} =(y^{t-1}-d^{t-1})+\sum_{t'=t-L}^{t-1} q^{t'} =S- d^{t-1},\ \  q^t  &= S - I^t =  d^{t-1},\ \forall t\geq 2.
    \end{align*}
It implies that $ y^{t} = x^{t} + q^{t-L} = S - \sum_{t'=t-L}^{t-1} d^{t'}$ for all $t\geq L+1$.

From this derivation, the loss $\ell(S, \bd)$ of the $S$ policy on demand sequence $\bd$  is
    \begin{align}\label{eq-01}
       \ell(S,\bd) = \frac{1}{T}\sum_{t=L+1}^{T+L} \left( c\left(S- \sum_{t'=t-L}^{t} d^{t'}  \right)
          + K\cdot\bbI\{q^{t-L} >0\}   \right).
    \end{align}
Note that $\ell(S, \bd)$ is convex in $S$ for any given $\bd$, because $c(\cdot)$ is a convex function and the fixed-cost component is independent of $S$.
It follows that, for a given $\bd$, the set of values of $S$ that incur low loss forms an interval.

This allows us to show that the Pseudo-dimension is at most 2, following an argument similar to \citet{balcan2021much}.
They analyze the dual function class $\mathcal{L}^* (\Pi) \triangleq  \{\ell(\cdot, \bd): \bd\in [0,U]^{T+L}\}$, where each function corresponds to a demand sequence $\bd$ and is defined over the domain of policy parameters.
While the argument becomes standard once the convexity of the dual function is observed, our contribution lies in recognizing the applicability of the Pseudo-dimension concept from learning theory and formalizing this connection for the first time. This insight introduces a new analytical toolbox to the field of data-driven inventory management.

\begin{theorem}[Proved in \Cref{pf:thm-base}]\label{thm-base}
For any
$b,h\in[0,1]$,
$K\geq 0$,
$L\in\mathbb{Z}_{\ge0}$,
$T\in\mathbb{Z}_{>0}$,
and $H>0$, we have
$\Pdim\left(\cL(\Pi_{S})\right) \leq 2.$
\end{theorem}

We can now combine \Cref{prop-rademacher}, \Cref{prop-P-dim} (with $B=(L+1)U+K$), and \Cref{thm-base} to obtain the following.
\begin{corollary}\label{col-basestock}
There exists an absolute constant $C>0$ such that
for any
$b,h\in[0,1]$,
$K\geq 0$,
$L\in\mathbb{Z}_{\ge0}$,
$U> 0$,
$T\in\mathbb{Z}_{>0}$,
$\mathcal{D}\in\Delta([0,U]^{T+L})$, $N\in\mathbb{Z}_{>0}$,
assuming the class of policies $\Pi_S$ is bounded by $H=(L+1)U$,
the expected EE and GE of $\Pi_S$  have the following upper bound:
$$
\mathbb{E}_{ \bd_1, \ldots, \bd_N  \sim \mathcal{D} }\left[ \EE \left(\cL(\Pi_S) \right) \right]  \leq    \mathbb{E}_{ \bd_1, \ldots, \bd_N  \sim \mathcal{D} }\left[ \GE \left(\cL(\Pi_S) \right) \right]  \leq
C \left((L+1)U+K\right)\sqrt{\frac1N}.
$$
\end{corollary}

\begin{remark}
\label{remark-1}
Because $\Pdim(\cL (\Pi_S)) \leq 2$ essentially arises from the convexity of the loss function w.r.t.\ the parameter, this argument can also be applied to upper-bound the Pseudo-dimension of other single-parameter policy classes in related inventory models. For instance, it is applicable to both the class of base-stock policies and the class of constant-order policies for the lost-sales model with lead times. In both cases, the loss function is known to be convex w.r.t.\ the policy parameter (e.g., \citealt{janakiraman2004lost,xin2016optimality}).

Moreover, a base-stock policy with a single $S$ is a special case of an $(s, S)$ policy with $s=S$. Similar analysis applies to $(s, S)$ policies with a fixed $S-s$ gap. For example, one can set $S-s$ using the following economic order quantity (EOQ) formula with backorders: $S-s=\sqrt{\frac{2 K {\mu} (h+b)}{hb}}$ (e.g., Theorem 3.5 of \citealt{snyder2019fundamentals}), where ${\mu}$ represents the mean demand.
This enhanced single-parameter policy can significantly outperform the base-stock policy numerically, as shown in \Cref{subsec-num-N}.
\end{remark}

\subsection{The Class of ($s, S$) Policies: Upper Bound}\label{sec-SS-ub}

In this section, we analyze the class of ($s, S$) policies $\Pi_{(s, S)} = \left\{(s, S): \Hlower\leq s\leq  S \leq H, S\geq 0\right\}$. Under such a policy, an order is placed to bring the inventory position up to the order-up-to level $S$ when the inventory position falls below the reorder point $s$. That is,
\begin{align*}
        q^t &= (S - I^t) \cdot \bbI \{I^t \leq s\}.
\end{align*}

In contrast with base-stock policies, bounding the policy parameters for $(s,S)$ policies is less clear.
For example, if $h$ is small, then due to the fixed cost $K$, it may be beneficial to set $S$ to be much larger than $(L+1)U$.
For consistency with $S$ and $S^t$ policies, we aim to bound the parameters so that the maximum holding or backlogging cost incurred in any period is at most $(L+1)U$, while still allowing $U\to\infty$ as $h\to0$ (as required for optimality) or $L\to-\infty$ as $b\to0$.
Because the maximum inventory level is always $S$, we restrict $S\le H$ where $H\triangleq(L+1)U/h$ to ensure that per-period holding cost cannot exceed $(L+1)U$.
The minimum inventory level occurs when the current inventory position $I^t$ is just above $s$ (a negative number), and then $(L+1)U$ demand occurs, so we pay a backlogging cost of $(-s+(L+1)U)b$.
Thus, we restrict $s\ge\Hlower$ where $\Hlower\triangleq(L+1)U(1-1/b)\le0$ to ensure that per-period backlogging cost cannot exceed $(L+1)U$.
Finally, the restrictions of $s\le S$ and $S\ge0$ are easily seen to be without loss of optimality, resulting in the policy space $\Pi_{(s, S)} = \left\{(s, S): \Hlower\leq s\leq  S \leq H, S\geq 0\right\}$ with $\Hlower,H$ defined as above.

A key ingredient in our analysis is the use of an equivalent definition of this policy class, namely, $\Pi_{(s, S)} = \left\{(\Delta, S): 0\leq \Delta \leq H -\Hlower, 0\leq S\leq H \right\}$, where $\Delta \triangleq S-s$ denotes the minimum order quantity. To track the inventory dynamics, we define a sequence of reordering periods $(t_j)_{j\in [J]}$ with $1 \leq t_j\leq t_{j+1} \leq T$, namely, $t_j$ is in the sequence if and only if  $q^{t_j} > 0$ and $t_j\leq T$. Here, $J$ represents the total number of reordering periods during the horizon, and it depends only on $\Delta$, not on $S$. We also ignore reordering periods after period $T$ because these orders will not arrive within the horizon and thus do not affect the loss.
Note that $t_1 = 1$ because $x^1\leq \Hlower$ by assumption, and
    $$
        t_{j+1}=  \min \left\{t > t_j: \sum_{t'=t_j}^{t-1}d^{t'} \geq \Delta,\ t\leq T \right\}.
    $$
Due to the lead time $L$, the loss function is
$$
 \ell(\pi, \bd) = \frac{1}{T}\sum_{j=1}^J \left(  K+  \sum_{\hat{t} = t_j +L}^{t_{j+1}+L-1} c\left(  S-\sum_{t' = t_j}^{\hat{t}} d^{t'}  \right) \right),
$$
and we assume $t_{J+1} \triangleq T+1$.
Based on the structure of the reordering sequence, we are now able to provide an upper bound on the Pseudo-dimension of $\cL(\Pi_{(s, S)})$.

\begin{theorem}[Proved in \Cref{pf:thm-ss-ub}]\label{thm-ss-ub}
There exist absolute constants $C>0$ and $T_0\in \mathbb{Z}_{>0}$ such that
for any
$b,h\in[0,1]$,
$K\geq 0$,
$L\in\mathbb{Z}_{\ge0}$,
$T\in\mathbb{Z}_{\geq T_0}$,
$H>0 $, and $\Hlower\leq H$,
we have
$\Pdim \left(\cL (\Pi_{(s, S)}) \right) \leq C \log T$.
\end{theorem}

The proof of \Cref{thm-ss-ub} also relies on the dual-class argument over the parameter space. However, unlike the proof of \Cref{thm-base}, naively applying the argument from \cite{balcan2021much} would lead to a much looser Pseudo-dimension bound of $O(T)$.
Instead, our analysis exploits the specific structure of  $(s, S)$ policies. Specifically, for a given demand sequence $\bd$, we first observe that:
(i) the minimum order quantity $\Delta$ fully determines the reordering sequence $(t_j)_{j\in[J]}$; and
(ii) given a fixed sequence $(t_j)_{j\in[J]}$, the loss function can be re-written to reveal convexity in $S$ (assuming $s$ is adjusted to maintain a constant $\Delta$).
As a result, for each fixed $(t_j)_{j\in[J]}$, the set of $S$ values leading to low loss on $\bd$ forms a (possibly empty) interval. The remaining step is to count the number of possible reordering sequences $(t_j)_{j\in[J]}$.
A priori, there are $2^T$ possible subsequences of $1,\ldots,T$, but we show that the sequence $(t_j)_{j\in[J]}$ can change only when $\Delta$ crosses the sum of demands over some interval (i.e., when it crosses a threshold of the form $d^t+\cdots+d^{t'}$ for some $1\le t\le t'\le T$). Therefore, as $\Delta$ varies, the sequence $(t_j)_{j\in[J]}$ can change at most $O(T^2)$ times. This implies that the set of policy parameters yielding low loss on $\bd$ can be captured by $O(T^2)$ axis-aligned rectangles in the $(\Delta, S)$ plane, which in turn implies that at most $O(\log T)$ samples can be shattered.
We now combine \Cref{prop-rademacher}, \Cref{prop-P-dim} (with $B=(L+1)U+K$), and \Cref{thm-ss-ub} to obtain the following result.

\begin{corollary}\label{col-ss}
There exist absolute constants $C>0$ and  $T_0\in \mathbb{Z}_{>0}$ such that
for any
$b,h\in[0,1]$,
$K\geq 0$,
$L\in\mathbb{Z}_{\ge0}$,
$U> 0$,
$T\in\mathbb{Z}_{\geq T_0}$,
$\mathcal{D}\in\Delta([0,U]^{T+L})$,
$N\in\mathbb{Z}_{>0}$,
assuming the class of policies $\Pi_{(s, S)}$ is bounded by $H=(L+1)U/h$ and $\Hlower=(L+1)U(1-1/b)$,
the expected EE and GE of $\Pi_{(s, S)}$ have the following upper bound:
 $$
  \mathbb{E}_{ \bd_1, \ldots, \bd_N  \sim \mathcal{D} }\left[ \EE \left(\cL( \Pi_{(s,S)}) \right) \right] \leq  \mathbb{E}_{ \bd_1, \ldots, \bd_N  \sim \mathcal{D} }\left[ \GE \left(\cL( \Pi_{(s,S)}) \right) \right]  \leq   C ((L+1)U+K) \sqrt{\frac{ \log T}{N}}.
   $$
\end{corollary}

\begin{remark}\label{remark-discrete}
For this two-parameter policy class, it may be tempting to try to prove a $T$-free bound by discretizing the two-dimensional parameter space into an $M\times M$ grid, which would lead to an EE of $O(\sqrt{\log M/N})$ due to the policy class having finite cardinality $M^2$ (e.g., see Corollary 2.3 of \citealt{shalev2014understanding}). However, one can show that the discretization error remains a constant bounded away from zero, even as $M \to \infty$, due to the infinitely fine granularity required for $\Delta$, which depends on the demand distribution.
More formally, we find an absolute constant $c>0$ such that for any $M>0$, there exists a distribution $\mathcal{D}$ with
\begin{align}\label{eq-discrete}
\inf_{\pi=(s,S)\in\left\{\ldots,-\frac2M,-\frac1M,0,\frac{1}{M},\frac{2}{M},\ldots\right\}^2}R(\pi)\ - \inf_{\pi=(s,S)\in \Pi_{(s, S)}}R(\pi) \ge c.
\end{align}
Therefore, the discretization method cannot produce an EE that vanishes to 0 as $N\to\infty$.
We defer the detailed argument of \eqref{eq-discrete} to \Cref{sec-discrete} in the appendix.
\end{remark}

\subsection{The Class of ($s, S$) Policies: Lower Bound}\label{sec-SS-lb}

We now bound $\mathbb{E}_{ \bd_1, \ldots, \bd_N  \sim \mathcal{D} }\left[ \GE \left(\cL( \Pi_{(s,S)}) \right) \right]$ from below.
Note that for binary function classes, the VC-dimension provides both lower and upper bounds on the expected GE (e.g., Theorem 6.8 of \citealt{shalev2014understanding}).
By contrast, a lower bound on the Pseudo-dimension of real-valued function classes does not necessarily translate to a lower bound on the expected GE.
In fact, the Pseudo$_\gamma$-dimension can provide both lower and upper bounds on the expected GE (e.g., \citealt{kearns1994efficient,alon1997scale}), which motivates us to analyze a lower bound on the more complex Pseudo$_\gamma$-dimension of $\cL (\Pi_{(s, S)})$.

\begin{theorem}[Proved in \Cref{pf:thm-ss-lb}]\label{thm-ss-lb-2}
Suppose $U=1$, $h=K=0$, and $H=\infty$.
There exist absolute constants $C>0 $ and $ T_0\in \mathbb{Z}_{>0}$ such that
for any
$b\in (0, 1/2]$,
$L\in\mathbb{Z}_{\ge0}$,
$ T\in \mathbb{Z}_{\ge T_0}$,
$\gamma \in [0, b/16]$,
and $\Hlower\leq -1$,
we have $\Pgamma{\gamma} \left(\cL(\Pi_{(s,S)}) \right) \geq  C\log T / \log\log T$.
\end{theorem}

The proof of \Cref{thm-ss-lb-2} involves constructing an instance in which demand sequences start with $P-1$ ones, followed by a single $1/2$, and then zeros, where $P$ is a product of prime numbers. The performance of an $(s,S)$ policy depends on whether $\Delta$, which determines the order frequency, divides $P$.
The number of prime numbers that can be multiplied together without exceeding $T$ determines how many samples can be $\gamma$-shattered. We use the celebrated Prime Number Theorem and the first Chebyshev function to show that this number is $\Omega(\log T/\log\log T)$. In addition, the shattering in \Cref{thm-ss-lb-2} is achieved using $(s,S)$ policies that satisfy the boundaries $\Hlower,H$ defined above.
Note that the Pseudo$_{\gamma}$-dimension is non-increasing in $\gamma \geq 0$, which implies that
\[
   \Pgamma{\gamma}\left(\cL (\Pi_{(s, S)})\right)  \leq  \Pgamma{0}\left(\cL (\Pi_{(s, S)})\right)= \Pdim\left(\cL (\Pi_{(s, S)})\right),
\]
suggesting the tightness of the order $O(\log T)$ for P-dim$\left(\cL (\Pi_{(s, S)})\right)$.
Finally, we show that Theorem \ref{thm-ss-lb-2} can lead to a lower bound on the expected GE of $\Pi_{(s, S)}$.

\begin{corollary}[Proved in \Cref{pf:col-ss-lb}]\label{col-ss-lb-2}
Suppose $U=1$, $h=K=0$, and $H=\infty$.
There exist a distribution $\mathcal{D}\in\Delta([0,U]^{T+L})$, absolute constants $C>0$ and $T_0\in \mathbb{Z}_{>0}$ such that
for any
$b\in(0, 1/2]$,
$L\in\mathbb{Z}_{\ge0}$,
$T\in \mathbb{Z}_{\ge T_0}$,
$\gamma \in (0, b/16]$,
and  $\Hlower\leq -1$,
 the expected GE of $\Pi_{(s, S)}$ has the following lower bound:
    $$
    \mathbb{E}_{ \bd_1, \ldots, \bd_N  \sim \mathcal{D} }\left[ \GE \left(\cL( \Pi_{(s,S)}) \right) \right]
    \geq C\gamma\sqrt{\frac{ \log T}{N\log\log T}}\quad \forall N\geq\Pgamma{\gamma}\left(\cL (\Pi_{(s, S)})\right).
   $$
\end{corollary}

The proof of \Cref{col-ss-lb-2} uses the $\gamma$-shattered samples from \Cref{thm-ss-lb-2} to establish a lower bound on the expected GE.
We note that our conversion from a lower bound on Pseudo$_\gamma$-dimension to a lower bound on GE differs from that in \citet{kearns1994efficient,alon1997scale}.
In either case, this shows that any error bound for $(s,S)$ policies derived through uniform convergence must grow with $T$, in contrast to the $S$ and $(S^t)$ policies.

\subsection{The Class of Non-Stationary Base-Stock Policies}\label{sec-ST}
In this section, we analyze the class of base-stock policies with non-stationary base-stock levels $\Pi_{(S^t)} = \left\{ (S^t)_{t=1}^{T+L}: (S^t)_{t=1}^{T+L} \in [0, H]^{T+L}\right\}$, assuming no fixed costs (i.e.,~$K=0$). As explained at the start of \Cref{sec-base-stock}, we set $H=(L+1)U$. Under such a policy, the inventory position is replenished to $S^t$ in period $t$: $q^t = \max\{S^t-I^t, 0\}$.
Given $x^1\leq 0$, $q^1 = S^1-I^1=S^1-x^1$, $y^t=x^1-\sum_{t'=1}^{t-1} d^{t'}$ for all $1\leq t\leq L$, and it is straightforward to verify that
\begin{align*}
I^t &=\ \max_{\hat{t}=1,\ldots, t-1}\left\{S^{\hat{t}}- \sum_{t' = \hat{t}}^{t-1} d^{t'}  \right\},
\ \ \forall t\geq 2 ,\\
y^t &=\ \max\left\{I^{t-L}, S^{t-L}\right\}-\sum_{t' = t-L}^{t-1}d^{t'} =\ \max_{\hat{t}=1,\ldots, t-L}\left\{S^{\hat{t}}-
\sum_{t' = \hat{t}}^{t-L-1} d^{t'}  \right\} - \sum_{t' = t-L}^{t-1}d^{t'},\ \ \forall t\geq L+1.
 \end{align*}
Thus, the loss in period $t\geq L+1$ is
     \begin{align*}
         \ell^t \left((S^{t})_{t=1}^{T+L}, \bd \right) =\ c(y^t-d^t) =\
        c\left(  \max_{\hat{t}=1,\ldots, t-L}\left\{S^{\hat{t}}- \sum_{t' = \hat{t}}^{t-L-1} d^{t'}  \right\}  - \sum_{t' = t-L}^{t}d^{t'} \right).
     \end{align*}
Recall $\cL (\Pi_{(S^t)}) = \left\{\ell\left((S^t)_{t=1}^{T+L} ,\cdot\right): (S^t)_{t=1}^{T+L} \in \Pi_{(S^t)}\right\} $ is the class of loss functions induced by  $\Pi_{(S^t)}$.

Before proceeding, we note that the Pseudo-dimension of $\cL (\Pi_{(S^t)})$ is $\Omega(T)$. This can be demonstrated by constructing an instance where each demand sequence has high demand in a specific period while maintaining low demand elsewhere. These samples can be shattered because low and high losses can be incurred by adjusting the base-stock levels to either match or mismatch with the high-demand periods, respectively.
We defer the details to \Cref{sec-pseudoDoesNotWork} in the appendix.
Thus, applying Pseudo-dimension to $(S^t)$ policies leads to at best an upper bound of $O(\sqrt{T/N})$ for the expected GE.
To obtain a tighter bound of $O(\sqrt{1/N})$, we analyze the stronger notion of Pseudo$_\gamma$-dimension.

Instead of directly bounding the Pseudo$_\gamma$-dimension of the loss functions $\cL (\Pi_{(S^t)})$, we decompose the loss function by time period and apply Talagrand's contraction lemma to reduce the problem to bounding the Pseudo$_\gamma$-dimension of the inventory levels $y^t$ induced by $\Pi_{(S^t)}$. Specifically, we use $y^t \left((S^{t'})_{t'=1}^{T+L}, \bd \right)$ to denote the inventory level after replenishment but before seeing demand in period $t $ under policy $(S^{t'})_{t'=1}^{T+L}$ on a given demand sequence $\bd$. Note that $y^t \left((S^{t'})_{t'=1}^{T+L}, \bd \right) =  y^t \left((S^{t'})_{t'=1}^{t-L}, \bd \right) \in [-LU, (L+1)U] \subseteq [-(L+1)U, (L+1)U]$ for any $(S^{t'})_{t'=1}^{T+L} \in \Pi_{(S^t)}$ and $\bd\in [0,U]^{T+L}$.
We further define a class of functions representing the normalized inventory level for $t =  L+1, \ldots, T+L$:
\begin{align*}
   \mathcal{Y} ^t (\Pi_{(S^t)} )
   \triangleq & \left\{ \frac{y^t \left((S^{t'})_{t'=1}^{t-L}, \bd \right)}{(L+1)U}: (S^{t'})_{t'=1}^{t-L}\in [0,H]^{t-L}
 \right\}.
\end{align*}
Note that the function class $\mathcal{Y}^t(\Pi_{S^t})$ takes values in the interval $[-1,1]$.

\begin{theorem}[Proved in \Cref{pf:thm-st-dim}]\label{thm-st-dim}
Suppose $K=0$.
For any $b,h\in[0,1]$,
$L\in\mathbb{Z}_{\ge0}$,
$U> 0$,
$T\in\mathbb{Z}_{>0}$,
$\bd\in\bbR^{T+L}_{\ge0}$, $\gamma\in (0,1]$,
assuming the class of policies $\Pi_{(S^t)}$ is bounded by $H=(L+1)U$,
we have $\Pgamma{\gamma} \left(\mathcal{Y} ^t (\Pi_{(S^t)} ) \right) \leq  2/\gamma +1\ \forall t  =  L+1, \ldots, T+L$.
\end{theorem}

To illustrate this key result, we provide a complete proof in the special case where $L=0$ and $U=1$, noting that we also have a longer proof for the fully general case. The general proof is deferred to \Cref{pf:thm-st-dim}.

\noindent\textbf{Proof of Theorem \ref{thm-st-dim} for the special case where $L=0$ and $U=1$.}
It suffices to show that the statement holds for $t=T$. Suppose $\Pgamma{\gamma} (\mathcal{Y} ^{T} (\Pi_{(S^t)} ) ) =m $, that is, there exist demand samples $\bd_1,\ldots, \bd_m$ that are $\gamma$-shattered with witnesses $\tau_1,\ldots, \tau_m$. Note that we must have $ \tau_i \in [0, 1] $ for $ i\in [m]$, because the normalized inventory levels in this special case as defined by $\mathcal{Y} ^{T} (\Pi_{(S^t)} )$ $\mathcal{Y} ^{T} (\Pi_{(S^t)} )= \left\{y^t \left((S^{t'})_{t'=1}^{t}, \bd \right): (S^{t'})_{t'=1}^{t}\in [0,1]^{t} \right\}$ lie in $[0,1]$ for $L=0$. Applying the definition of Pseudo$_\gamma$-dimension, for any fixed $i\in [m]$, there exists an $(S_i^t)$ policy indexed by $i$ such that
\[
\begin{aligned}
   y^{T} \left((S^{t }_i) , \bd_i \right) > \tau_i + \gamma, \text{ and } y^{T} \left((S^{t }_i) , \bd_j \right) \leq \tau_j - \gamma,\ \forall j \in [m]\setminus \{i\}.
\end{aligned}
\]
Let $t_i$ denote the last reordering point that affects the inventory level $y^{T} \left((S^{t }_i) , \bd_i \right)$, namely,
$t_i \triangleq \max\left\{ t'\in [T]: S^{t'}_i - I^{t'}\left((S^{t }_i) , \bd_i \right) > 0\right\}$, where $I^{t'}\left((S^{t }_i) , \bd_i \right) $ %
denotes the inventory position at the beginning of period $t'$ before replenishment by the $(S_i^t)$ policy on the demand sequence $\bd_i$. Define $d_i [t_1, t_2] \triangleq \sum_{t=t_1}^{t_2} d_i^t$ for all $1\leq t_1\leq t_2\leq T$.
It follows that
\begin{align*}
    & y^{T} \left((S^{t }_i), \bd_i\right)   = S_i^{t_i} - d_i[t_i, T-1],\\
     &y^{T} \left((S^{t }_i), \bd_j\right)   \geq  \max \left\{S^{t_i}_i , I^{t_i}\left((S^{t }_i), \bd_j \right)\right\} - d_j[t_i, T-1]\geq S_i^{t_i} - d_j[t_i, T-1], \ \forall   j\in [m]\setminus \{i\} \nonumber.
\end{align*}
Thus, we have for all $i\in[m]$ and $j\in[m]\setminus\{i\}$,
$$
S_i^{t_i}- d_i[t_i, T-1] >  \tau_i + \gamma
\text{ and } S_i^{t_i} -  d_j[t_i, T-1] \leq  \tau_j - \gamma,
$$
from which we derive
\begin{align}
& \tau_j - \gamma + d_j[t_i, T-1] \ge \ S_i^{t_i} >\ \tau_i + \gamma + d_i[t_i, T-1]\nonumber\\
\Longrightarrow\ \ & d_j[t_i, T-1] +  \tau_j \geq\ d_i[t_i, T-1] +  \tau_i  +2 \gamma.
\label{eq-51-special}
\end{align}

Without loss of generality, we assume that $t_1\geq t_2 \geq \cdots \geq t_m$. We claim that  for all $i\in [m]$,
\begin{equation}\label{eq-82-special}
  d_i[t_i, T-1] +  \tau_i \geq 2(i-1)\gamma- 1.
\end{equation}
It is straightforward to see that the claim holds for $i=1$, because $(d_1^t)_{t=1}^{T}$ are non-negative and $\tau_1 \geq -1$. Now suppose that (\ref{eq-82-special}) holds for some $i\in [m-1]$.
It follows that
\begin{align*}
   d_{i+1}[t_{i+1}, T-1] +  \tau_{i+1} \geq\ d_{i+1}[t_{i}, T-1] +  \tau_{i+1} \geq\ d_{i}[t_{i}, T-1] +  \tau_{i} + 2\gamma \geq\ 2i\gamma - 1,
\end{align*}
where the first inequality holds because $t_{i}\geq t_{i+1}$, the second inequality holds by applying \eqref{eq-51-special} with $j=i+1$, and the third inequality holds by the induction hypothesis.
This completes the induction and we have that the claim in \eqref{eq-82-special} holds.

To upper-bound the left-hand side of~\eqref{eq-82-special}, we use the fact that $d_i[t_i, T-1] \leq 1$, which holds because $S_i^t \leq 1$ and the $(S_i^t)$ policy when executed on $\bd_i$ does not replenish again up to time $T$.
Combining this with~\eqref{eq-82-special} and substituting $i=m$ yields $2(m-1)\gamma-1 \le 1+\tau_m$. Finally, we know that $\tau_m\leq 1$, from which it follows that $m\leq 3/2\gamma + 1$, completing the proof of Theorem \ref{thm-st-dim} for this special case. (Note that we obtain the bound $m\leq 2/\gamma+1$ for the fully general case.)  \hfill $\square$
\endproof

This proof constructs, for each trajectory $i$, a policy $(S_i^t)$ that ends with high inventory under $\bd_i$, and low inventory under all other trajectories.
Using the inventory dynamics and the last reordering point $t_i$, we essentially argue that $\bd_i$ must exhibit a low total demand from $t_i$ to the end, compared to any other sequence $\bd_j$, with the difference being at least $2\gamma$.
This allows us to directly compare the demand sequences $\bd_1,\ldots,\bd_m$ without considering policies, by identifying for each $\bd_i$ a time point $t_i$ such that its cumulative demand from $t_i$ onward is low compared to that of any other sequence. Finally, given the bounded demand support, $m$ must be $O(1/\gamma)$, completing the proof.
We can now combine \Cref{prop-rademacher}, \Cref{prop-gamma-dim}, and \Cref{thm-st-dim} to obtain the following.
Note that the proof of \Cref{col-st} requires applying Talagrand's contraction lemma to convert from analyzing losses to analyzing inventory levels, and also evaluating the integral from \Cref{prop-gamma-dim}.

\begin{corollary}[Proved in \Cref{pf:col-st}]\label{col-st} Suppose $K=0$.
    There exists an absolute constant $C>0$ such that
for any
$b,h\in[0,1]$,
$L\in\mathbb{Z}_{\ge0}$,
$U> 0$,
$T\in\mathbb{Z}_{>0}$,
$\mathcal{D}\in\Delta([0,U]^{T+L})$,
$N\in\mathbb{Z}_{>0}$,
assuming the class of policies $\Pi_{(S^t)}$ is bounded by $H=(L+1)U$,
    the expected EE and GE of $\Pi_{(S^t)}$ have the following upper bound:
   $$
     \mathbb{E}_{ \bd_1, \ldots, \bd_N  \sim \mathcal{D} }\left[ \EE\left(\cL   (\Pi_{(S^t)}) \right) \right] \leq \mathbb{E}_{ \bd_1, \ldots, \bd_N  \sim \mathcal{D} }\left[ \GE\left(\cL   (\Pi_{(S^t)}) \right) \right]  \leq   C(L+1)U\sqrt{\frac{ 1}{N}} .
   $$
\end{corollary}

As highlighted in \Cref{sec-comparison}, this $T$-free bound on the expected EE for $(S^t)$ policies in the multi-period inventory problem is quite surprising compared to the existing literature, demonstrating the advantages in combining specific inventory structures with the modern learning theory framework.
Our analysis also raises an interesting question: when is the stronger notion of Pseudo$_\gamma$-dimension necessary to establish the correct sample complexity? The data-driven algorithm design literature has typically found the standard Pseudo-dimension to be sufficient (e.g., \citealt{morgenstern2015pseudo,balcan2021much, balcan2023generalization}).
In contrast, this paper identifies an inventory setting in which both the GE and EE can be bounded by $O(\sqrt{1/N})$ using the Pseudo$_\gamma$-dimension, whereas the standard Pseudo-dimension remains $\Omega(T)$ and would lead to an inferior bound of $O(\sqrt{T/N})$.

\begin{remark}\label{remark-St}
The $T$-free result in \Cref{col-st} relies on the assumption \(K = 0\), %
which can be justified by the time period being long enough that an order is placed every period; hence, the fixed cost does not need to be modeled because it is a sunk cost paid each period. This is common in practical implementations (see, e.g., \citealt{xie2025deepstock}).
On the other hand, it remains an interesting question whether one can still obtain a $T$-free bound for Pseudo$_{\gamma}$-dimension, or for the expected GE when $K>0$.
We can construct an instance showing that $\Pgamma{\gamma} \left(\mathcal{Y} ^t (\Pi_{(S^t)} ) \right)=\Omega(T)$ when $K>0$. Intuitively, the indicators in the loss of $\frac{1}{T}\sum_{t=L+1}^{T+L} K \mathbb{I}\{y^t > x^t\}$ imply that a slight change in $(S^t)$ can lead to a dramatic change in the reordering pattern and the total number of reorders.
The detailed construction can be found in \Cref{sec-pseudogamma-nonzeroK}.
Moreover, following the same argument as in \Cref{col-ss-lb-2}, we can conclude $\mathbb{E}_{ \bd_1, \ldots, \bd_N  \sim \mathcal{D} }\left[ \GE\left(\cL   (\Pi_{(S^t)}) \right) \right] =\Omega(T)$ from $\Pgamma{\gamma} \left(\mathcal{Y} ^t (\Pi_{(S^t)} ) \right)=\Omega(T)$.
\end{remark}

\subsection{Independent Demands and Product ERM}\label{sec-PERM}
We now consider the setting where the true demand distribution $\mathcal{D}$ is independent across time; that is, $\mathcal{D}$ is a product distribution $\mathcal{D}^1 \times \cdots \times \mathcal{D}^{T+L}$, where $\mathcal{D}^t$ denotes the distribution of demand in period $t$.
To fully utilize the available samples, one may construct an empirical distribution under the independence assumption and then solve the corresponding empirical DP. This approach is also known as Product Empirical Risk Minimization (PERM) in \citet{guo2021generalizing}.

Specifically, given $N$ samples $\bd_i = (d_i^t)_{t\in [T+L]}$ for $i\in[N]$, one can construct an empirical distribution for each $\mathcal{D}^t$, with $t\in[T+L]$, using the samples $\{d_i^t\}_{i\in [N]}$. Then,
$N^{T+L}$ augmented samples can be generated as ${\bd}^{\times}_{\bii} \triangleq (d_{i_1}^1, \ldots, d_{i_{T+L}}^{T+L})$ for $\bii=\left(i_1,\ldots,i_{T+L}\right)\in[N]^{T+L}$, where
the superscript $\times$ indicates that the sample is formed via the product of marginal empirical distributions.
PERM outputs a policy $\pi^{\times}$ minimizing the empirical risk over the $N^{T+L}$ samples $R^{\times}(\pi) \triangleq \frac{1}{N^{T+L}} \sum_{\bii\in [N]^{T+L}} \ell(\pi, {\bd}^{\times}_{\bii})$, i.e., $\pi^{\times} \in \arg\inf_{\pi\in \Pi} R^{\times}(\pi)$.
The corresponding EE and GE are defined as
$\EE^{\times}(\cL(\Pi)) \triangleq R(\pi^{\times}) - R(\pi^*)$ and $\GE^{\times}(\cL(\Pi)) \triangleq \sup_{\pi \in \Pi}\left\{ R( {\pi}) -  {R}^{\times}( {\pi}) \right\}$, respectively. Similar to \eqref{ineq-ee-ge},
we have
\begin{align}
  \nonumber  \mathbb{E}_{ \bd_1, \ldots, \bd_N  \sim \mathcal{D} }[\EE^{\times}(\cL(\Pi))]& =  \mathbb{E}_{ \bd_1, \ldots, \bd_N  \sim \mathcal{D} }\left[R({\pi}^{\times}) - {R}^{\times}({\pi}^{\times}) + {R}^{\times}({\pi}^{\times}) - {R}^{\times}(\pi^*) +   {R}^{\times}(\pi^*) - R(\pi^*) \right] \nonumber\\
   \label{ineq-ind-ee} &\leq  \mathbb{E}_{ \bd_1, \ldots, \bd_N  \sim \mathcal{D} }[\GE^{\times}(\cL(\Pi))],
\end{align}
where the inequality follows from the optimality of ${\pi}^{\times}$ for minimizing $R^{\times}$, and the fact that $\mathbb{E}_{ \bd_1, \ldots, \bd_N  \sim \mathcal{D} }\left[ {R}^{\times}(\pi^*)\right]=  \frac{1}{N^{T+L}} \sum_{\bii\in [N]^{T+L}}\bE_{d_{i_t}^t\sim \mathcal{D}^t\ \forall t}  [ \ell(\pi, {\bd}^{\times}_{\bii})] =\frac{1}{N^{T+L}} \sum_{\bii\in [N]^{T+L}}\bE_{{\bd}^{\times}_{\bii}\sim \mathcal{D} }  [ \ell(\pi, {\bd}^{\times}_{\bii})] =  R(\pi^*) $ for independent demands.

We now provide an important observation regarding these $N^{T+L}$ samples.
\begin{lemma}
\label{lemma-ind}
The $N^{T+L}$ samples $\left\{ {\bd}^{\times}_{\bii} \right\}_{\bii\in[N]^{T+L}}$ can be partitioned into $N^{T+L-1}$ subsets, each containing $N$ mutually disjoint samples.
\end{lemma}
Here, mutually disjoint means that each entry $\{d_i^t\}_{i\in [N], t\in [T+L]}$ from the original dataset is used exactly once. Under the independence assumption, \Cref{lemma-ind} implies that the $N$ samples within each subset are drawn IID from $\mathcal{D}$, the original distribution over sequences.

\Cref{lemma-ind} can be shown using the following construction: for each $\bj = (j_1, \ldots, j_{T+L-1})\in [N]^{T+L-1}$, the associated subset consists of $N$ samples
\[
   {\bd}^{\times}\left(i, \bj\right) \triangleq  \bd^{\times} \left((i, {i}^*_1, \ldots, {i}^*_{T+L-1}) \right) = \left(d_i^1, d_{  {i}^*_1}^2, \ldots, d_{  {i}^*_{T+L-1}}^{T+L}\right)\  \ \forall i\in [N],
\]
where ${i}^*_t =   \left( i+ j_{t}-1 \mod N\right)+1 $ for ${t\in [T+L-1]}$.
To illustrate this construction, we present an example with $L=0$, $T=3$, and $N=2$. Consider the following table of demands:
\begin{center}
\begin{tabular}{|c|c|c|c|}
\hline
$d^t_i$ & $t=1$ & $t=2$ & $t=3$ \\
\hline
$i=1$ & 1 & 3 & 5 \\ \hline
$i=2$ & 2 & 4 & 6\\ \hline
\end{tabular}
\end{center}
Then, the partitioning of the 8 samples $\left\{ {\bd}^{\times}_{\bii} \right\}_{\bii\in[N]^{T+L}}$ into 4 subsets of mutually disjoint samples is
$$
\Big\{\{135,246\},\{136,245\},\{145,236\},\{146,235\}\Big\}.
$$

We now show that the expected GE of PERM is upper-bounded by that of ERM, as follows:
\begin{align}
 \mathbb{E}_{ \bd_1, \ldots, \bd_N  \sim \mathcal{D} }[\GE^{\times}(\cL(\Pi))]
\nonumber =&\ \mathbb{E}_{ \bd_1, \ldots, \bd_N  \sim \mathcal{D} } \left[\sup_{\pi\in\Pi}\frac1{N^{T+L-1}}\sum_{\bj\in[N]^{T+L-1}} \frac1N \sum_{i=1}^N\left(R(\pi) - \ell\left(\pi, {\bd}^{\times}\left(i, \bj\right)\right)\right)\right]
\\ \nonumber\le& \ \frac1{N^{T+L-1}}\sum_{\bj\in[N]^{T+L-1}} \mathbb{E}_{ {\bd}^{\times}\left(1, \bj\right), \ldots, {\bd}^{\times}\left(N, \bj\right)  \sim \mathcal{D} } \left[\sup_{\pi\in\Pi}\frac1N \sum_{i=1}^N\left(R(\pi) - \ell\left(\pi, {\bd}^{\times}\left(i, \bj\right)\right)\right)\right]
\\ \nonumber =&\ \frac1{N^{T+L-1}}\sum_{\bj\in[N]^{T+L-1}} \mathbb{E}_{ \bd_1, \ldots, \bd_N  \sim \mathcal{D} } \left[\sup_{\pi\in\Pi}\frac1N \sum_{i=1}^N\left(R(\pi) - \ell(\pi,\bd_i)\right)\right]
\\ \label{ineq-ind}=&\ \mathbb{E}\left[\sup_{\pi\in\Pi}\frac1N \sum_{i=1}^N \left(R(\pi) - \ell(\pi,\bd_i)\right)\right] =  \mathbb{E}_{ \bd_1, \ldots, \bd_N  \sim \mathcal{D} }[\GE (\cL(\Pi))],
\end{align}
where the first equality follows from \Cref{lemma-ind} and the first inequality uses the independence assumption. Therefore, by combining \eqref{ineq-ind-ee} and \eqref{ineq-ind}, all of the results that we derived for the GE of the ERM policy under arbitrary demands (from \Cref{sec-base-stock,sec-SS-ub,sec-SS-lb,sec-ST}) can be directly applied to the PERM policy under independent demands.
In particular, under the independence assumption, our theoretical guarantees extend to PERM, which can be efficiently computed using a backward DP approach.

\section{Numerical Experiments}\label{sec-numerical}

In this section, we conduct numerical experiments to evaluate the learning performance of three policy classes under the $\Pi$-constrained ERM approach in \Cref{subsec-num-T,subsec-num-N}, and to compare the ERM and PERM approaches in \Cref{subsec-num-perm}. Throughout this section, we set $L=0$, $h=1$, and $b=9$, corresponding to a service level of $b/(h+b)=90\%$.

Unless stated otherwise, ERM policies are computed as follows.
For $\Pi_{(s,S)}$, since demands (defined below) take integer values, we compute $\hat{\pi}=(\hat{s},\hat{S})$ via a brute-force search over integer values.
For $\Pi_{(S^t)}$, we use the black-box optimizer L-BFGS-B, which achieves low optimization error, as demonstrated later.
The best-in-class policy $\pi^*$ is approximated by the ERM policy computed using a dataset of 2,000 samples. Finally, the expectation in $R(\cdot)$ is approximated by the empirical average of $\ell(\cdot,\bd_j)$ over 2,000 independently generated samples $(\bd_j)_j$.

\subsection{Varying the Horizon Length $T$}\label{subsec-num-T}

In this subsection, we empirically examine the dependence of expected EE on $T$, as shown in \Cref{col-basestock,col-ss,col-st}.
We consider a setting with independent demands, in which demands $\{d^t\}$ in period $t$ is drawn from a normal distribution $N(\mu^t, (\sigma^t)^2)$.
The parameters $\mu^t$ and $\sigma^t$ are independently drawn from uniform distributions $U(5, 15)$ and $U(2.5, 7.5)$, respectively. Demands are then truncated and rounded to nonnegative integers upper-bounded by 20.
The sample size is fixed at $N = 20$, and we vary the horizon length over $T\in\{20, 40, 60, 80, 100\}.$

Since the demands are integer-valued, the optimal policy $\pi^{\opt}$ can be computed exactly via DP, with state given by the period $t$ and the current inventory level $x^t$.
The expectation in the expected EE, $\bbe_{\bd_1,\ldots,\bd_N}\!\left[ R(\hat{\pi}) - R(\pi^*) \right]$, is approximated using 20 independent realizations of the dataset $\{\bd_1,\ldots,\bd_N\}$.
We report the expected EE relative to the optimal loss $R(\pi^{\opt})$, i.e., $\bbe_{\bd_1,\ldots,\bd_N}\!\left[ R(\hat{\pi}) - R(\pi^*) \right] / R(\pi^{\opt})$, and the out-of-sample (OOS) loss ratio  relative to $R(\pi^{\opt})$, i.e.,   $\bbe_{\bd_1,\dots, \bd_N}[R(\hat{\pi})]/ R(\pi^{\opt})$.
For each $T$, we repeat the experiment for 10 independent instances (i.e., 10 different sets of $(\mu^t, \sigma^t)_{t=1}^T$), and report averages of both metrics.

\begin{figure}[h]
\centering
\subfigure[]{
\includegraphics[scale=0.38]{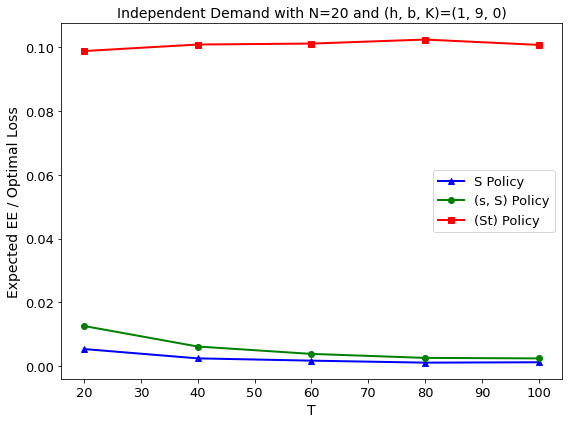}
}
\subfigure[]{
\includegraphics[scale=0.38]{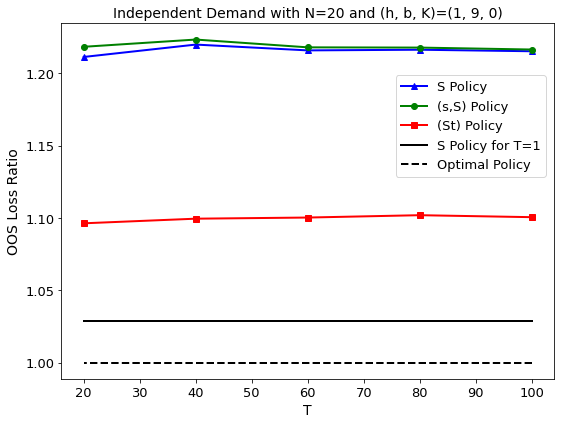}
}
\caption{The expected EE's relative to the optimal loss, and OOS loss ratios, of  $\Pi_S$, $\Pi_{(s, S)}$ and $\Pi_{(S^t)}$ policies when $K=0$. Note that for $T=1$, the OOS loss ratio of $\Pi_S$ and $\Pi_{(S^t)}$ (equivalent) is 1.029.}\label{fig-ee-reg-T-K0}
\end{figure}

We consider a setting with $K=0$, under which the optimal policy $\pi^{\opt}$ lies in  $\Pi_{(S^t)}$.
In \Cref{fig-ee-reg-T-K0}(a), we observe that the expected EE for all three policy classes %
does not increase with $T$, which is consistent with our $T$-free bounds for  $\Pi_{S}$ and $\Pi_{(S^t)}$, as proved in \Cref{col-basestock,col-st}, respectively.
For $\Pi_{(s, S)}$, since the true demands are integer-valued, the generalization error bound is also  $T$-free; see \Cref{remark-discrete}.
It also shows the expected EE is highest for $\Pi_{(S^t)}$, followed by $\Pi_{(s, S)}$, and lowest for $\Pi_{S}$, which aligns with the intuition that the more flexible policy class is more prone to overfitting.
Nonetheless, when the AE of $\Pi_{(S^t)}$ is small, the $T$-free property still ensures that its out-of-sample loss remains low as $T$ increases. Thus, if $\Pi_{(S^t)}$ performs the best for small $T$, then it remains appropriate to continue learning $(S^t)$ policies as $T$ scales, as verified in \Cref{fig-ee-reg-T-K0}(b).
We also test $T=1$, corresponding to the Newsvendor problem where $\Pi_S = \Pi_{(S^t)}$. In this setting, the OOS loss ratio of $\Pi_{(S^t)}$ is 1.029. That is, as $T$ increases from 1 to 100 (even to infinity), the regret only increases by about $8\%$, empirically demonstrating the surprisingly graceful scaling in $T$ of $\Pi_{(S^t)}$-constrained ERM policy.

\subsection{Varying the Sample Size $N$}\label{subsec-num-N}

In this subsection, we compare the numerical performance of the three policy classes w.r.t. $N$. We fix the horizon length $T$ (see values below), and vary the sample size over $N \in \{2, 4, 6, 8, 12, 16, 20\}$. We consider the OOS loss ratio, where the expectation in $\bbe_{\bd_1,\ldots,\bd_N}[R(\hat{\pi})]$ is approximated using 100 independent realizations of the dataset $\{\bd_1,\ldots,\bd_N\}$.
For each experimental setting, we repeat the experiment for 10 instances (i.e., 10 different sets of distribution parameters) and report the average OOS loss ratio.

\subsubsection{$\Pi_{(s, S)}$ vs. $\Pi_{S}$ when $K>0$}\label{subsecn-num-sS}

We first compare the policy classes $\Pi_{(s,S)}$ and $\Pi_S$  when $K>0$. We consider a setting in which demands $(d^t)_{t\in [T]}$ are IID across time and drawn from a normal distribution $N(\mu,\sigma^2)$. The parameters $\mu$ and $\sigma$ are independently drawn from uniform distributions $U(8,12)$ and $U(0.8\sigma_0,\,1.2\sigma_0)$, respectively, where $\sigma_0$ is a base standard deviation. Demands are then truncated and rounded to nonnegative integers upper-bounded by 20.
We select different values of $K$ based on the refined EOQ formula $\Delta = \sqrt{\frac{2K\mu(h+b)}{hb}}$ (see \Cref{sec-base-stock}). Letting $\Delta/\mu = P$, we set
$K = \frac{P^2 \mu h b}{2(h+b)} =4.5 P^2$
by fixing $\mu=10$, so that replenishment occurs approximately every $P$ periods.
Note that under this setting, $\Pi_{(s,S)}$ is asymptotically optimal as $T \to \infty$. We report the OOS loss ratio relative to the best-in-class policy $\pi^*_{(s,S)}$ of $\Pi_{(s,S)}$, defined as
${\bbe_{\bd_1,\ldots,\bd_N}[R(\hat{\pi})]} / {R(\pi^*_{(s,S)})}$. To mitigate large AE of $\Pi_S$, we also consider an EOQ-based variant,
defined as $\Pi_{\text{EOQ}}\triangleq \left\{(s, S):  S\in [0, H], s = S-\bar{\Delta} \right\}$, where $\bar{\Delta} \triangleq \sqrt{\frac{2K\bar{\mu} (h+b)}{hb}} $ and $\bar{\mu}$ denotes the empirical mean of the training demand samples.

\begin{figure}[h]
\centering
\subfigure[$T=20 $, $K=18$ ($P=2$), $\sigma_0=5$]{
\includegraphics[scale=0.38]{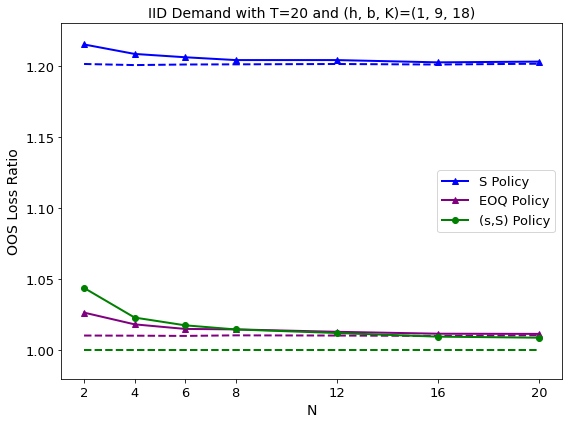}
}
\subfigure[$T$ changes from 20 to 10]{
\includegraphics[scale=0.38]{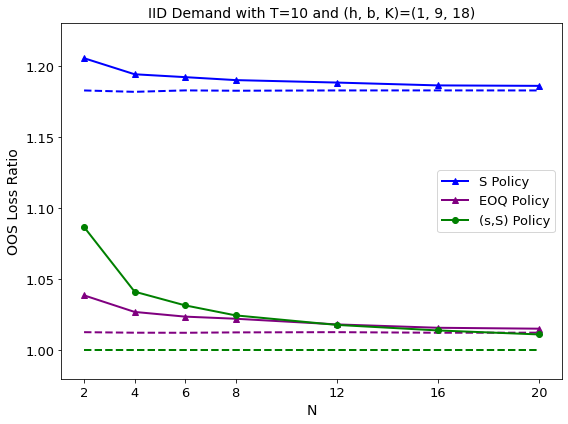}
}
\subfigure[$K$ changes from 18 to 9 ($P$ changes from 2 to $\sqrt{2}$)]{
\includegraphics[scale=0.38]{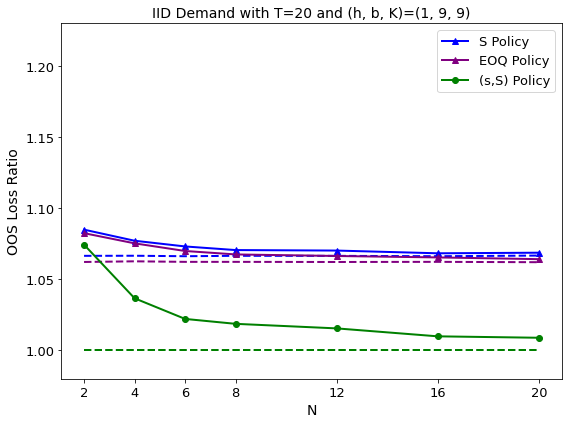}
}
\subfigure[$\sigma_0$ changes from 5 to 2]{
\includegraphics[scale=0.38]{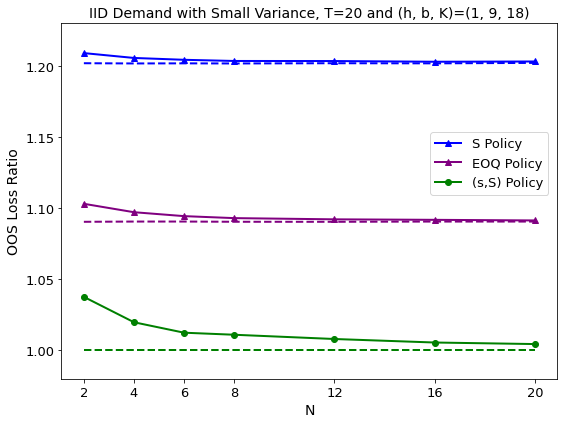}
}
\caption{The OOS loss ratios of $\Pi_{(s,S)}$ and $\Pi_{S}$ when $K>0$. Subfigures (b), (c), and (d) present the results for different values of $T$, $K$, and $\sigma_0$, respectively, compared with (a).}\label{fig-oos-sS}
\end{figure}

In \Cref{fig-oos-sS}, we present the OOS loss ratios for different values of $T$, $K$, and $\sigma_0$. In all four subfigures, $\Pi_{(s,S)}$ exhibits a larger expected EE than both $\Pi_{\text{EOQ}}$ and $\Pi_{S}$, as reflected by the larger gap between the solid and dashed curves.
Second, the OOS curves for $\Pi_{(s,S)}$ and $\Pi_{\text{EOQ}}$ intersect at a critical sample size $N^*$: $\Pi_{\text{EOQ}}$ outperforms $\Pi_{(s,S)}$ when $N \leq N^*$, whereas $\Pi_{(s,S)}$ outperforms $\Pi_{\text{EOQ}}$ when $N > N^*$.
Even under the simplest IID demands, the intersection occurs at a nontrivial sample size due to the larger EE of $\Pi_{(s,S)}$. It indicates that the simpler policy class $\Pi_{\text{EOQ}}$ can be preferable to $\Pi_{(s,S)}$ in data-limited regimes, aligning with the phenomenon that ``learning less is more.''
It has also been observed in \cite{lyu2021ucb} through numerical experiments that, in a lost-sales model with lead times, learning an optimal single-parameter $S$ policy is relatively easier, whereas learning an optimal two-parameter capped base-stock policy is more challenging but potentially more rewarding.
Third, in most scenarios, especially when $K$ is large, $\Pi_{\text{EOQ}}$ has exhibits a smaller or comparable AE relative to $\Pi_{S}$, as reflected by the gaps between their dashed lines. This suggests an advantage of using $\Pi_{\text{EOQ}}$ to reduce AE compared to $\Pi_{(s,S)}$ .

By comparing the four subfigures, we further observe how the EE and AE of each policy class vary across different scenarios.
Comparing (a) and (b), with a larger $T$,  we find that the reduction in expected EE is more significant for $\Pi_{(s,S)}$ than for $\Pi_{S}$ and $\Pi_{\text{EOQ}}$, suggesting that $\Pi_{(s,S)}$ benefits more from observing more IID demand realizations over a longer horizon.
Meanwhile, although the AE gap between $\Pi_{S}$ and $\Pi_{(s,S)}$ increases, the AE gap between $\Pi_{\text{EOQ}}$ and $\Pi_{(s,S)}$ remains almost unchanged.
Combining both EE and AE, for a larger $T$,
the intersection point occurs at a smaller sample size $N^*$.
In \Cref{fig-oos-sS}(c), when $K$ is smaller, the AE of $\Pi_{\text{EOQ}}$ increases significantly, and the intersection point shifts leftward (essentially disappearing so that $\Pi_{(s,S)}$ uniformly dominates), since $\bar{\Delta}$ becomes a poorer approximation of $\Delta = S - s$ as learned by $\Pi_{(s,S)}$.
Finally, in \Cref{fig-oos-sS}(d), lower demand variability (i.e., smaller variance) leads to a smaller EE for $\Pi_{(s,S)}$ and a smaller AE for $\Pi_{S}$. This indicates the optimal $\Delta$ is likely small, and thus $\Pi_{(s,S)}$ is less prone to overfitting. %

\begin{figure}[h]
\centering
\subfigure[$T=5,\, \mathsf{Nonst}=0.5,\, \sigma_0=5$]{
\includegraphics[scale=0.38]{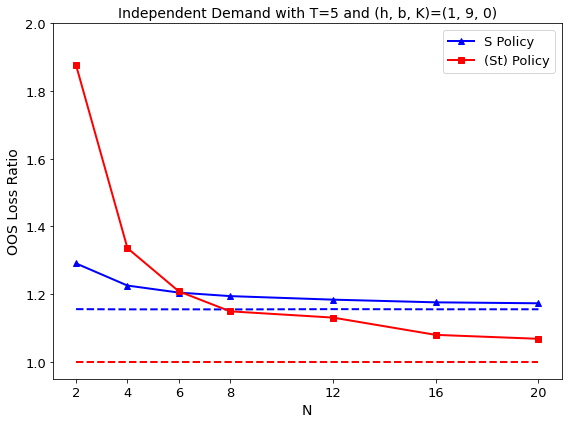}
}
\subfigure[$T$ changes from 5 to 10]{
\includegraphics[scale=0.38]{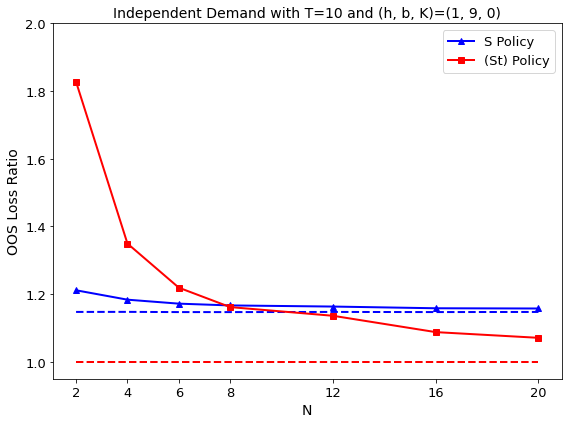}
}
\subfigure[$\mathsf{Nonst}$ changes from 0.5 to 0.2]{
\includegraphics[scale=0.38]{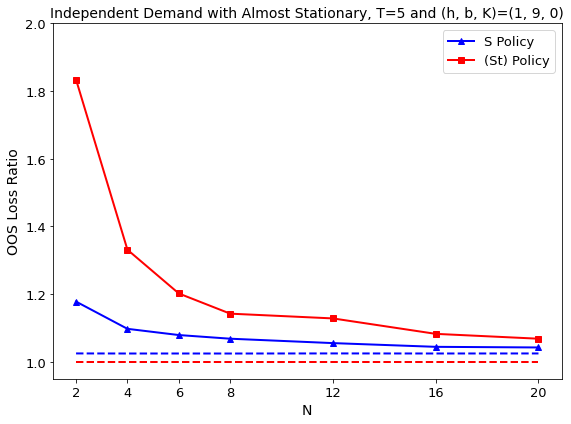}
}
\subfigure[$\sigma_0$ changes from 5 to 2]{
\includegraphics[scale=0.38]{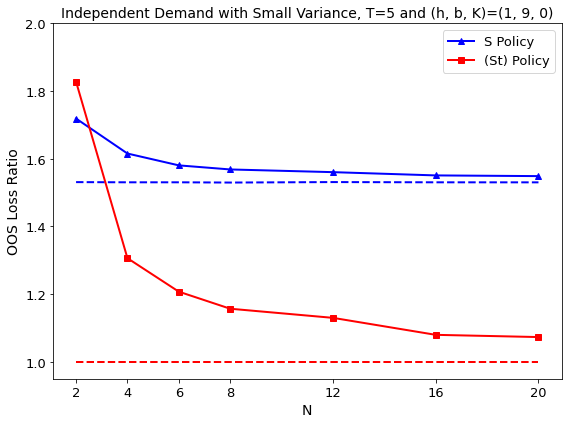}
}
\caption{The OOS loss ratios of $\Pi_{(S^t)}$ and $\Pi_{S}$ when $K=0$. Subfigures (b), (c), and (d) present the results for different values of $T$, $\mathsf{Nonst}$, and $\sigma_0$, respectively, compared with (a).}\label{fig-oos-St}
\end{figure}

\subsubsection{$\Pi_{(S^t)}$ vs. $\Pi_{S}$ when $K=0$}\label{subsec-num-St}

We now compare the policy classes $\Pi_{(S^t)}$ and $\Pi_S$  when $K=0$, and consider a setting in which demands $\{d^t\}$ in period $t$ is drawn from a normal distribution $N(\mu^t, (\sigma^t)^2)$.
The parameters $\mu^t$ and $\sigma^t$ are independently drawn from uniform distributions $U((1-\mathsf{Nonst})\cdot 10, (1+\mathsf{Nonst}) \cdot 10)$ and $U((1-\mathsf{Nonst})\cdot \sigma_0, (1+\mathsf{Nonst})\cdot \sigma_0)$, respectively, where $\sigma_0$ is a base standard deviation. The parameter $\mathsf{Nonst}$ controls nonstationarity, with smaller values corresponding to demand processes closer to stationary.
Demands are then truncated and rounded to nonnegative integers upper-bounded by 20.
We report the OOS loss ratio relative to the best-in-class policy $\pi^*_{(S^t)}$ in $\Pi_{(S^t)}$, defined as
${\bbe_{\bd_1,\ldots,\bd_N}[R(\hat{\pi})]} / {R(\pi^*_{(S^t)})}$, because the optimal policy lies in $\Pi_{(S^t)}$ in this setting.

In \Cref{fig-oos-St}, we present the OOS loss ratios for different values of $T$, $\mathsf{Nonst}$, and $\sigma_0$. Across all four subfigures, $\Pi_{(S^t)}$ exhibits a larger expected EE than $\Pi_S$, which is consistent with the results in \Cref{fig-ee-reg-T-K0}. An intersection at a sample size $N^*$ is also observed: $\Pi_S$ outperforms $\Pi_{(S^t)}$ when $N \leq N^*$, whereas $\Pi_{(S^t)}$ outperforms $\Pi_S$ when $N > N^*$.
In \Cref{fig-oos-St}(b), with a larger $T$,
the reduction in expected EE is more significant for $\Pi_{S}$ than for $\Pi_{(S^t)}$, and the AE gap between them remains unchanged, Thus, the intersection point $N^*$ occurs at a larger sample size $N^*$.
In \Cref{fig-oos-St}(c), with weaker nonstationarity (i.e., a smaller $\mathsf{Nonst}$), the AE of $\Pi_S$ is significantly reduced, and the intersection point shifts to a larger sample size (even though this is not shown in the current figure).
In \Cref{fig-oos-St}(d), with a smaller base variance $\sigma_0$,  the optimal policy $\pi^*_{(S^t)}$ becomes more concentrated around the time-varying means and $\Pi_S$ suffers from a larger AE, indicating that ${(S^t)}$ policies are preferable for most sample sizes.

\subsection{Comparing ERM to PERM}\label{subsec-num-perm}

To complement our numerical experiments for $\Pi$-constrained ERM, we now investigate the PERM approach discussed in \Cref{sec-PERM}.
Note that the policy $\pi^{\times}$ learned via the PERM approach, in which the empirical product distribution constructed from the dataset assumes independence across time, is always an $(S^t)$ policy.
Our goal is to compare PERM policy $\pi^{\times}$ with ERM policy $\hat{\pi} = (\hat{S}^t)$, with particular attention to their performance under independent demands versus temporally correlated demands. Throughout this subsection, we fix $K = 0$. We evaluate performance using the OOS ratio of ERM to PERM, defined as $
\mathbb{E}_{\mathbf{d}_1, \ldots, \mathbf{d}_N}\!\left[ R(\hat{\pi}) \right]
\big/
\mathbb{E}_{\mathbf{d}_1, \ldots, \mathbf{d}_N}\!\left[ R(\pi^{\times}) \right]. $
We conduct experiments on 10 problem instances, each with 100 independent realizations of $\{\mathbf{d}_1, \ldots, \mathbf{d}_N\}$, and report the average OOS ratio.

We begin with independent demands drawn from ${N}(\mu^t, (\sigma^t)^2)$, where $\mu^t$ and $\sigma^t$ are generated as described in \Cref{subsec-num-St}, with $\mathsf{Nonst} = 0.5$ and $\sigma_0 = 5$. In addition to the PERM policy $\pi^{\times}$ and the ERM policy $\hat{\pi}$ approximated using the black-box optimizer, we also include the exact ERM policy obtained via brute-force search. We consider a horizon length of $T = 2$ and sample sizes $N \in \{4, 8, 12, 16, 20, 24, 28, 32, 36\}.$ We also consider a longer horizon \(T = 5\); to ensure computational feasibility in this case, we scale the demand distributions to \({N}(\mu^t/5, (\sigma^t/5)^2)\) and truncate demands to have an upper bound of 4.

\begin{figure}[h]
\centering
\subfigure[$T=2$]{
\includegraphics[scale=0.38]{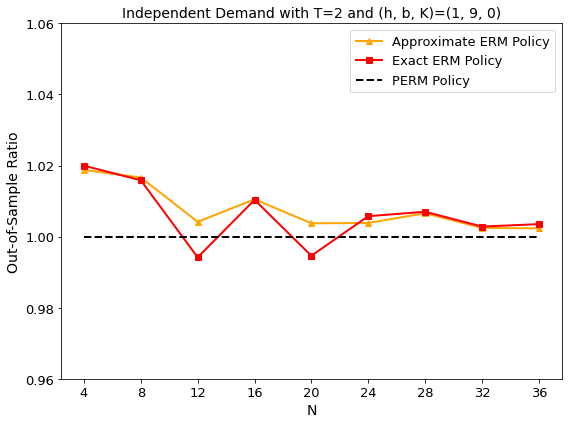}
}
\subfigure[$T=5$]{
\includegraphics[scale=0.38]{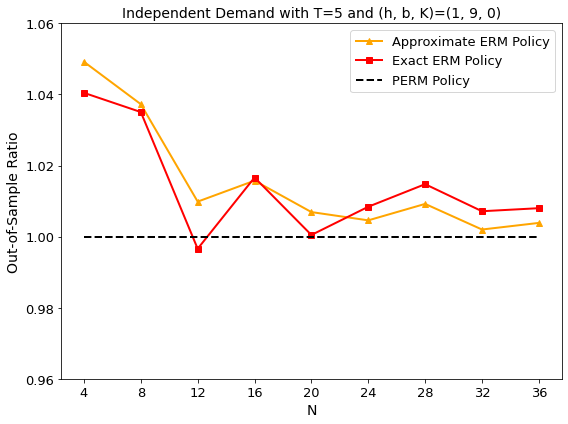}
}
\caption{The OOS ratios of ERM approaches compared to PERM approach under independent demands.}\label{fig-erm-perm-ind}
\end{figure}

As shown in \Cref{fig-erm-perm-ind}, for both $T=2$ and $T=5$, PERM outperforms ERM for all values of $N$, indicating that PERM is more sample efficient for learning $(S^t)$ policies when the true demand process is independent. This advantage arises because ERM does not exploit demand information across different trajectories.
Although the relative performance gap between ERM and PERM is substantial at small sample sizes, it decreases rapidly as $N$ increases. ERM policies exhibit strong performance and become nearly indistinguishable from PERM policies at moderate sample sizes (e.g., $N = 32$).
Comparing $T=5$ with $T=2$, the relative performance gap is larger; however, when $N \geq 12$, the OOS ratio of ERM policies falls below 1.02.
These results provide empirical evidence of the strong sample efficiency of the $\Pi$-constrained ERM approach.
In addition, the performance of the approximate ERM policy is very close to that of the exact ERM policy, with only minor fluctuations. This suggests that the ERM policy $\hat{\pi}$ obtained via the quasi-Newton method is a valid approximation of the exact ERM policy. Hence, the numerical results reported in \Cref{subsec-num-T} and \Cref{subsec-num-N} remain reliable when using the approximate ERM policy.

We next introduce correlation among the marginal distributions $N(\mu^t,(\sigma^t)^2)$ for $t\in [T]$, forming a $T$-variate normal distribution
$N(\boldsymbol{\mu}, \boldsymbol{\Sigma})$ with (pairwise) correlation coefficient $\rho$. Specifically, $\boldsymbol{\mu}=(\mu^1, \ldots, \mu^T)$ and  $(k,\ell)$-entry of $\boldsymbol{\Sigma} \in \bbR^{T\times T}$ is $\sigma^k\sigma^\ell\rho^{|k-\ell|}$.
We inherit the marginal distributions from the independent-demand experiments for both $T=2$ and $T=5$.
To induce stronger temporal dependence, we generate a finite support by drawing 5 or 20 IID samples from the joint normal distribution, and define the true demand distribution $\mathcal{D}$ to be uniform over this support. We learn the exact ERM policy $\hat{\pi}$ and the PERM policy $\pi^{\times}$ directly on the 5-sample or 20-sample support, which is equivalent to having access to an infinite training dataset drawn from $\mathcal{D}$.
We focus on the OOS ratio of the exact ERM policy compared to PERM as the correlation coefficient varies over $\rho\in \{-1, -0.8, -0.6, -0.4, -0.2,\ 0,\ 0.2,\ 0.4,\ 0.6,\ 0.8,\ 1\}$.

\begin{figure}[h]
\centering
\subfigure[$T=2$, 5-sample support]{
\includegraphics[scale=0.38]{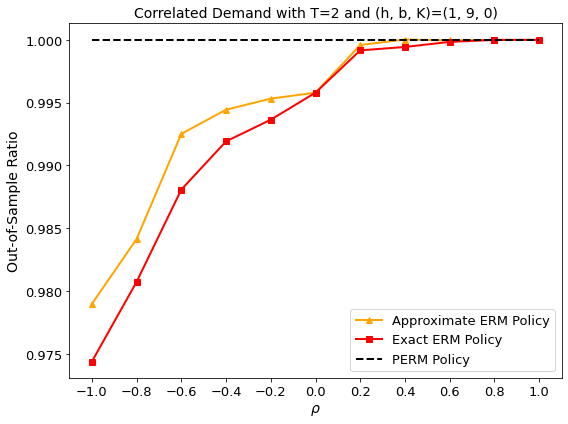}
}
\subfigure[$T=5$, 20-sample support]{
\includegraphics[scale=0.38]{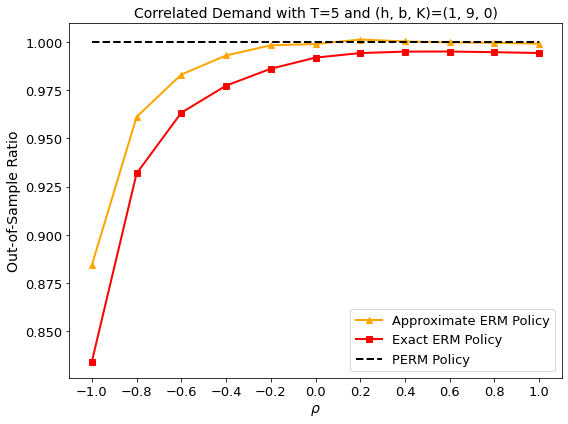}
}
\caption{The OOS ratios of ERM approaches compared to PERM approach under correlated demands.}\label{fig-erm-perm-jointnormal}
\end{figure}

As shown in \Cref{fig-erm-perm-jointnormal}, ERM outperforms PERM for most values of $\rho$. This suggests that PERM may lose temporal information when constructing a product distribution in the presence of strong dependence, whereas ERM preserves such information by directly optimizing the trajectory loss.
The figure further clarifies the regimes in which ERM is most likely to outperform PERM. In particular, ERM’s performance advantage increases as $\rho$ approaches $-1$, indicating stronger negative correlation, and also grows with the horizon length $T$.
Taken together across \Cref{fig-erm-perm-ind} and \Cref{fig-erm-perm-jointnormal}, these results suggest that when the sample size is moderate, ERM tends to be preferred over PERM---because the gain from using ERM in a correlated environment outweighs the loss from using ERM when the environment was actually (unrealistically) independent.

Overall, the independent and correlated regimes highlight a fundamental trade-off between sample efficiency and the preservation of temporal dependence when comparing PERM and ERM for learning $(S^t)$ policies.
This trade-off is central for decision-makers, who must balance concerns about data scarcity against the risks of model misspecification. A similar trade-off appears in \cite{zhang2025error} for $(s, S)$ policies, comparing SAA (interpreted as ERM) with Empirical DP (interpreted as PERM).
We note that the setting we study is entirely non-contextual. Extending this comparison to contextual settings, which are common in practice, would be both important and interesting. In such settings, $\Pi$-constrained ERM and PERM naturally correspond to the differentiable simulator approach constrained to learn base-stock policies and to model-based RL algorithms, respectively. In practical applications, the state often includes high-dimensional contextual vectors, while data collection is costly and limited. Hence, practitioners face the dual challenges of data scarcity and potential model misspecification or misidentification of the relevant contextual features driving demand. A related comparison in contextual settings is investigated in \cite{xie2025deepstock}.

\bibliographystyle{informs2014} %
\bibliography{mybib} %

@article{glasserman1995sensitivity,
  title={Sensitivity analysis for base-stock levels in multiechelon production-inventory systems},
  author={Glasserman, Paul and Tayur, Sridhar},
  journal={Management Science},
  volume={41},
  number={2},
  pages={263--281},
  year={1995},
  publisher={INFORMS}
}

@inproceedings{jiang2018open,
  title={Open problem: The dependence of sample complexity lower bounds on planning horizon},
  author={Jiang, Nan and Agarwal, Alekh},
  booktitle={Conference On Learning Theory},
  pages={3395--3398},
  year={2018},
  organization={PMLR}
}

@article{lin2022data,
  title={Data-driven newsvendor problem: Performance of the sample average approximation},
  author={Lin, Meichun and Huh, Woonghee Tim and Krishnan, Harish and Uichanco, Joline},
  journal={Operations Research},
  volume={70},
  number={4},
  pages={1996--2012},
  year={2022},
  publisher={INFORMS}
}

@article{levi2015data,
  title={The data-driven newsvendor problem: New bounds and insights},
  author={Levi, Retsef and Perakis, Georgia and Uichanco, Joline},
  journal={Operations Research},
  volume={63},
  number={6},
  pages={1294--1306},
  year={2015},
  publisher={INFORMS}
}

@article{huh2009nonparametric,
  title={A nonparametric asymptotic analysis of inventory planning with censored demand},
  author={Huh, Woonghee Tim and Rusmevichientong, Paat},
  journal={Mathematics of Operations Research},
  volume={34},
  number={1},
  pages={103--123},
  year={2009},
  publisher={INFORMS}
}

@article{besbes2023quality,
  title={From Contextual Data to Newsvendor Decisions: On the Actual Performance of Data-Driven Algorithms},
  author={Besbes, Omar and Ma, Will and Mouchtaki, Omar},
  journal={arXiv preprint arXiv:2302.08424},
  year={2023}
}

@article{guan2022randomized,
  title={Randomized Policy Optimization for Optimal Stopping},
  author={Guan, Xinyi and Mi{\v{s}}i{\'c}, Velibor V},
  journal={arXiv preprint arXiv:2203.13446},
  year={2022}
}

@article{balcan2023generalization,
  title={Generalization guarantees for multi-item profit maximization: Pricing, auctions, and randomized mechanisms},
  author={Balcan, Maria-Florina and Sandholm, Tuomas and Vitercik, Ellen},
  journal={Operations Research},
  volume={73},
  number={2},
  pages={648--663},
  year={2025}
}

@article{lykouris2021competitive,
  title={Competitive caching with machine learned advice},
  author={Lykouris, Thodoris and Vassilvitskii, Sergei},
  journal={Journal of the ACM},
  volume={68},
  number={4},
  pages={1--25},
  year={2021},
  publisher={ACM New York, NY}
}

@article{gupta2020data,
  title={Data-driven algorithm design},
  author={Gupta, Rishi and Roughgarden, Tim},
  journal={Communications of the ACM},
  volume={63},
  number={6},
  pages={87--94},
  year={2020},
  publisher={ACM New York, NY, USA}
}

@article{alon1997scale,
  title={Scale-sensitive dimensions, uniform convergence, and learnability},
  author={Alon, Noga and Ben-David, Shai and Cesa-Bianchi, Nicolo and Haussler, David},
  journal={Journal of the ACM},
  volume={44},
  number={4},
  pages={615--631},
  year={1997},
  publisher={ACM New York, NY, USA}
}

@article{balcan2020data,
  title={Data-driven algorithm design},
  author={Balcan, Maria-Florina},
  journal={arXiv preprint arXiv:2011.07177},
  year={2020}
}

@inproceedings{balcan2021much,
  title={How much data is sufficient to learn high-performing algorithms? generalization guarantees for data-driven algorithm design},
  author={Balcan, Maria-Florina and DeBlasio, Dan and Dick, Travis and Kingsford, Carl and Sandholm, Tuomas and Vitercik, Ellen},
  booktitle={Proceedings of the 53rd Annual ACM SIGACT Symposium on Theory of Computing},
  pages={919--932},
  year={2021}
}

@article{ban2019big,
  title={The big data newsvendor: Practical insights from machine learning},
  author={Ban, Gah-Yi and Rudin, Cynthia},
  journal={Operations Research},
  volume={67},
  number={1},
  pages={90--108},
  year={2019},
  publisher={INFORMS}
}

@article{ban2020confidence,
  title={Confidence intervals for data-driven inventory policies with demand censoring},
  author={Ban, Gah-Yi},
  journal={Operations Research},
  volume={68},
  number={2},
  pages={309--326},
  year={2020},
  publisher={INFORMS}
}

@article{besbes2023big,
  title={How big should your data really be? Data-driven newsvendor: learning one sample at a time},
  author={Besbes, Omar and Mouchtaki, Omar},
  journal={Management Science},
  volume={69},
  number={10},
  pages={5848--5865},
  year={2023}
}

@article{berend2013sharp,
  title={A sharp estimate of the binomial mean absolute deviation with applications},
  author={Berend, Daniel and Kontorovich, Aryeh},
  journal={Statistics \& Probability Letters},
  volume={83},
  number={4},
  pages={1254--1259},
  year={2013},
  publisher={Elsevier}
}

@article{chen2022using,
  title={Using neural networks to guide data-driven operational decisions},
  author={Chen, Ningyuan and Lagzi, Saman and Milner, Joseph},
  journal={Available at SSRN 4217092},
  year={2022}
}

@article{cheung2019sampling,
  title={Sampling-based approximation schemes for capacitated stochastic inventory control models},
  author={Cheung, Wang Chi and Simchi-Levi, David},
  journal={Mathematics of Operations Research},
  volume={44},
  number={2},
  pages={668--692},
  year={2019},
  publisher={INFORMS}
}

@article{fan2024don,
  title={Don’t Follow Reinforcement Learning Blindly: Lower Sample Complexity of Learning Optimal Inventory Control Policies with Fixed Ordering Cost},
  author={Fan, Xiaoyu and Chen, Boxiao and Olsen, Tava and Qin, Hanzhang and Zhou, Zhengyuan},
  journal={Production and Operations Management},
  pages={10591478251378851},
  year={2025},
  publisher={SAGE Publications Sage CA: Los Angeles, CA}
}

@article{gijsbrechts2022can,
  title={Can deep reinforcement learning improve inventory management? Performance on lost sales, dual-sourcing, and multi-echelon problems},
  author={Gijsbrechts, Joren and Boute, Robert N and Van Mieghem, Jan A and Zhang, Dennis J},
  journal={Manufacturing \& Service Operations Management},
  volume={24},
  number={3},
  pages={1349--1368},
  year={2022},
  publisher={INFORMS}
}

@inproceedings{guo2021generalizing,
  title={Generalizing complex hypotheses on product distributions: Auctions, prophet inequalities, and pandora’s problem},
  author={Guo, Chenghao and Huang, Zhiyi and Tang, Zhihao Gavin and Zhang, Xinzhi},
  booktitle={Conference on Learning Theory},
  pages={2248--2288},
  year={2021},
  organization={PMLR}
}

@article{han2023deep,
  title={Deep Neural Newsvendor},
  author={Han, Jinhui and Hu, Ming and Shen, Guohao},
  journal={arXiv preprint arXiv:2309.13830},
  year={2023}
}

@article{kearns1994efficient,
  title={Efficient distribution-free learning of probabilistic concepts},
  author={Kearns, Michael J and Schapire, Robert E},
  journal={Journal of Computer and System Sciences},
  volume={48},
  number={3},
  pages={464--497},
  year={1994},
  publisher={Elsevier}
}

@article{liu2023ai,
  title={Ai vs. human buyers: A study of alibaba’s inventory replenishment system},
  author={Liu, Jiaxi and Lin, Shuyi and Xin, Linwei and Zhang, Yidong},
  journal={INFORMS Journal on Applied Analytics},
  volume={53},
  number={5},
  pages={372--387},
  year={2023},
  publisher={INFORMS}
}

@article{lyu2021ucb,
  title={UCB-Type Learning Algorithms with Kaplan-Meier Estimator for Lost-Sales Inventory Models with Lead Times},
  author={Lyu, Chengyi and Zhang, Huanan and Xin, Linwei},
  Journal = {Operations Research},
  volume={72},
  number={4},
  pages={1317--1332},
  year={2024}
}

@article{madeka2022deep,
  title={Deep inventory management},
  author={Madeka, Dhruv and Torkkola, Kari and Eisenach, Carson and Luo, Anna and Foster, Dean P and Kakade, Sham M},
  journal={arXiv preprint arXiv:2210.03137},
  year={2022}
}

@incollection{mendelson2003Few,
  title = {A Few Notes on Statistical Learning Theory},
  booktitle = {Advanced Lectures on Machine Learning},
  author = {Mendelson, Shahar},
  editor = {Mendelson, Shahar and Smola, Alexander J.},
  year = {2003},
  series = {Lecture Notes in Computer Science, Volume 2600},
  pages = {1--40},
  publisher = {Springer},
  address = {Berlin, Heidelberg},
  langid = {english},
}

@book{mohri2018foundations,
  title={Foundations of Machine Learning},
  author={Mohri, Mehryar and Rostamizadeh, Afshin and Talwalkar, Ameet},
  year={2018},
  publisher={MIT Press},
  edition = {Second},
  address = {Cambridge, MA},
}

@article{morgenstern2015pseudo,
  title={On the pseudo-dimension of nearly optimal auctions},
  author={Morgenstern, Jamie H and Roughgarden, Tim},
  journal={Advances in Neural Information Processing Systems},
  volume={28},
  year={2015}
}

@article{levi2007provably,
  title={Provably near-optimal sampling-based policies for stochastic inventory control models},
  author={Levi, Retsef and Roundy, Robin O and Shmoys, David B},
  journal={Mathematics of Operations Research},
  volume={32},
  number={4},
  pages={821--839},
  year={2007},
  publisher={INFORMS}
}

@book{Pollard84,
  title={Convergence of Stochastic Processes},
  author={Pollard, David},
  year={1984},
  address={New York},
  publisher={Springer-Verlag}
}

@book{shalev2014understanding,
  title={Understanding Machine Learning: From Theory to Algorithms},
  author={Shalev-Shwartz, Shai and Ben-David, Shai},
  year={2014},
  publisher={Cambridge University Press},
  address   = "New York"
}

@book{simchi2013logic,
  title={The Logic of Logistics: Theory, Algorithms, and Applications for Logistics Management},
  author={Simchi-Levi, David and Chen, Xin and Bramel, Julien},
  address={New York},
  year={2014},
  publisher={Springer},
  Edition={Third} 
}

@article{yuan2021marrying,
  title={Marrying stochastic gradient descent with bandits: Learning algorithms for inventory systems with fixed costs},
  author={Yuan, Hao and Luo, Qi and Shi, Cong},
  journal={Management Science},
  volume={67},
  number={10},
  pages={6089--6115},
  year={2021},
  publisher={INFORMS}
}

@article{vapnik1971uniform,
  title={On the Uniform Convergence of Relative Frequencies of Events to Their Probabilities},
  author={Vapnik, VN and Chervonenkis, A Ya},
  journal={Theory of Probability \& Its Applications},
  volume={16},
  number={2},
  pages={264--280},
  year={1971},
  publisher={SIAM}
}

@article{zhang2020closing,
  title={Closing the gap: A learning algorithm for lost-sales inventory systems with lead times},
  author={Zhang, Huanan and Chao, Xiuli and Shi, Cong},
  journal={Management Science},
  volume={66},
  number={5},
  pages={1962--1980},
  year={2020},
  publisher={INFORMS}
}

@book{snyder2019fundamentals,
  title={Fundamentals of Supply Chain Theory},
  author={Snyder, Lawrence V. and Shen, Zuo-Jun Max},
  year={2019},
  edition = {Second},
  publisher={John Wiley \& Sons},
  address = {Hoboken, NJ},
}

@article{janakiraman2004lost,
  title={Lost-sales problems with stochastic lead times: Convexity results for base-stock policies},
  author={Janakiraman, Ganesh and Roundy, Robin O},
  journal={Operations Research},
  volume={52},
  number={5},
  pages={795--803},
  year={2004},
  publisher={INFORMS}
}

@article{xin2016optimality,
  title={Optimality gap of constant-order policies decays exponentially in the lead time for lost sales models},
  author={Xin, Linwei and Goldberg, David A},
  journal={Operations Research},
  volume={64},
  number={6},
  pages={1556--1565},
  year={2016},
  publisher={INFORMS}
}

@book{har2011geometric,
  title={Geometric Approximation Algorithms},
  author={Har-Peled, Sariel},
  year={2011},
  address={Providence, Rhode Island},
  publisher={American Mathematical Society}
}

@incollection{bousquet2003introduction,
  title={Introduction to statistical learning theory},
  author={Bousquet, Olivier and Boucheron, St{\'e}phane and Lugosi, G{\'a}bor},
  booktitle={Summer School on Machine Learning},
  pages={169--207},
  year={2003},
  publisher={Springer}
}

@article{qin2023sailing,
  title={Sailing through the Dark: Provably Sample-Efficient Inventory Control},
  author={Qin, Hanzhang and Simchi-Levi, David and Zhu, Ruihao},
  journal={Available at SSRN 4652347},
  year={2023}
}

@article{shalev2009stochastic,
  title={Stochastic Convex Optimization},
  author={Shalev-Shwartz, Shai and Shamir, Ohad and Srebro, Nathan and Sridharan, Karthik},
  journal={COLT},
  volume={2},
  number={4},
  year={2009}
}

@article{li2024breaking,
  title={Breaking the sample size barrier in model-based reinforcement learning with a generative model},
  author={Li, Gen and Wei, Yuting and Chi, Yuejie and Chen, Yuxin},
  journal={Operations Research},
  volume={72},
  number={1},
  pages={203--221},
  year={2024},
  publisher={INFORMS}
}

@article{dusart2010estimates,
  title={Estimates of some functions over primes without RH},
  author={Dusart, Pierre},
  journal={arXiv preprint arXiv:1002.0442},
  year={2010}
}

@article{rosser1962approximate,
  title={Approximate formulas for some functions of prime numbers},
  author={Rosser, J Barkley and Schoenfeld, Lowell},
  journal={Illinois Journal of Mathematics},
  volume={6},
  number={1},
  pages={64--94},
  year={1962},
  publisher={Duke University Press}
}

@article{gijsbrechts2025ai,
  title={AI in Inventory Management: The Disruptive Era of DRL and Beyond},
  author={Gijsbrechts, Joren and Boute, Robert N and Van Mieghem, Jan A and Zhang, Dennis},
  journal={Available at SSRN},
  year={2025}
}

@book{shapiro2021lectures,
  title={Lectures on Stochastic Programming: Modeling and Theory},
  author={Shapiro, Alexander and Dentcheva, Darinka and Ruszczy\'{n}ski, Andrzej},
  year={2009},
  publisher={SIAM},
  address={Philadelphia, PA}
}

@article{yin2021near,
  title={Near-optimal offline reinforcement learning via double variance reduction},
  author={Yin, Ming and Bai, Yu and Wang, Yu-Xiang},
  journal={Advances in Neural Information Processing Systems},
  volume={34},
  pages={7677--7688},
  year={2021}
}

@inproceedings{zhang2022horizon,
  title={Horizon-free reinforcement learning in polynomial time: the power of stationary policies},
  author={Zhang, Zihan and Ji, Xiangyang and Du, Simon},
  booktitle={Conference on Learning Theory},
  pages={3858--3904},
  year={2022},
  organization={PMLR}
}

@article{chen2024survey,
  title={Survey of data-driven newsvendor: Unified analysis and spectrum of achievable regrets},
  author={Chen, Zhuoxin and Ma, Will},
  journal={arXiv preprint arXiv:2409.03505},
  year={2024}
}

@article{agrawal2022learning,
  title={Learning in Structured MDPs with Convex Cost Functions: Improved Regret Bounds for Inventory Management},
  author={Agrawal, Shipra and Jia, Randy},
  journal={Operations Research},
  volume={70},
  number={3},
  pages={1646--1664},
  year={2022},
  doi={10.1287/opre.2022.2263}
}

@article{shi2016nonparametric,
  title={Nonparametric data-driven algorithms for multiproduct inventory systems with censored demand},
  author={Shi, Cong and Chen, Weidong and Duenyas, Izak},
  journal={Operations Research},
  volume={64},
  number={2},
  pages={362--370},
  year={2016},
  publisher={INFORMS}
}

@article{xie2025deepstock,
  title={DeepStock: Reinforcement Learning with Policy Regularizations for Inventory Management},
  author={Xie, Yaqi and Hao, Xinru and Liu, Jiaxi and Ma, Will and Xin, Linwei and Cao, Lei and Zhang, Yidong},
  journal={Available at SSRN 5784782},
  year={2025}
}

@article{zhang2025error,
  title={Error propagation in asymptotic analysis of the data-driven (s, S) inventory policy},
  author={Zhang, Xun and Ye, Zhi-Sheng and Haskell, William B},
  journal={Operations Research},
  volume={73},
  number={1},
  pages={1--21},
  year={2025},
  publisher={INFORMS}
}

@article{alvo2025deepreinforcement,
      title={Deep Reinforcement Learning for Inventory Networks: Toward Reliable Policy Optimization}, 
      author={Matias Alvo and Daniel Russo and Yash Kanoria and Minuk Lee},
      journal={arXiv preprint arXiv:2306.11246},
      year={2025} 
}

\ECSwitch
\ECDisclaimer
\ECHead{Electronic Companion for ``VC Theory for Inventory Policies''}

\section{Background on VC-Dimension and Pseudo-Dimension}\label{sec-background}

In this section, we present a variety of examples illustrating how to bound the VC-dimension and Pseudo-dimension. We maintain consistent notation with the main paper. Following conventions in learning theory,  $\bd$ is interpreted as either a training sample or a problem instance,  $\Pi $ represents a class of policies or algorithms, and $\ell(\pi, \bd)$ denotes the performance, e.g., loss (or reward), of policy $\pi$ on sample $\bd$. We focus on contexts involving a policy class  $\Pi\subseteq \bbR^k$, which is parameterized by $k$ real numbers.
Recall that the function class $\cL(\Pi) = \{\ell(\pi, \cdot): \pi\in \Pi\}$ consists of  loss functions w.r.t. sample $\bd$ across different policies $\pi$. Our goal is to bound the VC-dimension or Pseudo-dimension of $\cL(\Pi)$. Let us start with examples that directly bound $\text{VC-dim}\left(\cL(\Pi)\right)$ or $\Pdim\left(\cL(\Pi)\right)$.
We note that in machine learning textbooks (e.g.,\ \citealp{mohri2018foundations}), the VC-dimension is often introduced w.r.t. the classifier itself; however, because our focus is on decision problems, we introduce VC-dimension directly in relation to the loss functions induced by policies.

\begin{figure}[h]
    \centering
    \includegraphics[scale=0.55]{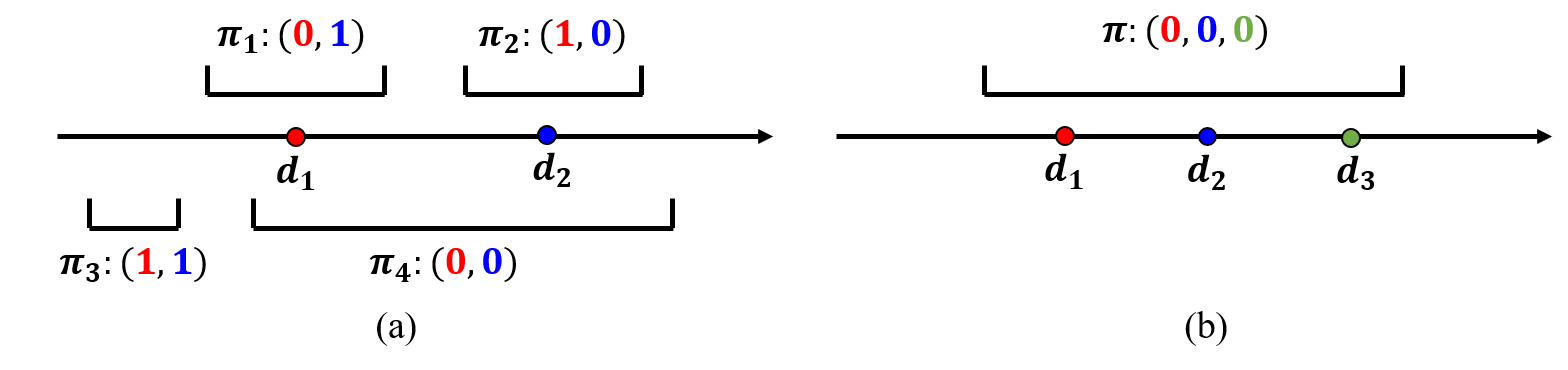}
    \caption{$\text{VC-dim}\left(\cL(\Pi)\right) = 2$ in Example \ref{eg-vcdim-1}.}
    \label{fig-eg-vcdim-1}
\end{figure}

\begin{example}\label{eg-vcdim-1}
    Consider $d\in \bbR$, $\Pi = \left\{(\underline{a}, \bar{a}): \underline{a}\leq \bar{a}\right\}\subseteq \bbR^2$, and $\ell(\pi, d) = \bbI\{ d\notin [ \underline{a}, \bar{a}] \}$ for $\pi = (\underline{a}, \bar{a})$. In this example, a policy $\pi$ is represented by a closed interval on the real line, a sample $d$ is represented by a point on the real line, and the loss function is set to zero if $\pi$ includes $d$, and one otherwise.
    As illustrated in Figure \ref{fig-eg-vcdim-1}(a), there exist two samples $d_1$ and $d_2$ with $d_1<d_2$, for which there exist four policies $\pi_1,\pi_2,\pi_3,\pi_4$ such that $\left(\bbI\{\ell(\pi, d_1)>0\},\ \bbI\{\ell(\pi, d_2)>0\}\right)$ %
    corresponds to the pairs $(0, 1), (1,0 ), (1,1), (0, 0)$, respectively. It implies that $\text{VC-dim}\left(\cL(\Pi)\right) \geq  2 $.
    In addition, as illustrated in Figure \ref{fig-eg-vcdim-1}(b), for any set of three samples $d_1, d_2, d_3$ labeled such that $d_1< d_2< d_3$, no policy $\pi$ exists such that $\left(\bbI\{\ell(\pi, d_1)>0\}, \bbI\{\ell(\pi, d_2)>0\}, \bbI\{\ell(\pi, d_3)>0\}  \right) = (0, 1, 0)$. (It is also immediately clear that if any two samples are identical, then the losses incurred on them for the same $\pi$ cannot be different.)
    Thus, $\text{VC-dim}\left(\cL(\Pi)\right) = 2$. %
\end{example}

\begin{figure}[h]
    \centering
    \includegraphics[scale=0.55]{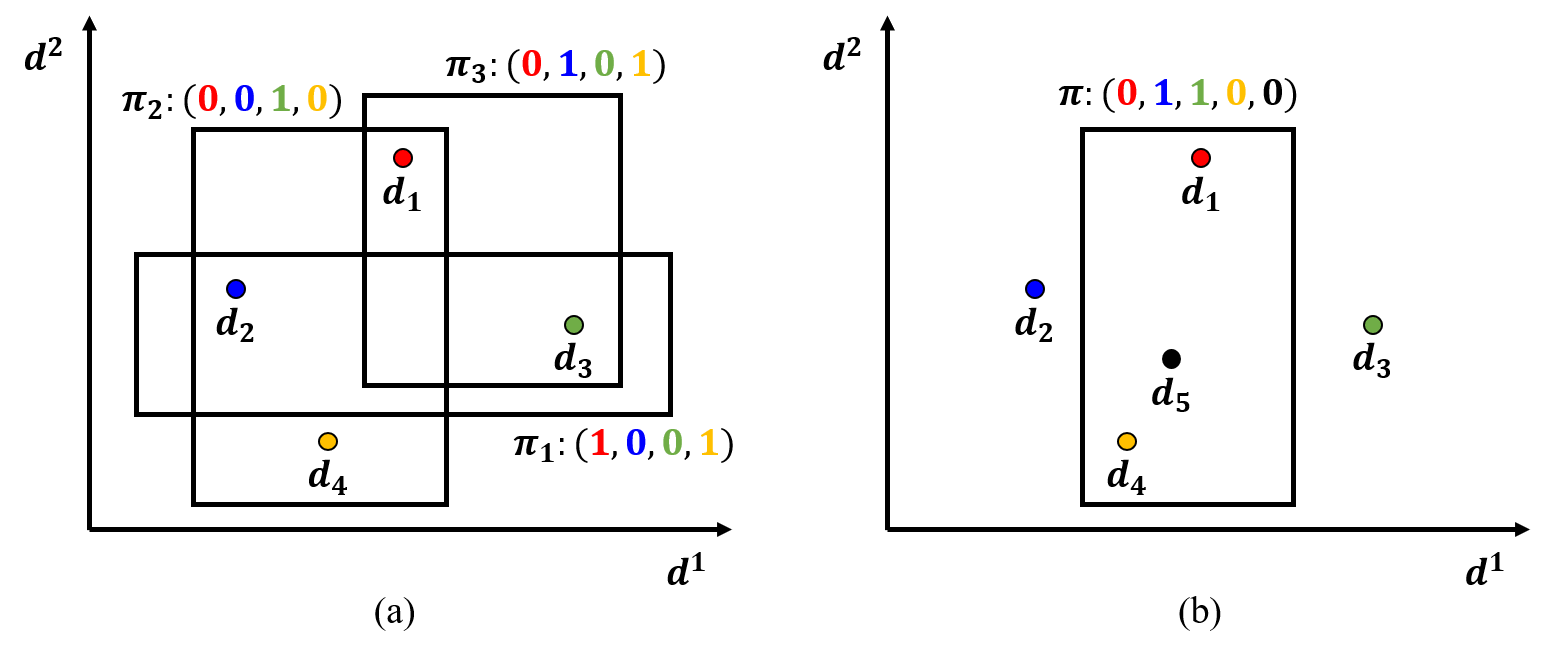}
    \caption{$\text{VC-dim}\left(\cL(\Pi)\right) = 4$ in Example \ref{eg-vcdim-2}.}
    \label{fig-eg-vcdim-2}
\end{figure}

\begin{example}\label{eg-vcdim-2}
We now extend the class of intervals as  described in Example \ref{eg-vcdim-1} to the class of axis-aligned rectangles in $\bbR^2$.
    Consider $\bd = (d^1, d^2)\in \bbR^2$, $\Pi = \left\{(\underline{a}, \bar{a}, \underline{b}, \bar{b}): \underline{a}\leq \bar{a},\ \underline{b}\leq \bar{b}\right\}\subseteq \bbR^4$ and $\ell(\pi, \bd) = \bbI\{ d^1 \notin [\underline{a}, \bar{a}] \text{ or } d^2 \notin [\underline{b}, \bar{b}] \}$ for $\pi = (\underline{a}, \bar{a}, \underline{b}, \bar{b})$.
    In this example, a policy $\pi$ is represented by an axis-aligned rectangle in $\bbR^2$, a sample $d$ is represented by a point in $\bbR^2$, and the loss function is set to zero if $\pi$ includes $d$, and one otherwise.
    As illustrated in Figure \ref{fig-eg-vcdim-2}(a), there exists a set of four samples $\bd_1, \bd_2, \bd_3, \bd_4$ that can be shattered by a set of axis-aligned rectangles, implying that $\text{VC-dim}\left(\cL(\Pi)\right) \geq  4 $.
    In addition, for any set of five samples, there must exist three samples, denoted by $\bd^*$, $\bd'$ and $\bd''$, such that $\bd^*$ is contained within the axis-aligned rectangle uniquely defined by $\bd'$ and $\bd''$ as opposite corners.
    For example, in Figure \ref{fig-eg-vcdim-2}(b), we have $\bd^* = \bd_5$, $\bd' = \bd_4$ and $\bd'' = \bd_1$. %
    It is easy to verify that there does not exist an axis-aligned rectangle $\pi$ such that $\left( \bbI\{\ell(\pi, \bd^*)>0\}, \bbI\{\ell(\pi, \bd')>0\}, \bbI\{\ell(\pi, \bd'')>0\}   \right) = (1, 0, 0)$.
    Thus, $\text{VC-dim}\left(\cL(\Pi)\right) = 4$.
\end{example}

\begin{figure}[h]
    \centering
    \includegraphics[scale=0.55]{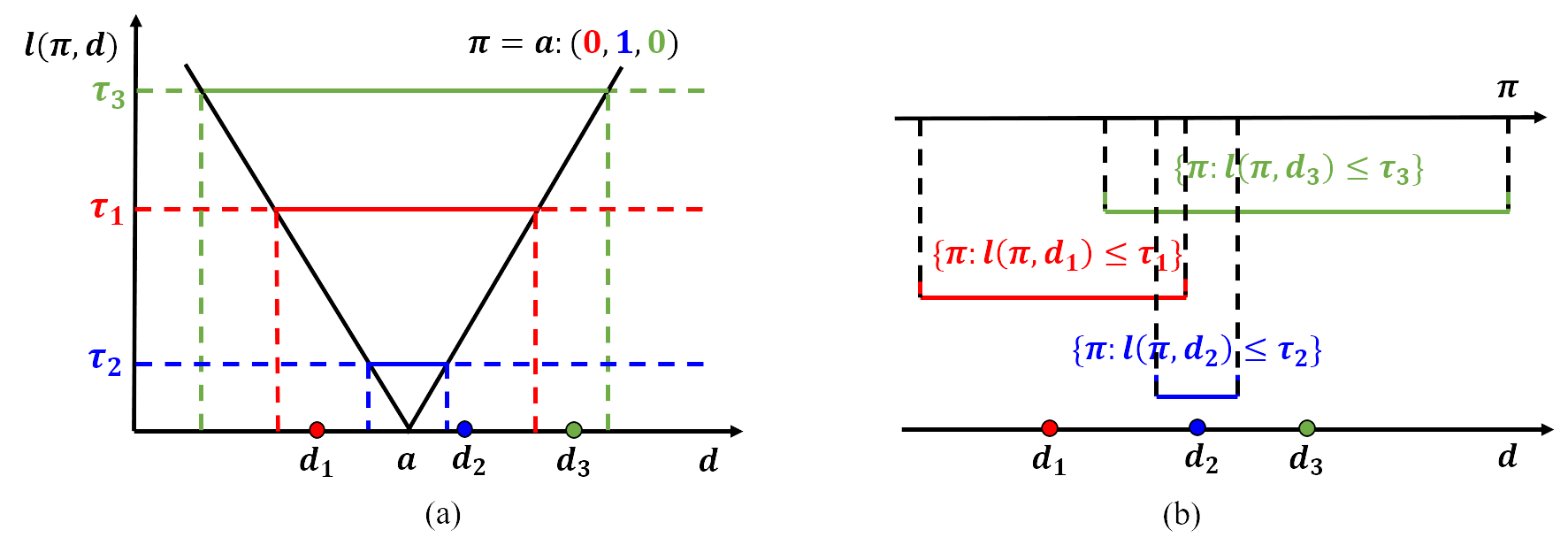}
    \caption{$\Pdim\left(\cL(\Pi)\right) = 2$ in Example \ref{eg-pdim-1}.}
    \label{fig-eg-pdim-1}
\end{figure}

\begin{example}\label{eg-pdim-1}
    The loss functions discussed in both Examples \ref{eg-vcdim-1} and \ref{eg-vcdim-2} are binary. We next discuss a class of loss functions that are real-valued and convex w.r.t. the sample $d$. Consider $d\in \bbR$, $\Pi = \{a:   a\in\bbR\}$, and with some $c_1,c_2>0$, $\ell(\pi, d) = c_1 [d-a]^+ + c_2 [a-d]^+$ for $\pi = a$.
    For each $\tau \geq 0$ and $\pi\in \Pi$, because of the convexity of $\ell(\pi, d)$ in $d$, $\left\{d: \ell(\pi, d)\leq \tau\right\}$ defines an interval in $d$, as shown in Figure \ref{fig-eg-pdim-1}(a). Similar to Example \ref{eg-vcdim-1}, it is easy to show that $\Pdim\left(\cL(\Pi)\right) \geq 2$. We next prove $\Pdim\left(\cL(\Pi)\right) \leq 2$. The argument is not as straightforward as in Example \ref{eg-vcdim-1}, as illustrated in  Figure \ref{fig-eg-pdim-1}(a). Specifically,
    for any three samples $d_1< d_2< d_3$, there exist thresholds $\tau_1, \tau_2, \tau_3$ and policy $\pi$ such that
    $\mathbf{v}\left(\pi, (d_i)_{i=1}^3, (\tau_i)_{i=1}^3\right)\triangleq \left(\bbI\{\ell(\pi, d_1)> \tau_1\},\ \bbI\{\ell(\pi, d_2)> \tau_2\},\ \bbI\{\ell(\pi, d_3)> \tau_3\} \right) = (0, 1, 0)$, due to the additional flexibility in selecting $\tau_i$. By contrast, this scenario cannot happen in the setting of Example \ref{eg-vcdim-1} without the ability to select $\tau_i$.
    However, we can alternatively show that for any $\tau_1, \tau_2, \tau_3, \pi_1, \pi_2$, it is impossible for both $\mathbf{v} \left(\pi_1, (d_i)_{i=1}^3, (\tau_i)_{i=1}^3 \right) = (0, 1, 0)$ and $\mathbf{v} \left(\pi_2, (d_i)_{i=1}^3, (\tau_i)_{i=1}^3 \right) = (1, 0, 1)$ to occur simultaneously. Thus, $\Pdim\left(\cL(\Pi)\right) \leq 2$. It is important to note that this argument is applicable not only to a broad class of general convex functions in single variables, but, due to the symmetry of shattering, it also extends to a class of general concave functions.

    The above approach to find an upper bound on the Pseudo-Dimension involves brute-forcing $\pi=a$ in $\bbR$ to calculate the values of $\mathbf{v} \left(\pi, (d_i)_{i=1}^3, (\tau_i)_{i=1}^3 \right)$. Alternatively, %
    we can first determine the ranges of $\pi$ where $\ell(\pi , d_i)\leq \tau_i$ holds for each $(d_i, \tau_i)$ pair, as illustrated in Figure \ref{fig-eg-pdim-1}(b), and then brute-force $\pi$. It is easy to see that there are at most 7 different values of $\pi$ resulting in distinct $\mathbf{v} (\pi, (d_i)_{i=1}^3, (\tau_i)_{i=1}^3 )$, which contradicts the fact that, by the definition of shattering, the number of distinct vectors $\left|\{\mathbf{v} (\pi, (d_i)_{i=1}^3, (\tau_i)_{i=1}^3 ): \pi\in \Pi\}\right|$ should be $2^3=8$.  This concludes that $\Pdim\left(\cL(\Pi)\right) \leq 2$.

\end{example}

Examples \ref{eg-vcdim-1}-\ref{eg-pdim-1} illustrate the direct approach to calculating VC-dimension and Pseudo-dimension: this involves first constructing a problem instance to establish a lower bound, followed by arguments to demonstrate the tightness of this lower bound.
This is the traditional approach, based on the definitions of VC-dimension and Pseudo-dimension. We refer readers to Section 3.3 of \cite{mohri2018foundations} for additional examples, including those involving hyperplanes in $\bbR^n$ and convex polygons in $\bbR^2$.
Note that all these examples are based on the structure of the primal loss function $\ell(\pi, \cdot)$ w.r.t. the sample $\bd$, where $\bd$ belongs to a low-dimensional Euclidean space. In general, the sample space can be quite complex, lacking obvious notions of Lipschitz continuity or smoothness to rely on, making the traditional approach challenging to implement (\citealp{balcan2021much}).
We now describe another approach for upper bounding the VC-dimension and Pseudo-dimension, which is based on the "dual" loss functions $\ell(\cdot, \bd): \Pi \rightarrow \bbR$. Recall that the dual function class $\cL^*(\Pi)$ of $\cL(\Pi)$ is defined as the set of functions $\ell(\cdot, \bd)$  indexed by the possible samples $\bd$.
This approach can be desirable if the parameter space $\Pi$ lies within a low-dimensional Euclidean space. In addition, many combinatorial problems share a clear-cut, useful structure; for example, for each sample $\bd$, the function $\ell(\cdot, \bd)$ can be piecewise structured (\citealp{balcan2020data,balcan2021much}). These advantages can make the analysis more tractable compared to a direct analysis of the primal loss function, as we have already demonstrated in the second part of Example \ref{eg-pdim-1}. Next, we provide another example that utilizes this dual approach.

\begin{figure}[h]
    \centering
    \includegraphics[scale=0.55]{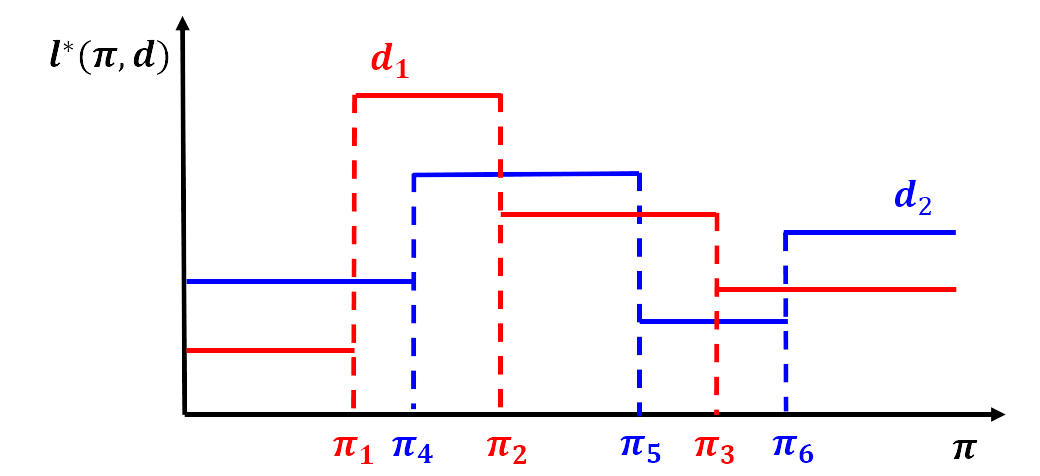}
    \caption{Example \ref{eg-pdim-2} with $C=3$.}
    \label{fig-eg-pdim-2}
\end{figure}

\begin{example}\label{eg-pdim-2}
Consider $\Pi \subseteq \bbR$ and suppose the dual loss function  $\ell(\cdot, \bd): \Pi \rightarrow \bbR$ is piecewise constant with at most  $C+1$ pieces for all $\bd$. Suppose that there exist $m$ samples $\bd_1, \ldots, \bd_m$ that are shattered by $\cL(\Pi)$ with witnesses $\tau_1, \ldots, \tau_m$. We now attempt to count the maximum number of distinct vectors $\left(\bbI\{\ell (\pi, \bd_1) > \tau_1\}, \ldots,  \bbI\{\ell (\pi, \bd_m) > \tau_m\} \right) $ that can be generated by all $\pi \in \Pi$.
Note that for any fixed $\bd$, the function $\ell(\cdot,\bd)$ can change value at most $C$ times over $\bbR$.
Thus, given $m$ such functions $\ell(\cdot,\bd_1),\ldots,\ell(\cdot,\bd_m)$, there can be at most $mC$ coordinates in $\bbR$ at which any of them changes value.  Therefore, the vector $(\ell(\pi,\bd_1),\ldots,\ell(\pi,\bd_m))$ can take at most $mC+1$ distinct values as $\pi$ ranges over $\bbR$, and hence $|(\bbI\{\ell(\pi,\bd_1)>\tau_1\},\ldots,\bbI\{\ell(\pi,\bd_m)>\tau_m\})|\le mC+1$.
We illustrate an example of this in \Cref{fig-eg-pdim-2}, where $m=2$, $C=3$, and value changes occur at $\pi_1,\ldots,\pi_6$.
In general, this implies that $2^m \leq mC+1$ due to the definition of shattering, and hence $m = O(\log C)$, i.e., $\Pdim\left(\cL(\Pi)\right) = O(\log C)$. In the example where $C=3$, we have $\Pdim\left(\cL(\Pi)\right) \leq 3$.
\end{example}

In summary, the dual approach focusing on the dual function class starts by analyzing the structure of $\ell(\cdot, \bd)$ for a given $\bd$. It then considers $m$ copies of this structure corresponding to $m$ shattered samples $\bd_1, \ldots, \bd_m$. The goal is to count the maximum number of distinct vectors $\left(\bbI\{\ell (\pi, \bd_1) > \tau_1\}, \ldots,  \bbI\{\ell(\pi, \bd_m) > \tau_m\} \right) $ that can be generated by all $\pi \in \Pi$.
We refer readers to \cite{balcan2021much} for broader applications of this approach, and to Chapter 5 of \cite{har2011geometric} for an in-depth discussion on the VC-dimension and its dual. The approach is particularly powerful in high-dimensional sample spaces. For example, in our setting, the sample $\bd$ is $T$-dimensional and we are interested in investigating $\Pdim\left(\cL(\Pi)\right)$ for large $T$.
Moreover, this approach usually requires a thorough analysis of policies or algorithms, augmented by the application of domain-specific knowledge, such as inventory theory. We adopt this approach in the proofs of \Cref{thm-base,thm-ss-ub} in this paper. Specifically, the proof of Theorem \ref{thm-base} relies on the fact that the loss function $\ell(S, \bd)$ is convex w.r.t. the base-stock level $S$ for each $\bd$.
The proof of Theorem \ref{thm-ss-ub} requires a more delicate analysis, utilizing the underlying structure of the loss function  $\ell((s, S), \bd)$ w.r.t. $(s, S)$. For example, for a given $\bd$, our analysis leads to the conclusion that the condition $\bbI\{\ell((\Delta, S), \bd)\leq \tau\}=1$ defines a polynomial number of axis-aligned rectangles in the $(\Delta, S)$ plane with $\Delta = S-s$. Fortunately, in both cases, this dual approach yields upper bounds for the Pseudo-dimension that are relatively tight, in terms of dependence on $T$.

\section{Instructive Constructions}

\subsection{Approximation Error from Discretization}\label{sec-discrete}
In this section, we present a formal proof of \eqref{eq-discrete}. Let $M$ be a positive integer. Consider the setting with $U=1$, $L=0$, cost parameters $b=h=1/2,K=0$, a sufficiently large even $T$, and a deterministic demand sequence
\[
   \bd = \left(1,\ \frac{1}{2M},\ 1,\ \frac{1}{2M},\ \ldots, 1,\ \frac{1}{2M}\right).
\]
Note that  $\Hlower = -1$ and $H=2$ in this example. It is straightforward to see that the policy $\pi^{\dagger} = (\Delta, S) = \left(1+ \frac{1}{2M},\ 1+\frac{1}{2M}\right)$ results in a replenishment at the beginning of every odd period, leading to a loss of $\frac{h}{4M}=\frac{1}{8M}$.

We now focus on $\Pi_M$, the class of discretized $(s, S)$ policies where $S$ and $\Delta$ are multiples of $\frac{1}{M}$.
For policies with $ \Delta = 0$, replenishment occurs in every period. Thus, for any $S$, the loss incurred by these policies is
\[
   \frac{1}{2}\left( \frac{1}{2}\left|S-1\right| +  \frac{1}{2}\left|S-\frac{1}{2M}\right| \right) \geq\ \frac{1 -\frac{1}{2M}}{4} =\  \Omega(1).
\]
For policies with $\frac{1}{M} \leq \Delta\leq 1$, replenishment occurs at  $t=1$ and every even period. Thus, for any $S$, the loss is at least
\[
   \frac{1}{T}\left[ \frac{1}{2} \left|S-1\right| + \frac{T-2}{2} \left( \frac{1}{2}\left|S-\frac{1}{2M}\right| +  \frac{1}{2}\left|S-\frac{1}{2M}-1\right|\right) \right] \geq\ \frac{T-2}{4T} =\  \Omega(1).
\]
For policies with $ \Delta\geq 1+ \frac{1}{M}$, each replenishment cycle contains at least three periods and the demand realizations over each cycle must have both $1$ and $\frac{1}{2M}$. Thus, one can verify that for any $S$, the loss is at least (asymptotically)
\[
\begin{aligned}
    \  & \frac{1}{3}\left(\frac{1}{2} \left|S-1\right| + \frac{1}{2} \left|S-1- \frac{1}{2M}\right| +\frac{1}{2} \left|S-2- \frac{1}{2M}\right|
   \right)\\
   =\ &  \frac{1}{3}\left( \frac{1}{2} \left|S-\frac{1}{2M}\right| + \frac{1}{2} \left|S-1- \frac{1}{2M}\right| + \frac{1}{2} \left|S-1- \frac{1}{M}\right|
   \right)   \geq\ \frac{1+ \frac{1}{2M}}{6} =\  \Omega(1).
\end{aligned}
\]
For example, when $\Delta= 1+ \frac{1}{M}$, the loss is
\[
   \frac{\lfloor\frac{T}{6}\rfloor}{2T}\left(\left|S-1\right| +  \left|S-1- \frac{1}{2M}\right| + \left|S-2- \frac{1}{2M}\right| + \left|S-\frac{1}{2M}\right| + \left|S-1- \frac{1}{2M}\right| + \left|S-1- \frac{1}{M}\right| \right)=\ \Omega(1).
\]
Therefore, for any $M$ and any discretized policy $\pi_M\in\Pi_M$, $R(\pi_M) - R(\pi^{\dagger}) %
=\Omega(1)$, which is independent of $M$. This concludes that the approximation error from discretization remains a constant even for an arbitrary large $M$.
The proof of \eqref{eq-discrete} is completed.

\subsection{Pseudo-Dimension Lower Bound for $(S^t)$ Policies} \label{sec-pseudoDoesNotWork}
In this section, we show that there exists $T$ demand sequences that can be shattered by $\Pi_{(S^t)}$, suggesting the failure of using Pseudo-dimension to bound the expected generalization error of $(S^t)$ policies. Consider the instance with $b=1$, $h=0$, $U=1$, and $L=0$. It follows that all the base-stock levels $S^t$ and demand realizations lie within the interval $[0,1]$. The $T$ demand sequences $(\bd_i)_{i=1}^T$ are defined as follows:  $d_i^t = 1$ if $t=i$ and $d_i^t = 1/2$ otherwise. For any $A\subseteq [T]$, consider the following policy $(S^t_A)_{t=1}^T$: $S^t_A = 1/2$ if $t\in A$ and $S^t_A = 1$ otherwise. It is straightforward to see that $\ell\left((S^t_A)_{t=1}^{T}, \bd_i\right) = 1/(2T)$ if $i\in A$ and $\ell\left((S^t_A)_{t=1}^{T}, \bd_i\right) = 0$ otherwise. Thus, with witnesses $\tau_i = 0$ for all $i\in [T]$, $(\bd_i)_{i=1}^T$ are shattered by the $(S^t)_{t=1}^T$ policy class, implying $\Pdim \left(\cL (\Pi_{(S^t)} ) \right) \geq  T$.

\subsection{Pseudo$_{\gamma}$-Dimension  Lower Bound for $(S^t)$ Policies When $K>0$} \label{sec-pseudogamma-nonzeroK}
In this section, we provide an instance such that there exist absolute constants $C>0$ and $T_0\in \mathbb{Z}_{>0}$ such that for any $\gamma\in [0, K/T)$, $\Pgamma{\gamma}(\cL(\Pi_{(S^t)} ))=T/5\ \forall T\geq T_0$.
Consider the instance with $h=b=0$, $K\in(0,1]$, $U=1$, $L=0$ and $T\geq 5$.
The loss function is
$
\ell((S^t), \bd) = \frac{1}{T}\sum_{t=1}^T K\mathbb{I}\{y^t> x^t\}.
$
Let $m=T-4$ and $\delta = 1/m$, and we construct
$m$ demand sequences $(\bd_i)_{i=1}^m$ as follows: $d_i^{t} = i\delta$ if $t=i$, $d_i^{t} = (m-i)\delta$ if $t=m+1$, and $d_i^{t} = 0$ otherwise.
For any $A\subseteq [m]$, consider the following policy $(S^t_A)_{t=1}^T$: $S_A^1=1$, $S^{t+1}_A = 1-(t-1)\delta$ if $t\in A$, $S_A^{m+2}=1-(m-1/3)\delta$, $S_A^{m+3}=1-(m-2/3)\delta$, $S_A^{m+4}=1-(m-1)\delta$, and $S^t_A = 0$ otherwise. It is straightforward to verify that for each $i\in A$, the sequence $\bd_i$ has exactly one reorder of quantity $\delta$ at period $t=i+1$;
for each  $i\notin A$, the sequence $\bd_i$  has three reorders, each of quantity $\delta/3$, at periods $t=m+1, m+2, m+3$. That is, $\ell((S^t), \bd) =K/T$ for each $i\in A$, $\bd_i$, and $\ell((S^t), \bd) =3K/T$ for each $i\notin A$,.Therefore, by letting $\tau_i=2K/T$, for any $\gamma\in [0, K/T)$, we have $\Pgamma{\gamma}(\cL(\Pi_{(S^t)} ))\geq m=T-4 \geq T/5\ \forall T\geq 5$.

\begin{table}[H]
    \centering
    \caption{Instances constructed in \Cref{sec-pseudoDoesNotWork} and \Cref{sec-pseudogamma-nonzeroK}.}

    \renewcommand{\arraystretch}{1.4}
    \begin{tabular}{|c|c|c|c|c|c|}
        \hline
        $d_i^t$ & $t=1$ & $t=2$ & $t=3$ & \ldots & $t=T$\\
        \hline
        $i=1$ & 1 & $\tfrac12$ & $\tfrac12$ & \ldots & $\tfrac12$ \\ \hline
        $i=2$ & $\tfrac12$ & 1 & $\tfrac12$ & \ldots & $\tfrac12$ \\ \hline
        $i=3$ & $\tfrac12$ & $\tfrac12$  & 1 & \ldots & $\tfrac12$ \\ \hline
        \vdots & \vdots & \vdots & \vdots & $\ddots$ & \vdots \\ \hline
        $i=T$ & $\tfrac12$ & $\tfrac12$ & $\tfrac12$ & \ldots & 1 \\ \hline
    \end{tabular}

    \vspace{1em} %

    \begin{tabular}{|c|c|c|c|c|c|c|c|c|c|c|}
        \hline
        $d_i^t$ & $t=1$ & $t=2$ & $t=3$ & \ldots & $t=m$ & $t=m+1$ & $t=m+2$ & \ldots & $t=m+4$ \\
        \hline
        $i=1$ & $\tfrac1m$ & 0 & 0 & \ldots & 0 & $\tfrac{m-1}{m}$ & 0 & \ldots & 0 \\ \hline
        $i=2$ & 0 & $\tfrac2m$ & 0 & \ldots & 0 & $\tfrac{m-2}{m}$ & 0 & \ldots & 0 \\ \hline
        $i=3$ & 0 & 0 & $\tfrac3m$ & \ldots & 0 & $\tfrac{m-3}{m}$ & 0 & \ldots & 0 \\ \hline
        \vdots & \vdots & \vdots & \vdots & $\ddots$ & \vdots & \vdots & \vdots & \vdots & \vdots \\ \hline
        $i=m$ & 0 & 0 & 0 & \ldots & $\frac{m}{m}$ & $\tfrac{m-m}{m}$ & 0 & \ldots & 0 \\ \hline
    \end{tabular}
\end{table}

\section{Missing Proofs}

\subsection{Proof of \Cref{thm-base}} \label{pf:thm-base}

Suppose $\Pdim\left(\cL(\Pi_{S})\right) = m$. By the equivalent definition of Pseudo-dimension, there exist $m$ demand samples $\bd_1,\ldots, \bd_m$ and corresponding witnesses $\tau_1, \ldots, \tau_m\in \bbR$ such that $|E| = 2^m$, where
\begin{align*}
 E \triangleq  \left\{\left(
\begin{array}{c}
     \bbI\{\ell(S, \bd_1) > \tau_1\}  \\
     \vdots\\
     \bbI\{\ell(S,\bd_m) > \tau_m\}
\end{array}\right): S\in \Pi_{S}
\right\} .
\end{align*}
From \eqref{eq-01}, $\ell(S,\bd) $ is convex in $S$ for any given $\bd$. Thus, for any give $\bd$ and $\tau$, the set of $S$ satisfying $\bbI\{\ell(S, \bd) \leq \tau\} = 1$ is an interval  in $\Pi_S \subseteq \bbR$. Thus, $ \bbI\{\ell(S, \bd_i) \leq \tau_i\}$ for $ i\in [m]$ generate   $m$ intervals, which divide $\bbR$ into at most $2m+1$ non-overlapping regions. Considering any one of these $2m+1$ non-overlapping regions in $\Pi_S$, we know $( \bbI\{\ell(S, \bd_1) > \tau_1\}, \ldots,  \bbI\{\ell(S, \bd_m) > \tau_m\})$  is invariant for $S$ within this region.
Thus, by searching $S$ in $\bbR$, $E$ has at most $2m+1$ different elements. To have $|E| = 2^m$, we must have $2^m \leq 2m+1$, implying $\Pdim(\cL(\Pi_{S})) = m\leq 2$. The proof of Theorem \ref{thm-base} is completed.

\subsection{Proof of \Cref{thm-ss-ub}} \label{pf:thm-ss-ub}

\begin{figure}
    \centering
    \includegraphics[scale=0.5]{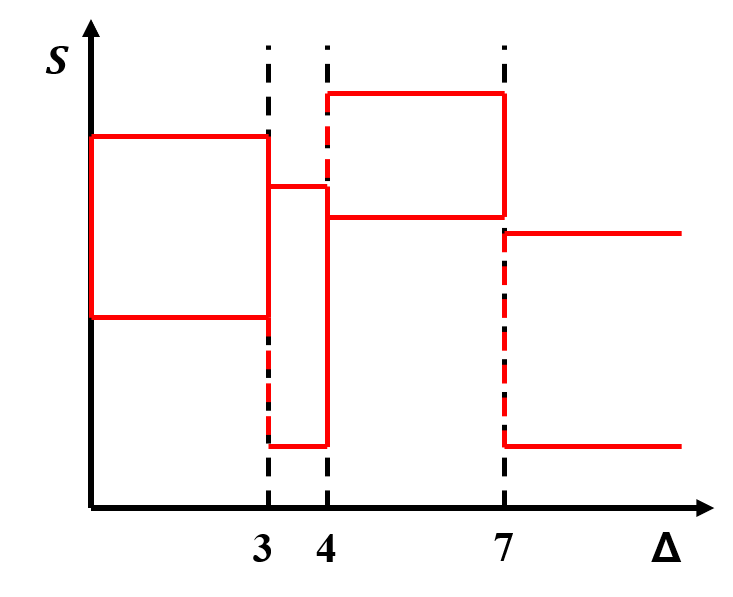}
    \caption{Consider an instance where $T = 3$ and the demand sequence is $\bd = (4, 3, 1)$. For $\Delta\in [0, 3]$, the reordering sequence is $\{1, 2, 3\}$; for $\Delta\in (3, 4]$, the reordering sequence is $\{1, 2 \}$; for $\Delta\in (4, 7]$, the reordering sequence is $\{1, 3 \}$; for $\Delta\in (7, \infty)$,  the reordering sequence is $\{1  \}$. Thus, this configuration results in a maximum of 4 non-overlapping axis-aligned rectangles.}\label{fig:proof}
\end{figure}

We first investigate the structure of the dual  class $\cL^* (\Pi_{(s, S)}) =  \{\ell((\cdot,\cdot), \bd): \bd\in [0,U]^{T+L} \}$ in the $(\Delta, S)$ space.
We claim that for given $\bd$ and $\tau \in \bbR$, the region of $(\Delta, S) $ satisfying $ \bbI\{\ell((\Delta, S), \bd) \leq \tau\}=1$ consists of at most $T(T-1)/2 + 1$ non-overlapping axis-aligned rectangles in $\bbR_+^2$. (Note that the range of $(\Delta, S)$ is relaxed to  $\bbR^2_+$.)
It is straightforward to see that  for any given $\Delta$, $\ell((\Delta,S), \bd)$ is convex in $S$, because the reordering sequence is fixed and $c(\cdot)$ is convex. It implies that for given $\Delta$, $ \bbI\{\ell((\Delta, S), \bd) \leq \tau\}=1$ occurs in an interval of $S$.
Moreover,  by the definition of $(s, S)$ policy, for any given $S$, the $(\Delta + \delta, S)$ policy  or the $(\Delta - \delta, S)$ policy  have the same reordering sequence and loss function with the  $(\Delta, S)$ policy for some small $\delta > 0$. Thus,  the region of $(\Delta, S) $ satisfying   $ \bbI\{\ell((\Delta, S), \bd) \leq \tau\}=1$  consists of multiple non-overlapping axis-aligned rectangles in $\bbR^2_+$.
Furthermore, for a given $\bd$, the number of rectangles is  no greater than the number of different reordering sequences that can be constructed with the demand sequence.
Note that a reordering sequence is determined by $\Delta$ and summations of consecutive demand values. By the definition of reordering sequence, it only depends on the first $T-1$  demand values $d^1,\ldots, d^{T-1}$. Because the number of different consecutive segments in $(d^1,\ldots, d^{T-1})$ with length $n$ is $T-n$, the number of different values by summing  consecutive demands in first $T-1$ periods is at most $\sum_{n=1}^{T-1} (T-n)= T(T-1)/2$.
As $\Delta$ increases from 0 to infinity, the current reordering sequence changes to a new one only when $\Delta$ reaches one of these $T(T-1)/2$ values. Thus, the region of $(\Delta, S) $ satisfying $ \bbI\{\ell((\Delta, S), \bd) \leq \tau\}=1$ consists of at most $T(T-1)/2 + 1$ non-overlapping axis-aligned rectangles in $ \bbR^2_+$. Among these, one rectangle is unbounded as $\Delta \rightarrow \infty$. See an illustration for $T=3$ in Figure \ref{fig:proof}.

Suppose $\Pdim\left(\cL(\Pi_{(s, S)}) \right) = m$. By  definition, there exist $m$ samples $\bd_1,\ldots, \bd_m$ and corresponding witnesses $\tau_1, \ldots, \tau_m\in \bbR$ such that $|E| = 2^m$, where
\begin{align}
\nonumber E\triangleq \left\{\left(
\begin{array}{c}
     \bbI\{\ell((\Delta,S), \bd_1) > \tau_1\}  \\
     \vdots\\
     \bbI\{\ell((\Delta,S), \bd_m) > \tau_N\}
\end{array}\right): (\Delta, S)\in \Pi_{(s, S)}
\right\} .
\end{align}
Similar to the proof of Theorem \ref{thm-base}, our goal is to count the maximum number of regions that $\bbI\{\ell((\Delta,S), \bd_i) \leq \tau_i\} = 1$ can divide a plane  of $(\Delta,S)$ into for each $ i \in [m] $.
By the above claim, the region of $ \bbI\{\ell((\Delta, S), \bd) \leq \tau \}=1$ generates at most $T(T-1)/2 $ vertical lines  and $2(T(T-1)/2+1)$ horizontal lines.
Thus, for all $ i\in [m] $, $\bbI\{\ell((\Delta,S),\bd_i) \leq \tau_i \}$ can generate at most $mT(T-1)/2 $ vertical lines  and $2m(T(T-1)/2+1)$ horizontal lines, dividing $\bbR^2_+$ into at most $(mT(T-1)/2 +1) (2m(T(T-1)/2+1) +1)=O(m^2 T^4)$ regions in total.
It is easy to verify that $(mT(T-1)/2 +1) (2m(T(T-1)/2+1) +1)\leq m^2T^4/2$ for all $T\geq 3.$ To have $|E| = 2^m$, we must have $2^m \leq   m^2T^4/2$, implying $m\leq 4\log_2T^4=16\log_2T$ for all $T\geq 3$.
The proof of Theorem \ref{thm-ss-ub} is completed. %

\subsection{Proof of \Cref{thm-ss-lb-2}} \label{pf:thm-ss-lb}

Let $\left\{p_i\right\}_{i=1}^m$ be the first $m$ prime numbers (e.g., 2, 3, 5, 7, 11, 13, 17, \ldots). Define $P_i \triangleq
\frac{1}{p_i}\prod_{j=1}^m p_j$ for each $i\in [m]$.
We claim that with $b\in (0, 1/2]$ and $h=K=0$, there exists a set of  $m$ demand sequences $\bd_1,\ldots, \bd_m$ that can be $\gamma$-shattered by $\cL (\Pi_{(s, S)})$ for $\gamma \in [0, b/16]$ if $T \geq 2\prod_{j=1}^m p_j$. In this setting, the range of policy parameters is $\Hlower\leq -1$ and $H = \infty$.
We construct the demand sequences $(\bd_i)_{i=1}^m$ as follows: for each $i\in [m] $,
\begin{align*}
    d_i^{t} = \left\{
    \begin{array}{ll}
      1,   &  \text{ if } t < P_i + L-1; \\
      1/2,   &  \text{ if } t = P_i + L; \\
       0,  &  \text{ if } t > P_i + L. \
    \end{array}
    \right.
\end{align*}
We will argue that a policy with $S-s=\Delta$ performs poorly on a sequence $\bd_i$ if and only if $\Delta$ exactly divides $P_i$, and $\Delta$ should be a multiple of $p_i$ to prevent this from occurring. In particular, for any subset $A\subseteq[m]$ that we would like to perform well on, let us consider $\Delta_A =  \prod_{i\in A} p_i $ (note that $\Delta_{\emptyset} = 1$) and $S_A =\Delta_A - 1 + L$ (i.e., $s_A = -1+ L$). It is easy to see $i\notin A$ if and only if  $\Delta_A $ divides $P_i$.

If $i\notin A$, then the ending inventory level vector $(y^t-d^t)_{t=L+1}^{T+L} = (\Delta_A-2, \Delta_A-3, \ldots,0, -1, \Delta_A-2, \Delta_A-3, \ldots,0, -1, \ldots, \Delta_A-2, \Delta_A-3, \ldots,0, -1/2, -1/2, \ldots, -1/2)$, namely, consisting of exactly $\left(\frac{P_i}{\Delta_A} -1\right)$ cycles $ (\Delta_A-2, \Delta_A-3, \ldots,0, -1)$ and the $\left(\frac{P_i}{\Delta_A} \right)$-th cycle ends with $-\frac{1}{2} $, leading to $y^t-d^t = -\frac{1}{2}$ for all $t\geq P_i+L$. Thus, the loss is $ \frac{1}{T} \left(b\left(\frac{P_i}{\Delta_A}-1\right)  + \frac{b}{2}\left(T-P_i+1\right)\right)$.
If $i \in A$, then the vector $(y^t-d^t)_{t=L+1}^{T+L} = \left(\Delta_A-2, \Delta_A-3, \ldots,0, -1, \Delta_A-2, \Delta_A-3, \ldots,0, -1, \ldots, \Delta_A-2, \Delta_A-3, \ldots,r_A+1, r_A, r_A,  \ldots, r_A\right)$, namely, consisting of $\lfloor \frac{P_i}{\Delta_A} \rfloor$ cycles $ (\Delta_A-2, \Delta_A-3, \ldots, 0, -1)$ and the $\lceil \frac{P_i}{\Delta_A}\rceil$-th cycle ends with $r_A =\Delta_A - 1 -  \left(P_i- {\Delta_A} \lfloor  \frac{P_i}{\Delta_A}  \rfloor\right) +\frac{1}{2} \geq 0$, leading to $y^t-d^t = r_A$ for all $t\geq P_i+L$. Thus, the loss is no greater than $\frac{b}{T}  \frac{P_i}{\Delta_A} $.
In summary, the loss under policy $(\Delta_A, S_A)$ on demand $\bd_i$ is
\begin{align*}
    \ell((\Delta_A, S_A), \bd_i) & =\ \frac{b}{T} \left(\frac{P_i}{\Delta_A}-1  + \frac{T-P_i+1}{2}\right) \geq\  \frac{b}{T} \left( \frac{T-P_i +1}{2}\right) ,\ \ \ \text{ if } i \notin A,\\
    \ell((\Delta_A, S_A), \bd_i)& <\ \frac{b}{T} \frac{P_i}{\Delta_A}   \leq\  \frac{b}{T} \frac{P_i}{p_i} ,\ \ \ \text{ if } i \in A.
\end{align*}
We then set the witnesses correspondingly: $\tau_i = \frac{b}{2T} \left( \frac{T-P_i+1}{2}+ \frac{P_i}{p_i} \right) $. Because $T =  2\prod_{j=1}^m p_j = 4P_1\geq 4P_i$, we have
$\frac{b}{T} \left( \frac{T-P_i+1}{2}- \frac{P_i}{p_i} \right) > \frac{b}{T} \left(  \frac{T-P_i}{2}-  P_i  \right) \geq b/8$.
Thus, it is straightforward to verify that for any $\gamma \in [0, b/16]$,
\begin{align*}
    \ell((\Delta_A,S_A), \bd_i) > \tau_i+\gamma,\   \text{ if } i \notin A,\text{ and }
    \ell((\Delta_A,S_A), \bd_i) \leq \tau_i-\gamma,\   \text{ if } i \in A.
\end{align*}
That is, the set of demand sequences $\bd_1, \ldots, \bd_m$ can be $\gamma$-shattered by $\cL (\Pi_{(s, S)})$.

This construction is valid when $T\ge2\prod_{j=1}^m p_j$, or equivalently, $\log(T/2)\ge\sum_{j=1}^m\log p_j$, which connects to the first Chebyshev function $\theta(x)\triangleq \sum\limits_{p\leq x} \log p$, where the sum is taken over all prime numbers $p$ less than or equal to $x$. From Proposition 5.1 of \cite{dusart2010estimates}, we obtain the following non-asymptotic bound:
\[
   \sum_{j=1}^m\log p_j = \sum_{j:p_j\leq p_m}\log p_j \leq\ 1.000028 p_m \quad \text{for all $m\geq 1$}.
\]
In addition, the following non-asymptotic version of the celebrated Prime Number Theorem provides an upper bound on the $m$-th prime number $p_m$ (Theorem 3 of \citealp{rosser1962approximate}):
\[
   p_m<\ m(\log m+\log \log m) \quad  \text{for all $m\geq 6$}.
\]
Therefore, we conclude that $\sum_{j=1}^m\log p_j\le 3m \log m$ for all $m\ge 6$.
For this construction to be valid, it suffices to have $\log(T/2)\ge 3m\log m$ (assuming $m\ge 6$).
It follows that for all $T$ satisfying $\log T /(\log \log T)\geq 18$, we can set %
\begin{equation}\label{eq-101}
\begin{aligned}
   m=\lfloor\log T /(3\log \log T)\rfloor\geq& \log T /(3\log \log T)-1
   \geq  \left(\frac{1}{3}-\frac{1}{18}\right)\log T /\log \log T= \frac{5}{18} \log T /\log \log T,
\end{aligned}
\end{equation}
so that $\Pi_{(s, S)}$ can $\gamma$-shatter $m$ samples, establishing the statement of \Cref{thm-ss-lb-2}.

\subsection{Proof of \Cref{col-ss-lb-2}} \label{pf:col-ss-lb}

Suppose $\hat{\bd}_1,\ldots,\hat{\bd}_m$ are $\gamma$-shattered by $\cL(\Pi_{(s, S)})$ for a constant $\gamma>0$, with witnesses $\tau_1,\ldots,\tau_m$, where $m = \Pgamma{\gamma}\left(\cL (\Pi_{(s, S)})\right)$.
Let $\mathcal{D}$ be the uniform distribution over $\hat{\bd}_1,\ldots,\hat{\bd}_m$, implying that  $R(\pi)=\frac1m\sum_{i=1}^m\ell(\pi,\hat{\bd}_i)$ for all $\pi$.
For the dataset $\bd_1, \ldots, \bd_N$, let $M_i$ denote the number of times that $\hat{\bd}_i$ is drawn for each $ i\in [m]$, where $\sum_{i=1}^m M_i=N$. We have
\begin{align*}
 &\ \mathbb{E}_{ \bd_1, \ldots, \bd_N  \sim \mathcal{D} }\left[ \GE \left(\cL( \Pi_{(s,S)}) \right) \right] \\
 = &\ \mathbb{E}_{\bd_1,\ldots, \bd_N\sim \mathcal{D}} \left[\sup_{\pi\in\Pi_{(s, S)}}\left\{R(\pi)-\hat{R}(\pi)\right\} \right]\\
=&\ \mathbb{E}_{\bd_1,\ldots, \bd_N\sim \mathcal{D}} \left[\sup_{\pi\in\Pi_{(s, S)}} \left\{\frac1m\sum_{i=1}^m\ell(\pi,\hat{\bd}_i)-\frac1N\sum_{i=1}^mM_i\ell(\pi,\hat{\bd}_i)\right\} \right]
\\ =&\ \mathbb{E}_{\bd_1,\ldots, \bd_N\sim \mathcal{D}} \left[\sup_{\pi\in\Pi_{(s, S)}}\left\{\sum_{i=1}^m\frac1m\left(\ell(\pi,\hat{\bd}_i)-\tau_i \right)-\sum_{i=1}^m\frac{M_i}N \left(\ell(\pi,\hat{\bd}_i)-\tau_i\right)\right\} \right]
\\ =&\ \mathbb{E}_{\bd_1,\ldots, \bd_N\sim \mathcal{D}} \left[\sup_{\pi\in\Pi_{(s, S)}}\sum_{i=1}^m\left(\frac1m-\frac{M_i}N\right)\left(\ell(\pi,\hat{\bd}_i)-\tau_i \right)\right]
\\ \ge&\ \mathbb{E}_{\bd_1,\ldots, \bd_N\sim \mathcal{D}} \left[\sum_{i=1}^m \left|\frac1m-\frac{M_i}N\right|\gamma \right]
\\ =&\ \frac{m\gamma}{N}  \mathbb{E}\left[\left|\mathrm{Bin}\left(N,\frac{1}{m}\right)-\frac{N}{m}\right|\right].
\end{align*}
Here, $\mathrm{Bin}\left(N,1/m\right)$ denotes a random variable that follows Binomial distribution with parameters $N$ and $1/m$;
the third equality holds because $\mathbb{E}_{\bd_1,\ldots, \bd_N\sim \mathcal{D}}[M_i/N]=1/m$;  the last equality follows from the uniform distribution over these sample $\hat{\bd}_i, \forall i\in [m]$; the inequality holds because of the definition of $\gamma$-shattering, which implies the existence of $\pi\in \Pi_{(s, S)}$ such that
\begin{align*}
\ell(\pi,\hat{\bd}_i) > \tau_i +\gamma\ \ \ \  \text{if $\frac{1}{m}-\frac{M_i}{N} > 0$},\\
\ell(\pi,\hat{\bd}_i) \leq \tau_i -\gamma\ \ \ \  \text{if $\frac{1}{m}-\frac{M_i}{N} \leq 0$}.
\end{align*}
By Theorem 1 of \cite{berend2013sharp}, for each $N \geq m\geq 2$,
$$
  \mathbb{E}\left[\left|\mathrm{Bin}\left(N,\frac{1}{m}\right)-\frac{N}{m}\right|\right] \geq \frac{1}{\sqrt{2}}{\sqrt{N\frac{1}{m}\left(1-\frac{1}{m}\right)}}.
$$
Note that \eqref{eq-101} in the proof of Theorem \ref{thm-ss-lb-2} shows that there exists an absolute constant $T_0$ such that $\Pgamma{\gamma}\left(\cL (\Pi_{(s, S)})\right)\geq \frac{5}{18}\log T/\log\log T $ and $\Pgamma{\gamma}\left(\cL (\Pi_{(s, S)})\right)\geq 6$ for all $T\geq T_0$. It follows that
\begin{align*}
    \mathbb{E}_{ \bd_1, \ldots, \bd_N  \sim \mathcal{D} }\left[ \GE \left(\cL( \Pi_{(s,S)}) \right) \right] \geq \frac{\gamma}{\sqrt{2}}\sqrt{\frac{\Pgamma{\gamma}\left(\cL (\Pi_{(s, S)})\right)-1}{N}} \geq \frac{\gamma}{\sqrt{2}} \sqrt{\frac{ \frac{5}{6} \cdot  \frac{5}{18}\log T/\log\log T}{N}} \quad  \forall T\geq T_0.
\end{align*}
The proof of Corollary \ref{col-ss-lb-2} is completed.

\subsection{Proof of \Cref{thm-st-dim}} \label{pf:thm-st-dim}

It suffices to show that the statement holds for $t=T+L$. Suppose $\Pgamma{\gamma} (\mathcal{Y} ^{T+L} (\Pi_{(S^t)} ) ) =m $, that is, there exist demand samples $\bd_1,\ldots, \bd_m$ that are $\gamma$-shattered with witnesses $\tau_1,\ldots, \tau_m$. Note that we must have $  \tau_i \in [-1, 1] $ for $ i\in [m]$, because the normalized inventory levels as defined by $\mathcal{Y} ^{T+L} (\Pi_{(S^t)} )$ lie in $[-1,1]$. Applying the definition of Pseudo$_\gamma$-dimension, for any fixed $i\in [m]$, there exists a policy $(S_i^t)_{t=1}^{T+L}$ such that
\[
\begin{aligned}
\frac{y^{T+L} \left((S^{t }_i)_{t =1}^{T }, \bd_i \right)}{(L+1)U}   > \tau_i + \gamma, \text{ and } \frac{y^{T+L} \left((S^{t }_i)_{t =1}^{T }, \bd_j \right)}{(L+1)U} \leq \tau_j - \gamma,\ \forall j \in [m]\setminus \{i\}.
\end{aligned}
\]
Note that $y^{T+L}$ is affected by the first $T$ base-stock levels $(S^{t }_i)_{t =1}^{T }$ due to the lead time $L$.
Let $t_i$ denote the last reorder point that affects the inventory level $y^{T+L} \left((S^{t }_i)_{t =1}^{T }, \bd_i \right)$, namely,
$t_i \triangleq \max\left\{ t'\in [T]: S^{t'}_i - I^{t'}\left((S^{t }_i)_{t =1}^{T }, \bd_i \right) > 0\right\}$, where $I^{t'}\left((S^{t }_i)_{t =1}^{T }, \bd_i \right) $ denotes the inventory position at the beginning of period $t'$ before replenishment by the policy $(S^{t }_i)_{t =1}^{T}$ on the demand sequence $\bd_i$. Define $\bd_i [t_1, t_2] \triangleq \sum_{t=t_1}^{t_2} d_i^t$ for all $1\leq t_1\leq t_2\leq T+L$.
It follows that
\begin{align}
    & y^{T+L} \left((S^{t }_i)_{t =1}^{T }, \bd_i\right)   = S_i^{t_i} - d_i[t_i, T+L-1], \nonumber\\
     &y^{T+L} \left((S^{t }_i)_{t =1}^{T }, \bd_j\right)   \geq  \max \left\{S^{t_i}_i , I^{t_i}\left((S^{t }_i)_{t =1}^{T }, \bd_j \right)\right\} - d_j[t_i, T+L-1]\geq S_i^{t_i} - d_j[t_i, T+L-1], \ \forall   j\in [m]\setminus \{i\} \nonumber.
\end{align}
Thus, we have for all $i\in[m]$ and $j\in[m]\setminus\{i\}$
$$
\frac{S_i^{t_i}- d_i[t_i, T+L-1]}{(L+1)U} >  \tau_i + \gamma
\text{ and } \frac{S_i^{t_i} -  d_j[t_i, T+L-1]}{(L+1)U} \leq  \tau_j - \gamma,
$$
from which we derive
\begin{align}
&& \tau_j - \gamma + \frac{ d_j[t_i, T+L-1]}{(L+1)U} \ge \frac{S_i^{t_i}}{(L+1)U} > \tau_i + \gamma + \frac{d_i[t_i, T+L-1]}{(L+1)U}\nonumber
\\ \Longrightarrow &&   \frac{d_j[t_i, T+L-1]} {(L+1)U} +  \tau_j \geq \frac{d_i[t_i, T+L-1]}{(L+1)U}  +  \tau_i  +2 \gamma.
\label{eq-51}
\end{align}

Without loss of generality, we assume that $t_1\geq t_2 \geq \cdots \geq t_m$. We claim that  for all $i\in [m]$,
\begin{equation}\label{eq-82}
 \frac{d_i[t_i, T+L-1]}{(L+1)U}   +  \tau_i \geq 2(i-1)\gamma- 1.
\end{equation}
It is straightforward to see that the claim holds for $i=1$, because $(d_1^t)_{t=1}^{T+L}$ are non-negative and $\tau_1 \geq -1$. Now suppose that (\ref{eq-82}) holds for some $i\in [m-1]$.
It follows that
\begin{align*}
\frac{d_{i+1}[t_{i+1}, T+L-1]}{(L+1)U}+  \tau_{i+1} \geq \frac{d_{i+1}[t_{i}, T+L-1]}{(L+1)U}+  \tau_{i+1} \geq \frac{d_{i}[t_{i}, T+L-1]}{(L+1)U}+  \tau_{i} + 2\gamma \geq 2i\gamma - 1,
\end{align*}
where the first inequality holds because $t_{i}\geq t_{i+1}$, the second inequality holds by applying \eqref{eq-51} with $j=i+1$, and the third inequality holds by the induction hypothesis.
This completes the induction and we have that the claim in \eqref{eq-82} holds.

To bound from above the LHS of~\eqref{eq-82}, we use the fact that $d_i[t_i, T-1] \leq H$, which holds because $S_i^t \leq H$ and policy $(S_i^t)_{t=1}^T$ when executed on $\bd_i$ does not replenish again up to time $T$. Therefore,
\begin{equation}
\frac{d_i [t_i, T+L-1]}{(L+1)U}  =\ \frac{d_i [t_i, T-1] + d_i [T, T+L-1]}{(L+1)U}
\leq\ \frac{H+LU}{(L+1)U}\leq\ 2. \nonumber
\end{equation}
Combining this with~\eqref{eq-82} and substituting $i=m$ yields $2(m-1)\gamma-1 \le 2+\tau_m$. Finally, we know that $\tau_m\leq 1$, from which it follows that $m\leq 2/\gamma + 1$, completing the proof of Theorem \ref{thm-st-dim}.

\subsection{Proof of \Cref{col-st}} \label{pf:col-st}

We first decouple $\mathbb{E}_{ \bd_1, \ldots, \bd_N  \sim \mathcal{D} }\left[ \GE(\cL   (\Pi_{(S^t)})) \right]$ into  the expected generalization error of $(S^{t'})_{t'=1}^{t-L}$ policies for time $t=L+1,\ldots, T+L$:
\begin{align}
     \nonumber & \mathbb{E}_{ \bd_1, \ldots, \bd_N  \sim \mathcal{D} }\left[ \GE \left(\cL   (\Pi_{(S^t)}) \right)\right]\\
     =\ \nonumber &\mathbb{E}_{\bd_1, \ldots, \bd_N  \sim \mathcal{D} } \left[ \sup_{(S^t)_{t=1}^{T+L} \in \Pi_{(S^t)}} \left\{ R ((S^t)_{t=1}^{T+L} ) -  \hat{R}((S^t)_{t=1}^{T+L} ) \right\} \right] \\
     \nonumber =\ & \mathbb{E}_{\bd_1, \ldots, \bd_N  \sim \mathcal{D} } \left[ \sup_{(S^{t'})_{t'=1}^{T+L} \in [0,H]^{T+L}} \left\{ \mathbb{E}_{\bd\sim \mathcal{D}} \left[\frac{1}{T} \sum_{t=L+1}^{T+L} \ell^t \left((S^{t'})_{t'=1}^{T+L}, \bd\right) \right] -  \frac{1}{N} \sum_{i=1}^N\frac{1}{T} \sum_{t=L+1}^{T+L} \ell^t \left((S^{t'})_{t'=1}^{T+L}, \bd_i\right) \right\} \right]\\
    \nonumber  =\ & \mathbb{E}_{\bd_1, \ldots, \bd_N  \sim \mathcal{D} } \left[ \sup_{(S^{t'})_{t'=1}^{T+L} \in [0,H]^{T+L}} \left\{\frac{1}{T} \sum_{t=L+1}^{T+L}  \left(\mathbb{E}_{\bd\sim \mathcal{D}} \left[ \ell^t \left((S^{t'})_{t'=1}^{t-L}, \bd\right) \right] -  \frac{1}{N} \sum_{i=1}^N  \ell^t \left((S^{t'})_{t'=1}^{t-L},\bd_i \right)\right) \right\} \right]\\
    \nonumber  \leq \ & \mathbb{E}_{\bd_1, \ldots, \bd_N  \sim \mathcal{D} } \left[\frac{1}{T} \sum_{t=L+1}^{T+L}  \sup_{ (S^{t'})_{t'=1}^{t-L} \in [0,H]^{t-L}} \left\{  \mathbb{E}_{\bd\sim \mathcal{D}} \left[ \ell^t \left((S^{t'})_{t'=1}^{t-L}, \bd\right) \right] -  \frac{1}{N} \sum_{i=1}^N  \ell^t \left((S^{t'})_{t'=1}^{t-L}, \bd_i \right) \right\} \right]\\
     \nonumber = \ & \frac{1}{T} \sum_{t=L+1}^{T+L} \mathbb{E}_{\bd_1, \ldots, \bd_N  \sim \mathcal{D}}\left[  \sup_{ (S^{t'})_{t'=1}^{t-L} \in [0,H]^{t-L}} \left\{  \mathbb{E}_{\bd\sim \mathcal{D}} \left[ \ell^t \left((S^{t'})_{t'=1}^{t-L}, \bd\right)\right] -  \frac{1}{N} \sum_{i=1}^N  \ell^t \left((S^{t'})_{t'=1}^{t-L}, \bd\right) \right\} \right]\\
       = \ & \frac{1}{T} \sum_{t=L+1}^{T+L} \mathbb{E}_{ \bd_1, \ldots, \bd_N  \sim \mathcal{D} }\left[ \GE \left(\cL^t   (\Pi_{(S^t)}) \right)\right],
      \label{eq-52}
\end{align}
where we define function classes $\cL^t (\Pi_{(S^t)}) \triangleq \left\{\ell^t \left((S^{t'})_{t'=1}^{t-L},\cdot \right): (S^{t'})_{t'=1}^{t-L} \in [0,H]^{t-L} \right\} $ for $t = L+1, \ldots, T+L$.
By Proposition \ref{prop-rademacher},
\begin{align} \label{eq-53}
 \mathbb{E}_{ \bd_1, \ldots, \bd_N  \sim \mathcal{D} }\left[ \GE \left(\cL^t   (\Pi_{(S^t)}) \right)\right] \leq 2\mathbb{E}_{\bd_1, \ldots, \bd_N  \sim \mathcal{D}}\left[ \Rade\left( \cL^t   (\Pi_{(S^t)})  \circ (\bd_i)_{i=1}^N \right) \right].
\end{align}
Note that $\ell^t \left((S^{t'})_{t'=1}^{t-L},  \bd \right)$ is 1-Lipschitz in $y^t \left((S^{t'})_{t'=1}^{t-L}, \bd \right)$, and thus is $((L+1)U)$-Lipschitz in $y^t \left((S^{t'})_{t'=1}^{t-L}, \bd \right)/((L+1)U)$.
By the Talagrand's contraction lemma (e.g., Lemma 26.9 of \citealt{shalev2014understanding}), it follows that
\begin{align}
    \label{eq-54} \mathbb{E}_{\bd_1, \ldots, \bd_N  \sim \mathcal{D} }\left[ \Rade\left(\cL^t   (\Pi_{(S^t)}) \circ (\bd_i)_{i=1}^N \right) \right]
    \leq \ (L+1)U \mathbb{E}_{\bd_1, \ldots, \bd_N  \sim \mathcal{D}}\left[ \Rade\left(\mathcal{Y}^t   (\Pi_{(S^t)}) \circ (\bd_i)_{i=1}^N \right) \right].
\end{align}
By Proposition \ref{prop-gamma-dim}, there exists constants $C_1, C_2>0$ such that
\begin{align}
   \label{eq-55} \mathbb{E}_{\bd_1, \ldots, \bd_N  \sim \mathcal{D}}\left[ \Rade\left(\mathcal{Y}^t   (\Pi_{(S^t)}) \circ (\bd_i)_{i=1}^N \right) \right] \leq \ & \frac{C_1}{\sqrt{N}} \int_0^1 \sqrt{ \log\left(\frac{2}{\gamma} \right)\cdot \Pgamma{C_2\gamma}\left(\mathcal{Y}^t   (\Pi_{(S^t)})\right) }d\gamma.
\end{align}
Note that by Theorem \ref{thm-st-dim}, $\Pgamma{\gamma}\left(\mathcal{Y}^t   (\Pi_{(S^t)})\right) \leq 2/\gamma +1 $ for all $t=L+1, \ldots, T+L$ and  $\gamma\in (0, 1]$, and it is straightforward to see that   $\Pgamma{\gamma}\left(\mathcal{Y}^t   (\Pi_{(S^t)})\right) =0 $ for any $\gamma>1$, because $\mathcal{Y}^t   (\Pi_{(S^t)})$ is bounded in $[-1, 1]$. Therefore, combining \eqref{eq-52}-\eqref{eq-55}, we have
\begin{align}
   \nonumber & \mathbb{E}_{ \bd_1, \ldots, \bd_N  \sim \mathcal{D} }\left[ \GE \left(\cL   (\Pi_{(S^t)}) \right)\right]\\
  \nonumber \leq \ & \frac{2(L+1)U C_1}{\sqrt{N}} \int_0^1 \sqrt{ \log \left(\frac{2}{\gamma} \right)\cdot \Pgamma{C_2\gamma} \left(\mathcal{Y}^t   (\Pi_{(S^t)}) \right) }\ d\gamma\\
  \nonumber \leq \ & \frac{2(L+1)U C_1}{\sqrt{N}} \int_0^1 \sqrt{ \log \left(\frac{2}{\gamma} \right)\cdot\left( \frac{2}{C_2\gamma} +1 \right) }\ d\gamma \\
  \nonumber \leq \ & \frac{2(L+1)UC_1}{\sqrt{N}} \int_0^1 \sqrt{ \log \left(\frac{2}{\gamma} \right)\cdot \max\left\{\frac{2}{C_2}, 1\right\} \cdot \frac{2}{\gamma}  }\ d\gamma\\
 \nonumber \leq \ &  2C_1 \sqrt{\max\left\{\frac{2}{C_2}, 1\right\}} \left(2\sqrt{2 \log 2} + 4\sqrt{2\pi}\right) (L+1)U\sqrt{\frac{1}{N}}.
\end{align}
The last inequality holds because, by changing the variable $\gamma=2e^{-t^2}$,
\begin{align*}
    \int_0^1 \sqrt{ \log \left(\frac{2}{\gamma} \right) \cdot \frac{2}{\gamma}  }\ d\gamma
    &= 4 \int_{\sqrt{\log 2}}^{\infty} t^2  e^{-\frac{t^2}{2}}  dt
    = 4 \left(
  -t e^{-\frac{t^2}{2}} \Big|_{\sqrt{\log 2}}^{\infty}  +  \int_{\sqrt{\log 2}}^{\infty}  e^{-\frac{t^2}{2}}  dt \right)
  \leq 2\sqrt{2 \log 2} + 4\sqrt{2\pi}.
\end{align*}
The proof of Corollary \ref{col-st} is completed.

\end{document}